\documentclass{article}

\PassOptionsToPackage{numbers, compress}{natbib}

\usepackage[preprint]{neurips_2025}

\usepackage[utf8]{inputenc} 
\usepackage[T1]{fontenc}    
\usepackage{hyperref}       
\usepackage{url}            
\usepackage{booktabs}       
\usepackage{amsfonts}       
\usepackage{nicefrac}       
\usepackage{microtype}      
\usepackage[table]{xcolor}         
\usepackage{graphicx}
\usepackage{amsmath}
\usepackage{amssymb}
\usepackage{bbm}
\usepackage{rotating}
\usepackage{multicol}
\usepackage{multirow}
\usepackage{algorithm}
\usepackage{algorithmic}
\usepackage{tocbasic}
\usepackage{titletoc}
\usepackage{bm} 
\usepackage{subcaption}
\usepackage{tablefootnote}
\usepackage{wrapfig}

\definecolor{limegreen}{rgb}{0.2, 0.8, 0.2}
\definecolor{worsered}{HTML}{B80000}
\definecolor{venetianred}{rgb}{0.78, 0.03, 0.08}
\definecolor{princetonorange}{rgb}{1.0, 0.56, 0.0}
\definecolor{royalpurple}{rgb}{0.47, 0.32, 0.66}
\definecolor{airforceblue}{rgb}{0.36, 0.54, 0.66}
\definecolor{customorange}{HTML}{ff7f0e}
\definecolor{customblue}{HTML}{1f77b4}
\definecolor{customgold}{HTML}{ffcd3f}
\definecolor{customyellow}{HTML}{f6b600}
\definecolor{myblue}{HTML}{2353bf}
\definecolor{custombrown}{HTML}{8c564b}

\definecolor{teasergreen}{HTML}{588937}
\definecolor{teaserorange}{HTML}{D56315}

\definecolor{umapred}{HTML}{800337}
\definecolor{umapyellow}{HTML}{ffc000}
\definecolor{umapgreen}{HTML}{548235}
\definecolor{umaporange}{HTML}{f3b066}

\definecolor{toyorange}{HTML}{EE7100}
\definecolor{toyred}{HTML}{F54B2F}
\definecolor{toyblue}{HTML}{4472C4}

\def\FullMethodName{Canonical Latent Representation Identifier}
\def\FullMethodNameBold{\textbf{C}anonical \textbf{LA}tent \textbf{R}epresentation \textbf{ID}entifier}
\def\Cano{CLARep}
\def\FullCano{Canonical Latent Representation}
\def\FullCanoBold{\textbf{C}anonical \textbf{LA}tent \textbf{Rep}resentation}
\def\Canofeat{Canonical Feature}
\def\Canoimg{Canonical Sample}
\def\MethodName{CLARID}
\def\CanoDistill{CaDistill}

\def\ExtraDir{extraneous direction}

\def\DMFit{DMDistill}
\def\DMDistill{CFGDistill}

\def\DDIMinv{$F_{inv}$}
\def\DDIMgen{$F_{dec}$}

\definecolor{myred}{HTML}{FF0000} 
\definecolor{mypurple}{HTML}{7030A0} 

\usepackage{pifont}
\usepackage{enumitem}
\usepackage{wrapfig}

\title{\FullCano{}s \\ in Conditional Diffusion Models}

\author{%
  Yitao Xu \quad Tong Zhang \quad Ehsan Pajouheshgar  \quad Sabine Süsstrunk\\
  Image and Visual Representation Lab \\
  École polytechnique fédérale de Lausanne,Lausanne, Switzerland \\
  \texttt{\{yitao.xu,tong.zhang,ehsan.pajouheshgar,sabine.susstrunk\}@epfl.ch} \\
}

\begin{document}

\maketitle

\begin{abstract}
    Conditional diffusion models (CDMs) have shown impressive performance across a range of generative tasks. Their ability to model the full data distribution has opened new avenues for analysis-by-synthesis in downstream discriminative learning. However, this same modeling capacity causes CDMs to entangle the class-defining features with irrelevant context, posing challenges to extracting robust and interpretable representations.
    To this end, we identify \textit{\FullCanoBold{}s} (\Cano{}s), latent codes whose internal CDM features preserve essential categorical information while discarding non-discriminative signals. When decoded, \Cano{}s produce representative samples for each class, offering an interpretable and compact summary of the core class semantics with minimal irrelevant details.
    %
    Exploiting \Cano{}s, we develop a novel diffusion-based feature distillation paradigm, \textbf{\textit{\CanoDistill{}}}. While the student has full access to the training set, the CDM as teacher transfers core class knowledge only via \Cano{}s, which amounts to merely 10\% of the training data in size.
    After training, the student achieves strong adversarial robustness and generalization ability, focusing more on the class signals instead of spurious background cues.
    %
    Our findings suggest that CDMs can serve not just as image generators but also as compact, interpretable teachers that can drive robust representation learning.

\end{abstract}

\section{Introduction}
\begin{wrapfigure}{r}{0.54\linewidth}
    \vspace{-15pt}
    \centering
    \includegraphics[width=\linewidth, trim={228 78 230 41},clip]{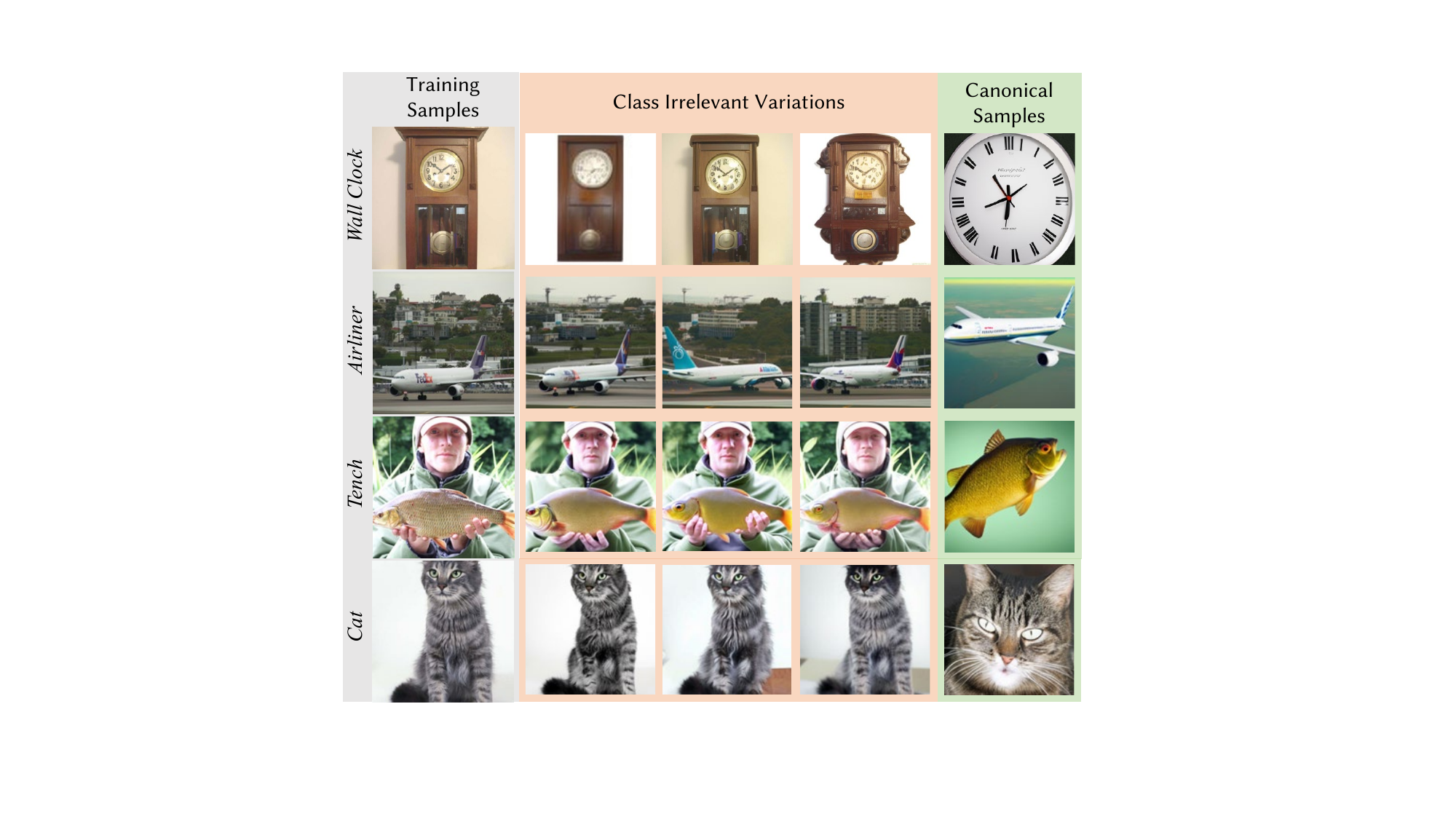}
    \caption{
    Conditional diffusion models (CDMs) encode both \textcolor{teasergreen}{core class features} and \textcolor{teaserorange}{irrelevant context} in a CDM.
    \textbf{\textcolor{gray}{Left}}: training samples.
    \textbf{\textcolor{teaserorange}{Middle}}: samples obtained by modifying CDM latent codes that preserve the class label but alter class-irrelevant parts.
    \textbf{\textcolor{teasergreen}{Right}}: \Canoimg{}s produced by \MethodName{}, which retain the \textcolor{teasergreen}{class-defining content} while removing \textcolor{teaserorange}{extraneous context}. \MethodName{} benefits the extraction of robust and interpretable representations from the CDMs.
    }
    \label{fig:teaser}
\end{wrapfigure}

Diffusion models (DMs) excel at generative modeling of images~\cite{dm-beats-gan-synthesis,ddpm,ldm,ddim,score-gen}. When conditioned on class labels \cite{uvit,cfg,dit} or text prompts \cite{sdxl,ldm}, conditional diffusion models (CDMs) faithfully generate samples with desired characteristics of the condition. 
This generative capability has sparked a wave of analysis-by-synthesis approaches~\cite{dm-discriminative5-ddpm-seg,dm-discriminative4-kaiming,dreamteacher,dm-discriminative2-not-all-dm-eval,dm-discriminative1,dm-discriminative3-ddae,dm-as-representation-learner,dm-discriminative7-sd-transfer-brute,dm-discriminative6-seg-cross-attn}, where DMs are used to probe or improve downstream discriminative tasks. However, a key challenge remains: since DMs model the full data distribution, they often encode redundant or irrelevant information, which can obscure the discriminative signal. 
For example, in the \textit{Tench} row of Figure \ref{fig:teaser}, modifying the CDM latent code of the training sample changes the angler in the background while the fish remains almost unchanged, showing that the model encodes background cues that correlate with the class but not semantically essential to it.
%
This entanglement between class semantics and extraneous factors limits interpretability and hinders the effective use of CDMs in representation learning.
This motivates our central question:
\begin{center}
  \begin{minipage}{.95\linewidth} 
    \vspace{-6pt}
    \emph{How can we identify the underlying core categorical semantics in a conditional diffusion model?}
  \end{minipage}
  \vspace{-6pt}
\end{center}
We answer this by introducing \FullMethodNameBold{} (\textbf{\MethodName{}}), a method for identifying the latent codes in CDMs that capture essential categorical information and filter out irrelevant details. We call the resulting latent codes \FullCanoBold{}s (\Cano{}s).
We begin with the assumption that the essential semantics of a class typically lie on a low-dimensional manifold embedded in the high-dimensional data space~\cite{deep-feature-factorization,diffusion-low-dim,zhang2019scalable}. We postulate that such semantic manifolds also exist within the latent space of diffusion models. 
Our key insight is that altering the latent code along the tangent directions of the manifold, referred to as \ExtraDir{}s, modifies visual appearance without affecting class identity.
We find that projecting out the \ExtraDir{}s in the latent space of CDMs effectively eliminates class-irrelevant factors such as background clutter or co-occurring objects from other categories. When projected back to the data space, \Cano{}s produce representative samples of each category, namely \textit{\Canoimg{}s}, providing an intuitive and interpretable summary of the essential categorical semantics. Additionally, the internal CDM features of \Cano{}s, \textit{i.e. \Canofeat{}s}, contain mostly the core class information.
We first validate our method in a toy hierarchical generative model, where \MethodName{} recovers a low-dimensional class manifold while standard classifier-free guidance (CFG) produces dispersed, high-likelihood samples. Scaling up, we apply \MethodName{} to ImageNet-pretrained CDMs and develop strategies for finding the projection time step and the number of \ExtraDir{}s. 
Our method also generalizes to text-conditioned DMs and is compatible with different diffusion samplers.

Building on the discovery of \Cano{}, we propose a novel feature distillation paradigm, \textbf{\textit{\CanoDistill{}}}. 
This method leverages the interpretable nature of \Cano{}s, which encapsulates the core semantics of each class, to supervise a student network.
%
\textbf{\textit{\CanoDistill{}}} aligns the student network's representations on both \Canoimg{}s and original training samples with \Canofeat{}s using a novel feature distillation loss, which helps the student network encode the core class information. 
The student network's representations of the original training samples are also forced to be close to those of the \Canoimg{}s, treating them as anchors in the student's representation space. 
%
The student learns on the full training set, while the teacher CDM transfers the essential class knowledge by only exploiting \Cano{}s, which amounts to merely 10\% of the training data in size. In contrast, existing state-of-the-art methods require transferring the teacher CDM’s knowledge using the entire training set.
\textbf{\textit{\CanoDistill{}}} improves the student's adversarial robustness as well as the generalization capability on CIFAR10 \cite{cifar10} and ImageNet \cite{deng2009imagenet}. Our contributions are as follows:
\begin{itemize}[leftmargin=*]
    \item We introduce \textit{\FullCanoBold{}} (\Cano{}) in CDMs—latent codes whose internal CDM features, \textit{i.e. \Canofeat{}s}, encapsulate core categorical semantics with minimal irrelevant signals. When decoded to the data space, these latent codes produce \textit{\Canoimg{}s}, which serve as compact and interpretable prototypes for each class.
    \item To extract \textit{\Cano{}s}, we propose \FullMethodNameBold{} (\MethodName{}), a method that identifies these latent codes by projecting out non-discriminative directions in the CDM’s latent space. The optimal configurations of \MethodName{} are selected through a systematic analysis of the CDM's features.
    \item Leveraging the \Cano{}s, we develop a novel diffusion-based feature distillation paradigm, \textbf{\textit{\CanoDistill{}}}. While the student is being trained on the full training set, the CDM as the teacher transfers essential class knowledge only via exploiting \Cano{}s, which amounts to merely 10~\% of the data. The resulting student network achieves strong adversarial robustness and generalization performance, focusing more on the core class information.
\end{itemize}

\vspace{-5pt}
\section{Related Works}
\vspace{-5pt}
\subsection{Interpretability in diffusion models}
\vspace{-3pt}
Recent research in diffusion models (DMs) reveals the interpretable semantic information in them. We categorize these efforts into two main groups. The first line of this work focuses on semantic editing by manipulating the reverse diffusion trajectory to produce semantically meaningful changes in generated images \cite{jac-low-dim-local,icfg-jac-edit,dm-already-semantic,riemannian-diffusion-edit}.
\citet{dm-already-semantic} uncover a semantic space inside a pre-trained DM, termed h-space, where particular vector directions yield high-quality image editing results. \citet{riemannian-diffusion-edit} analyze the latent input space, namely x-space, of DMs from a Riemannian geometry perspective. They define the pullback metric on x-space from the h-space Euclidean metric, obtaining certain vector directions in x-space that can yield semantic editing results. 
\citet{jac-low-dim-local} provides more theoretical insights into this framework and extends it to local editing scenarios. 

Another line of work leverages the attention mechanism \cite{nlp-transformer} in DMs to interpret the conditional information \cite{crossattn5,prompt2prompt-crossattn1,crossattn6,crossattn2,crossattn4,crossattn3}. \citet{prompt2prompt-crossattn1} propose that the cross-attention map between the text prompts and image tokens in text-conditioned DMs encodes rich spatial cues. 
Building on this insight, subsequent studies analyze these attention maps to improve the precision and controllability of semantic image editing \cite{crossattn5,crossattn6,crossattn2}.
%
Other efforts investigate this property in visual recognition tasks such as semantic segmentation \cite{crossattn4,crossattn3}, using the attention maps to interpret the model’s spatial reasoning. All existing methods either focus on image editing or are tailored to a specific DM architecture. We take the first step to uncover the underlying core class semantics in DMs without any supervision. Our method is compatible with different DM architectures and samplers.

\vspace{-5pt}
\subsection{DMs as teachers in analysis-by-synthesis}
\vspace{-4pt}
\textbf{DM-based feature distillation}. 
Recent works show that the intermediate features in DMs contain rich discriminative information \cite{dm-discriminative5-ddpm-seg,dm-discriminative4-kaiming,dreamteacher,dm-discriminative2-not-all-dm-eval,dm-discriminative1,dm-discriminative3-ddae,dm-as-representation-learner,dm-discriminative7-sd-transfer-brute,dm-discriminative6-seg-cross-attn}. Here, we focus on the utilization of DMs as teachers in feature distillation frameworks. \citet{dreamteacher} proposes a framework in which the intermediate features of the student network are aligned to those of a generative teacher, improving the student's performance on dense prediction tasks. \citet{dm-as-representation-learner} use reinforcement learning to select a proper diffusion timestep for distillation, enhancing the student's performance in image classification, semantic segmentation, and landmark detection benchmarks. 

\textbf{DMs for data generation and augmentation}. 
DMs faithfully model the full training data distribution, which allows the generation of new training samples or augmentation of the existing ones \cite{diffusion-aug-first,aug1m,dreamda-semantic-perturb,gen-data-robust1-advtrain,gen-data-zero-shot-few-shot,gen-data-diffusemix,imagenet-clone,gen-data-robust2-adv,diffaug,gen-data-da-fusion-few-shot,gen-data-robust3-adv}. \citet{gen-data-robust1-advtrain} and \citet{gen-data-robust2-adv} improve the robustness of adversarially-trained classifiers by using diffusion-generated data. \citet{aug1m} demonstrate that supplementing training data with diffusion‑generated images leads to consistent gains on out‑of‑distribution test sets. 
Diffusion-generated data is also effective in data-scarce settings, \textit{i.e.}, zero-shot and few-shot learning \cite{dreamda-semantic-perturb,gen-data-zero-shot-few-shot,gen-data-da-fusion-few-shot}. Regarding data augmentation, \citet{gen-data-diffusemix} propose blending images while preserving their labels using pre-trained text-to-image DMs \cite{ldm}. \citet{diffaug} utilize denoised samples for augmenting the training data, improving the generalization of downstream classifiers. 


Despite the progress, current methods use raw diffusion features and outputs, which contain class-irrelevant information. Such irrelevant signals prevent the student from efficiently and accurately learning the class semantics, leading to vulnerable models. On the contrary, our method transfers the core class semantics using \Cano{}s to the student, enhancing its adversarial robustness and generalization capability. Notably, \Cano{}s amounts to only 10\% of the original data in size.



\vspace{-5pt}
\section{Method}
\vspace{-5pt}
\label{sec:cafol}
\begin{figure*}[]
    \centering
    \includegraphics[width=\linewidth, trim={0 60 0 30},clip]{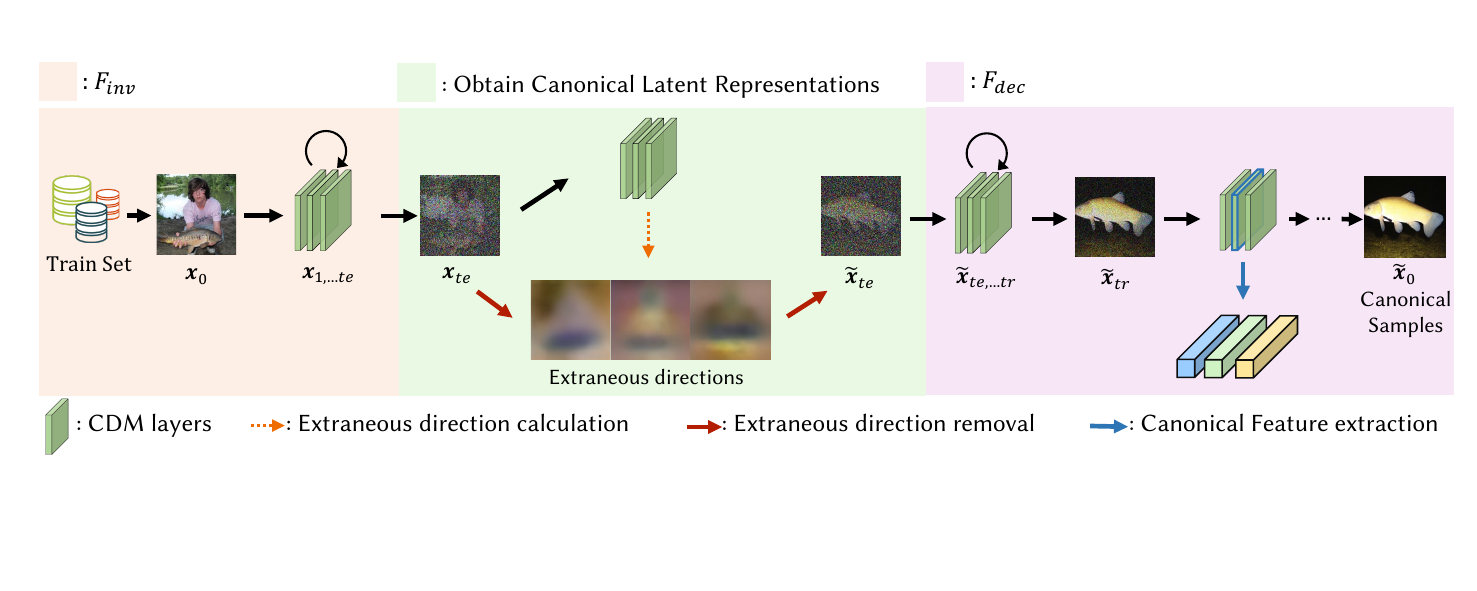}
    \caption{Overview of \FullMethodName{} (\MethodName{}). Starting from a training sample $\bm{x}_0$, we invert it using a CDM until $t=t_e$. We compute the \ExtraDir{}s using the method described in Section~\ref{sec:cafol-overall-pipeline}, and remove them from the inverted sample $\bm{x}_{te}$ to obtain a \Cano{} $\tilde{\bm{x}}_{te}$. We then generate the \Canoimg{} $\tilde{\bm{x}}_0$ and extract the \Canofeat{} at timestep $t=t_r$. The CDM receives the ground truth condition of $\bm{x}_0$ in the whole process, here the \textit{Tench} class.}
    \label{fig:cafol-overview}
    \vspace{-10pt}
\end{figure*}
An overview of \FullMethodName{} (\MethodName{}) is shown in Figure~\ref{fig:cafol-overview}. We describe the procedure of finding the \Cano{} of a given sample in Section~\ref{sec:cafol-overall-pipeline}. We then provide the intuition of the effect of \MethodName{} in a proof-of-concept experiment in Section~\ref{sec:cafol-toy}. 

\vspace{-5pt}
\subsection{Preliminaries}
\vspace{-5pt}
\label{sec:preliminary}

Diffusion models (DMs) generate images by learning to reverse a fixed forward diffusion process \cite{ddpm}. Let $\bm{x}_0$ be a training sample and $t\in\{1,...,T\}$ denote the diffusion time steps. A forward kernel $q(\bm{x}_t\mid \bm{x}_{t-1})$ progressively corrupts $\bm{x}_0$ into a noisy sample $\bm{x}_t$. A neural network $\bm{f}_{\theta,t}(\bm{x}_t)$, parameterized by $\theta$, is trained to approximate the reverse process $p_{\theta}(\bm{x}_{t-1} \mid \bm{x}_t)$. 
More details are in Section~\ref{sec:supp-preliminary} in the Appendix. Certain parameterizations of the diffusion process \cite{edm,edm2,ddim} allow for partial or full inversion of a given input sample, producing a noisy sample~$x_t$ that preserves the semantic information of $x_0$ \cite{goldennoise}.
Hereafter, we denote the inversion process of a sample by \DDIMinv{} and the corresponding decoding process as \DDIMgen{}. 

\subsection{The overall procedure of \FullMethodName{} (\MethodName{})}
\label{sec:cafol-overall-pipeline}
Given a sample $\bm{x}_0$ as the seed, which belongs to class condition $\bm{c}$, we first invert it to timestep $t_e$ via \DDIMinv{}. Denote $\bm{x}_{te}$ as the latent code of $\bm{x}_0$ at $t_e$. In this latent space, we identify a set of \ExtraDir{}s that preserve class identity yet induce large changes.
Concretely, let $\bm{f}_\theta(\cdot)$ be the CDM’s feature extractor at layer $l$ and timestep $t_e$. A first‐order Taylor expansion around $\bm{x}_{te}$ gives:
\begin{equation}
    \bm{f}^{l}_{\theta, te}(\bm{x}_{te}+\bm{v}) \approx \bm{f}^{l}_{\theta, te}(\bm{x}_{te}) + \nabla \bm{f}^{l}_{\theta, te}(\bm{x}_{te}) \cdot \bm{v} = \bm{f}^{l}_{\theta, te}(\bm{x}_{te}) + \bm{J}^{l}_{\theta, te}(\bm{x}_{te}) \cdot \bm{v}.
    \label{eq:first-order-taylor}
\end{equation}
$\bm{J}^{l}_{\theta, te}(\bm{x}_{te})$ denotes the Jabobian of $\bm{f}^{l}_{\theta, te}(\bm{x}_{te})$. Hereafter, we drop $\theta$ and $l$ to avoid clutter. A vector $\bm{v}$ that can cause large changes in the output carries some semantic information \cite{riemannian-diffusion-edit,latent-space-traversal-jac-meaningful}, but such information is \textbf{not necessarily class-relevant}, as shown in Figure~\ref{fig:teaser}.
The change caused by vector $\bm{v}$ is defined as the L2-norm of the Jacobian vector product: $|| \bm{J}_{te}(\bm{x}_{te}) \cdot \bm{v} ||_2$. Accordingly, the directions that lead to large changes in the output are the right singular vectors of $\bm{J}_{te}$. When $\bm{f}$ receives conditional information $\bm{c}$, those directions will \textbf{preserve the class identity} while altering the appearance of the decoded sample. We show examples in Section~\ref{sec:supp-extra-dir-class-preserve} in the Appendix. We therefore remove the components of $\bm{x}_{te}$ along those \ExtraDir{}s to obtain \Cano{} $\tilde{\bm{x}}_{te}$, by projecting $\bm{x}_{te}$ onto the subspace which is orthogonal to these directions, as described in Eq.\ref{eq:project-away}. 
\begin{equation}
    \tilde{\bm{x}}_{te}:= \bm{x}_{te, \perp V_k}
    = \bigl(\mathbf{I} - \mathbf{V}_{k} \mathbf{V}_{k}^{\mathsf T}\bigr)\,\bm{x}_{te}.
    \label{eq:project-away}
\end{equation}
Here, $\bm{V}=[\bm{v}_1,\bm{v}_2,...]$ is the right singular vector matrix of $\bm{J}_{te}$, and $k$ denotes the number of singular vectors to be removed. We then apply \DDIMgen{} to $\tilde{\bm{x}}_{te}$ to obtain one \Canoimg{} $\tilde{\bm{x}}_{0}$ for the input condition $\bm{c}$. Note that the CDM is conditioned on $\bm{c}$ in the whole procedure. 
Additional discussion on the choice of the layer index $l$ for computing $\bm{J}_{te}$ is provided in Section~\ref{sec:supp-jac-layer-index} of the Appendix.

\vspace{-5pt}
\subsubsection{Finding the optimal $t_e$ and $k$}
\label{sec:optimal-kt}
Choosing an appropriate $t_e$ is critical for \MethodName{} to be effective. If $t_e$ is too small, \textit{e.g.,} $t_e \approx \frac{T}{100}$, the synthesized sample barely changes and remains close to the original input. On the other hand, when $t_e$ is too large, the conditional signal becomes ineffective and fails to guide the generation~\cite{critical-stage2,critical-stage1}.
To find the largest time step where the conditional signal is still able to steer the CDM’s output toward the desired class, we perform a two-stage sampling process starting from $p(\bm{x}_T)\sim\mathcal{N}(\bm{0},\bm{I})$:
\vspace{-5pt}
\begin{enumerate}[leftmargin=*]
    \item \textbf{Unconditional stage} \((T \le t < t_e)\):  
          Forward the model without class conditioning, \textit{i.e.}, $\bm{c}=\varnothing$.
    \item \textbf{Conditional stage} \((t_e \le t \le 0)\):  
          The class condition is introduced and retained down to \(t=0\).
\end{enumerate}
\vspace{-5pt}
We generate $m$ samples for each condition in a given dataset, and measure the accuracy of pre-trained classifiers on the generated samples. We find the saturation point of the accuracy curve as the appropriate $t_e$. We show the accuracy curve and more details for an ImageNet $256 \times 256$-pretrained DiT \cite{dit} and a Stable Diffusion model \cite{ldm} in Section~\ref{sec:supp-te} in the Appendix.

We fix the inversion time step $t_e$ across all samples, but the number of directions to discard should be adapted to each sample. 
To decide $k$, we examine how strongly the top-$k$ \ExtraDir{}s alter the sample’s visual appearance, quantified by the explained variance ratio (EVR) $S_k=\sum^k_{i=1} \sigma^2_i / \sum^n_{j=1} \sigma^2_j$, where $\sigma_i$ is the $i$th singular value of $\bm{J}_{te}$ and $n$ is a hyperparameter. Intuitively, the \ExtraDir{} that leads to larger variations carries less core class semantics. We compute a sequence of the EVR $S_1,S_2,...,S_n$ and set the elbow point of the sequence to be the optimal $k$ for a given sample. Compared to a fixed $k$, the adaptive choice will find the point where the effect of the \ExtraDir{}s diminishes.
Visual examples illustrating the necessity of adapting $k$ for each sample, along with details of deciding $n$ and the calculation of the elbow point, are provided in Section~\ref{sec:supp-k} in the Appendix. 

\begin{figure}
    \centering
    \begin{minipage}{0.45\textwidth}
        \centering
        \vspace{-10pt}
        \includegraphics[width=\textwidth,trim={80 80 80 60},clip]{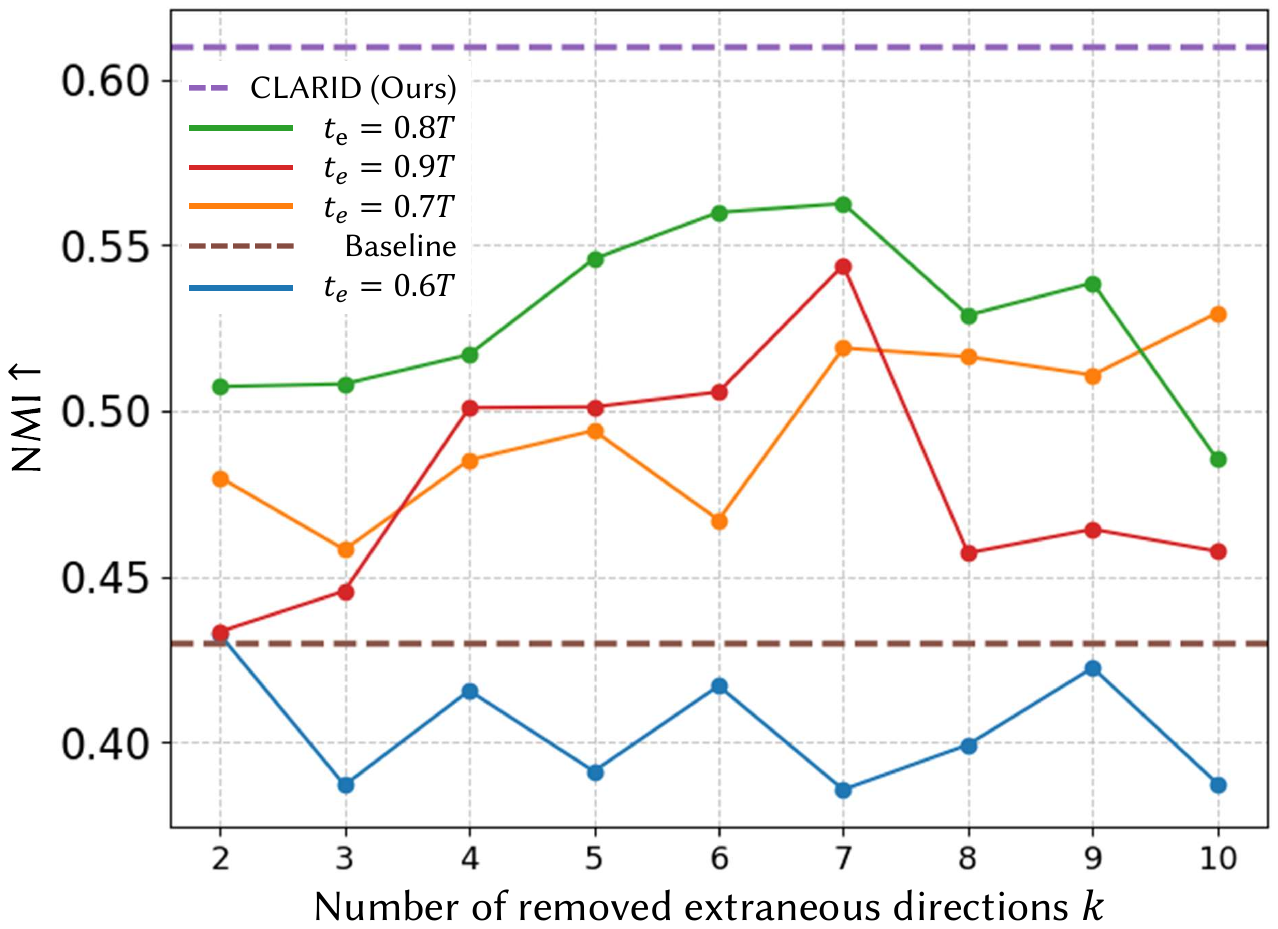}
        \caption{The normalized mutual information (NMI, higher is better) between cluster assignments of CDM features and the ground truth labels. \MethodName{} achieves the highest NMI. Baseline is the original CDM features.}
        \label{fig:nmi}
    \end{minipage}\hfill
    \hspace{1pt}
    \begin{minipage}{0.53\textwidth}
        \centering
        \includegraphics[width=\textwidth,trim={10 85 10 90},clip]{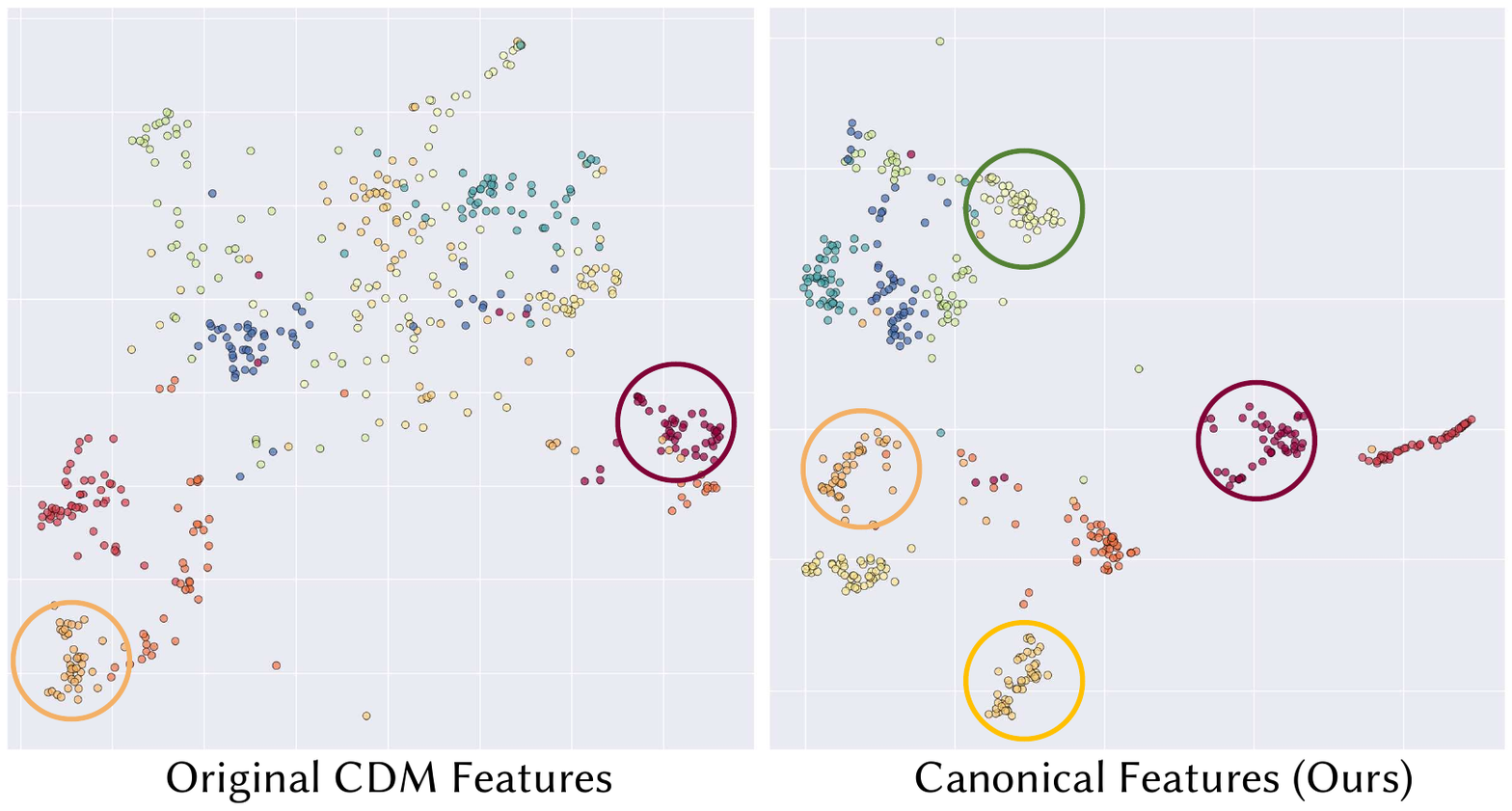}
        \caption{A 2D UMAP \cite{umap} projection of the CDM feature space, showing clusters for ten classes. Colors indicate classes. \MethodName{} yields more compact feature clusters than the original samples (\textcolor{umapgreen}{green}, \textcolor{umapyellow}{yellow}) and preserve the existing ones (\textcolor{umapred}{red}, \textcolor{umaporange}{orange}).}
        \label{fig:umap}
    \end{minipage}
    \vspace{-10pt}
\end{figure}

\textbf{Empirical validation}. To demonstrate the effectiveness of our strategies on finding $t_e$ and $k$, we quantify the feature quality of a CDM and show that our strategies achieve the best one. We also provide visual comparisons in Figure~\ref{fig:jac-kt-compare}, \ref{fig:jac-different-k} in the Appendix. 
Intuitively, by retaining the essential class features with minimal class-irrelevant information, the CDM features associated with the \Cano{}s (\Canofeat{}s) should be more easily separable in the feature space.
To show this, we first perform K-means clustering on the CDM features of 1000 samples from 20 different ImageNet classes (details in Section~\ref{sec:imagenet20} in the Appendix), as well as their corresponding \Canofeat{}s. We then quantify the feature quality using normalized mutual information (NMI) between the cluster assignments and the ground truth class labels. This method is training-free, hence efficient. Section~\ref{sec:imagenet20} in the Appendix provides evidence on the validity of our feature quality quantification method. We perform the analysis on an ImageNet $256 \times 256$-pretrained DiT-XL model \cite{dit}. The result is shown in Figure~\ref{fig:nmi}. Removing different numbers of \ExtraDir{}s, \textit{i.e.,} varying $k$, yields different NMI scores, whereas \MethodName{} adaptively chooses $k$ for each sample and achieves the highest NMI. Qualitatively, \MethodName{} produces feature clusters that are more compact than those formed by the original samples while preserving the existing clusters, as shown in Figure \ref{fig:umap}.
We refer the readers to Section~\ref{sec:imagenet20-layer-timestep} in the Appendix for the details of the feature extraction setup, including both the network layer and its associated time step $t_r$.

\vspace{-5pt}
\subsection{A proof-of-concept experiment}
\vspace{-5pt}
\label{sec:cafol-toy}

\begin{figure}[htp]
  \centering
  \begin{subfigure}[b]{0.36\textwidth}
    \centering
    \includegraphics[width=\textwidth,trim={70 0 80 0},clip]{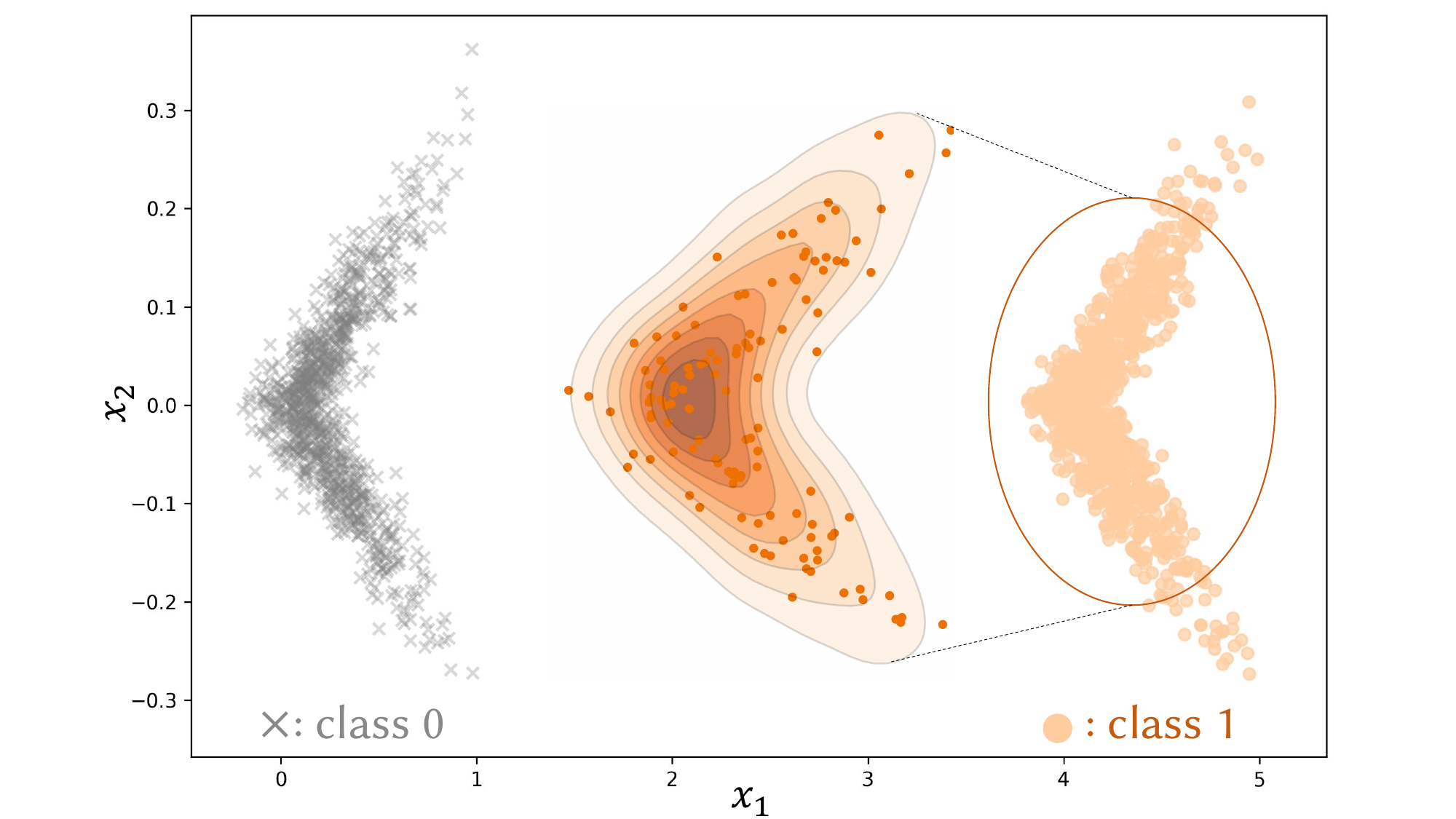}
    \caption{}%
    \vspace{-5pt}
    \label{fig:toy-data}
  \end{subfigure}
  \begin{subfigure}[b]{0.19\textwidth}
    \centering
    \includegraphics[width=\textwidth]{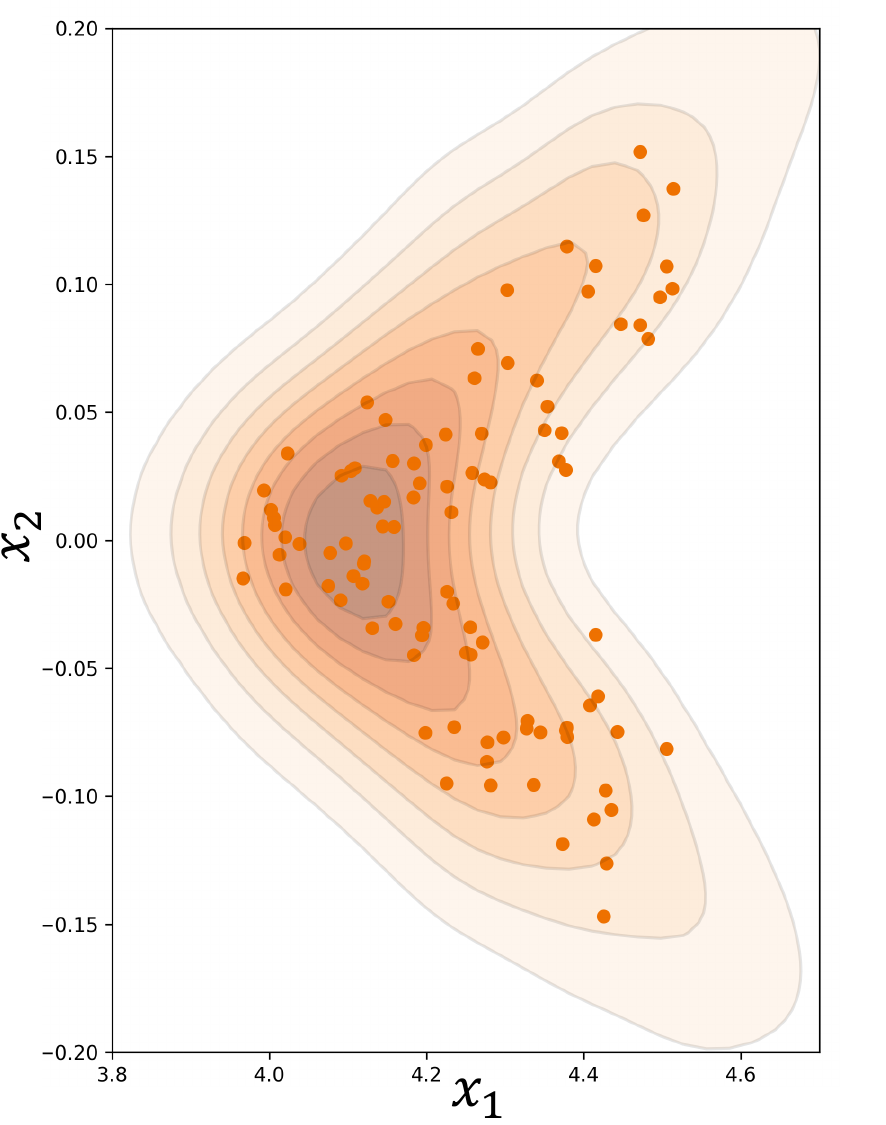}
    \caption{}%
    \vspace{-5pt}
    \label{fig:toy-ori}
  \end{subfigure}
  \hspace{5pt}
  \begin{subfigure}[b]{0.19\textwidth}
    \centering
    \includegraphics[width=\textwidth]{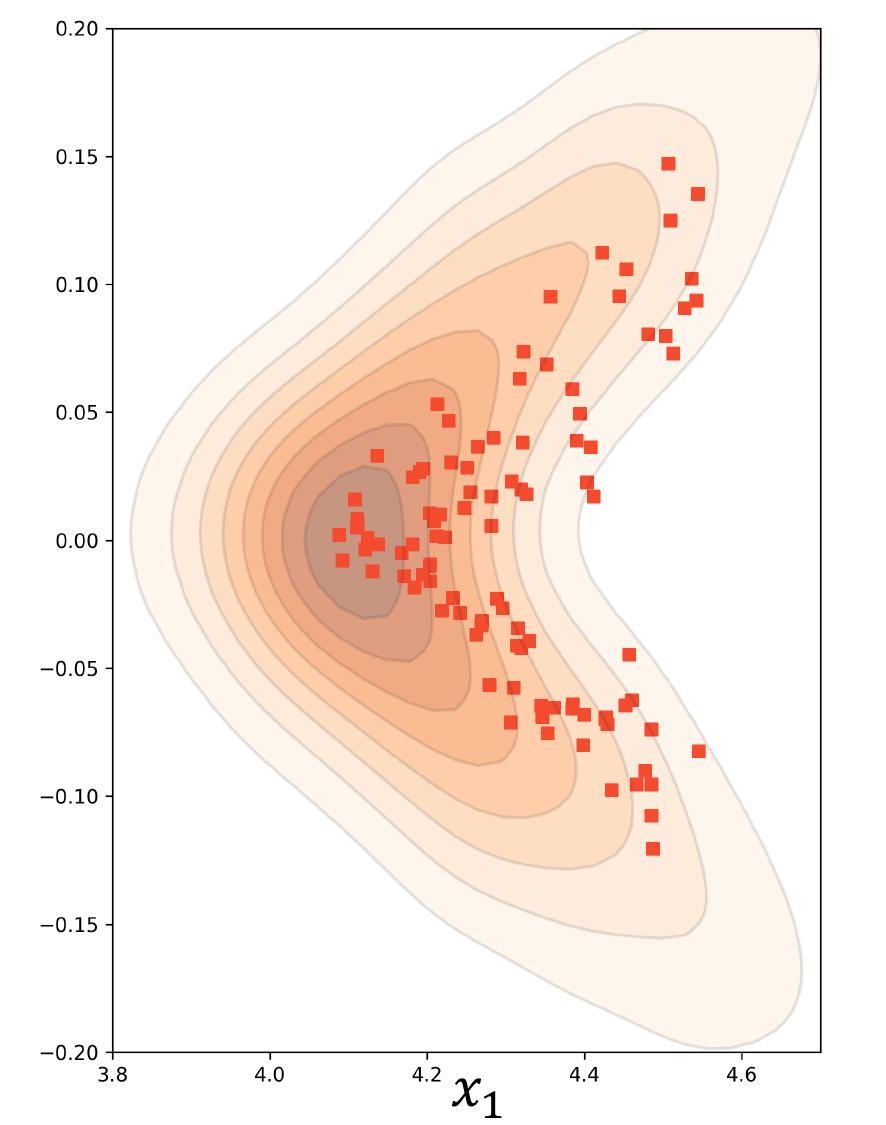}
    \caption{}%
    \vspace{-5pt}
    \label{fig:toy-cfg}
  \end{subfigure}
  \hspace{5pt}
  \begin{subfigure}[b]{0.19\textwidth}
    \centering
    \includegraphics[width=\textwidth]{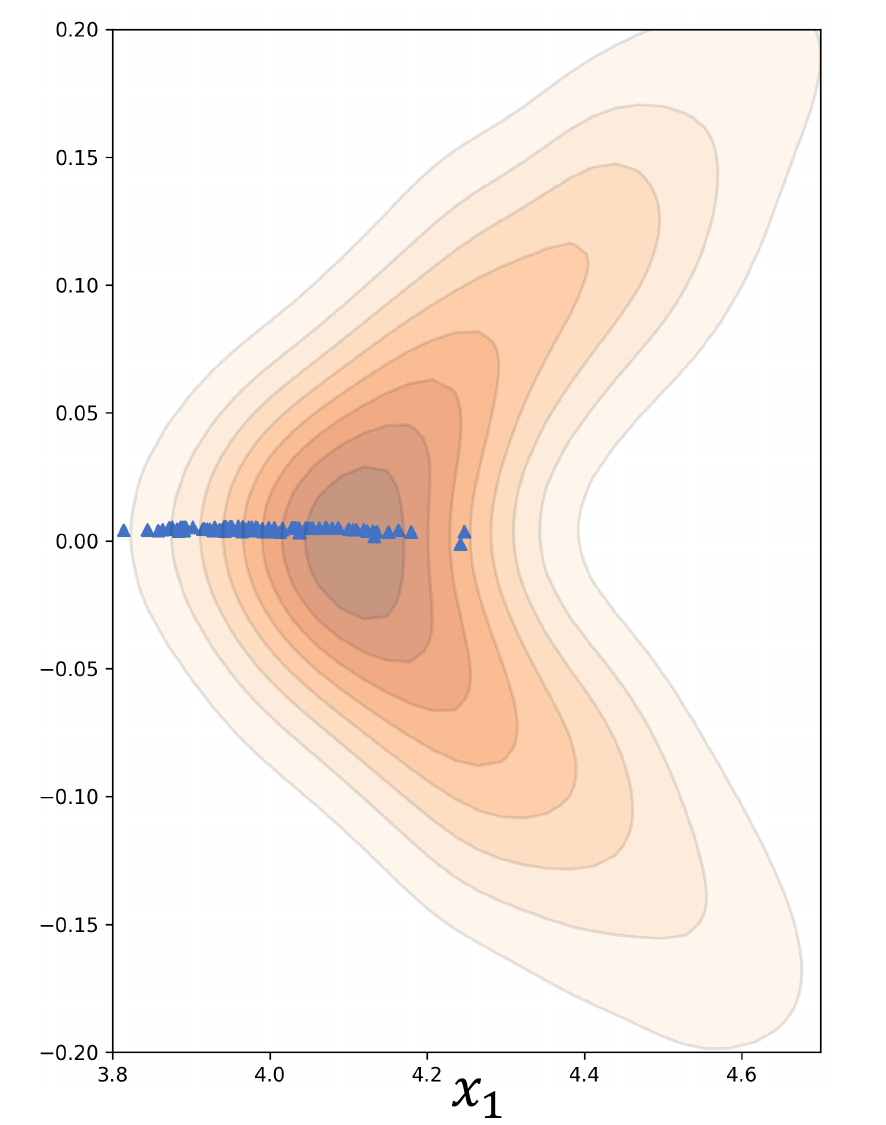}
    \caption{}%
    \vspace{-5pt}
    \label{fig:toy-jac}
  \end{subfigure}
  \caption{A toy example of \MethodName{}. \textbf{(a)}: The samples of \textcolor{gray}{class 0} and \textcolor{toyorange}{class 1}; \textbf{(b,c,d)}: The generated class-1 samples of \textbf{(b):} \textcolor{toyorange}{Plain \DDIMgen{}}, \textbf{(c):} \textcolor{red}{CFG}, and \textbf{(d):} \textcolor{toyblue}{\MethodName{}}, respectively, after applying \DDIMinv{} to the samples. \MethodName{} produces \Canoimg{}s that lie on a 1D manifold inside class 1, offering an intuitive visual summary of the core class semantics.
  }
  \label{fig:toy-all}
\end{figure}
We demonstrate the effect of \MethodName{} with a simple yet illustrative example. Figure~\ref{fig:toy-data} shows samples generated from our toy generative process and the corresponding density map. The process first generates class-specific samples on a segment $L=\{ (x_1, 0) | 4y-0.1 \le x_1 \le 4y+0.1 \}$, where $y \in \{ 0,1 \}$ denotes the class labels. It then adds class-independent noise to the points to generate the observed data. The samples from class 0 are included solely to introduce an inter-class contrast. We train a small class-conditional diffusion model on this data. After training, we perform \MethodName{} on randomly selected samples from class 1. As a baseline, we take the latent codes obtained with \DDIMinv{} and perform classifier-free guidance (CFG), steering the generation process toward regions with higher class-1 likelihood. The results are shown in Figure~\ref{fig:toy-all}. Notably, \MethodName{} pushes most samples to a 1D manifold inside class 1, whereas CFG mainly steers the samples away from class 0. This low-dimensional manifold described by \Canoimg{}s can be regarded as a summarization of class 1 information in this case. The underlying structure revealed by \Canoimg{}s corresponds to one of the true generative processes for the observed data, which is the one used in our toy model.
We refer readers to Section~\ref{sec:supp-toy-detail} in the Appendix for details of the toy experiments. 
Reliably recovering the exact generative model is interactable due to the identifiability issue \cite{identifiability}. The solution generally requires extra inductive bias in modeling data distribution, which we leave for future work.




\vspace{-5pt}
\subsection{Qualitative results}
\vspace{-5pt}
\label{sec:qualitative-cafol}
\begin{table*}[]
  \centering
  \resizebox{0.92\textwidth}{!}{
  \begin{tabular}{
    c|cccc||cccc                            
  }
  
  & Class & Original & CFG  & \MethodName{} & Class & Original & CFG  & \MethodName{} \\
  \midrule
  \multirow{3}{*}[5pt]{\rotatebox{90}{\parbox{3.64cm}{\centering \textbf{\hspace{0pt} DiT \cite{dit}}}}} & \rotatebox{90}{\parbox{1.64cm}{\centering \textbf{\hspace{0pt} Tench}}}
    & \includegraphics[width=0.15\textwidth]{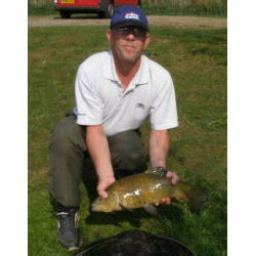}
    & \includegraphics[width=0.15\textwidth]{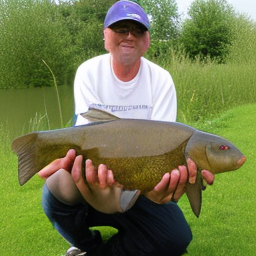}
    & \includegraphics[width=0.15\textwidth]{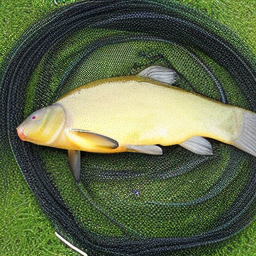}
    & \rotatebox{90}{\parbox{1.64cm}{\centering \textbf{\hspace{0pt} Tench}}}
    & \includegraphics[width=0.15\textwidth]{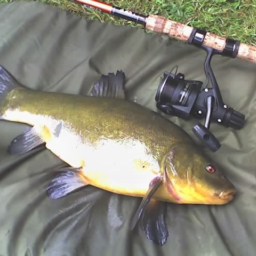}
    & \includegraphics[width=0.15\textwidth]{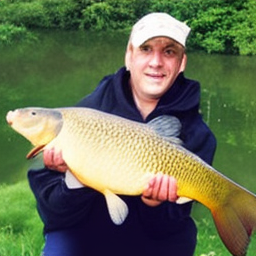}
    & \includegraphics[width=0.15\textwidth]{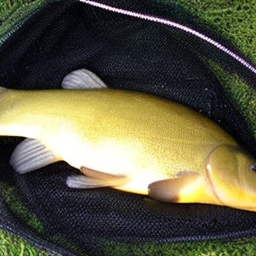} \\
  & \rotatebox{90}{\parbox{1.64cm}{\centering \textbf{\hspace{0pt} Wall Clock}}}
    & \includegraphics[width=0.15\textwidth]{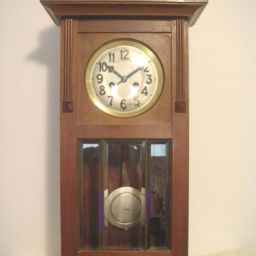}
    & \includegraphics[width=0.15\textwidth]{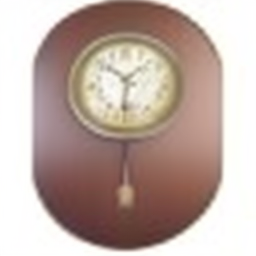}
    & \includegraphics[width=0.15\textwidth]{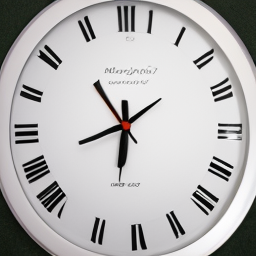}
    & \rotatebox{90}{\parbox{1.64cm}{\centering \textbf{\hspace{0pt} Wall Clock}}}
    & \includegraphics[width=0.15\textwidth]{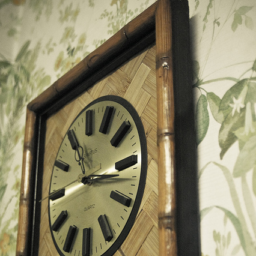}
    & \includegraphics[width=0.15\textwidth]{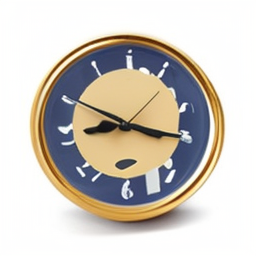}
    & \includegraphics[width=0.15\textwidth]{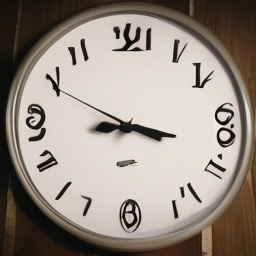} \\
  & \rotatebox{90}{\parbox{1.64cm}{\centering \textbf{\hspace{0pt} Tabby Cat}}}
    & \includegraphics[width=0.15\textwidth]{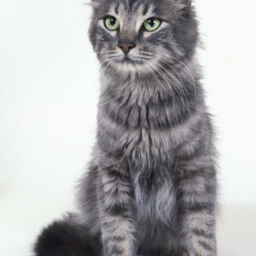}
    & \includegraphics[width=0.15\textwidth]{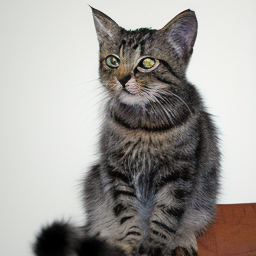}
    & \includegraphics[width=0.15\textwidth]{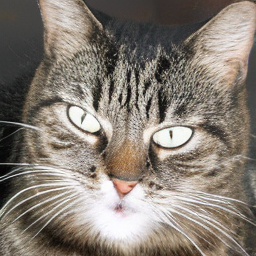}
    & \rotatebox{90}{\parbox{1.64cm}{\centering \textbf{\hspace{0pt} Wine Bottle}}}
    & \includegraphics[width=0.15\textwidth]{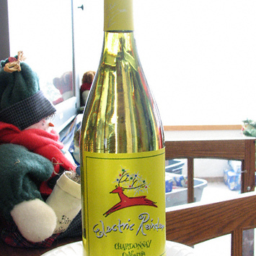}
    & \includegraphics[width=0.15\textwidth]{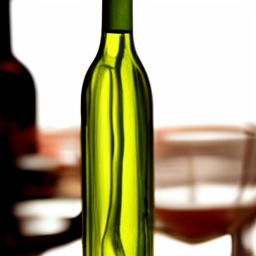}
    & \includegraphics[width=0.15\textwidth]{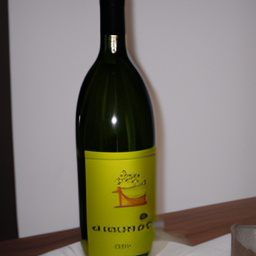} \\
    \midrule
  \multirow{3}{*}[5pt]{\rotatebox{90}{\parbox{3.64cm}{\centering \textbf{\hspace{0pt} SD \cite{ldm}}}}} & \rotatebox{90}{\parbox{1.64cm}{\centering \textbf{\hspace{0pt} Tench}}}
    & \includegraphics[width=0.15\textwidth]{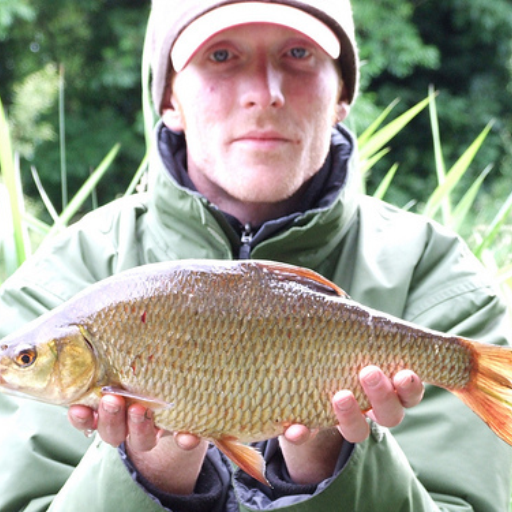}
    & \includegraphics[width=0.15\textwidth]{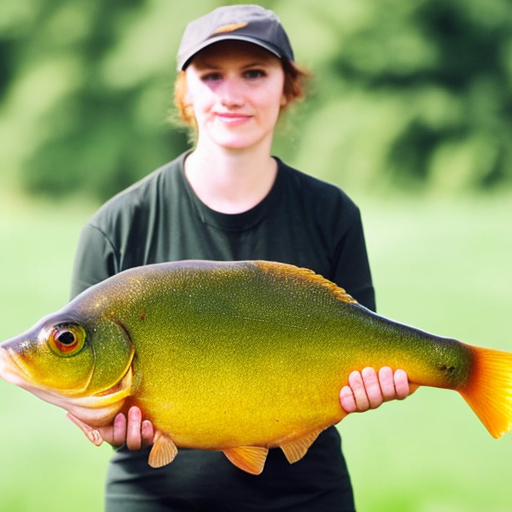}
    & \includegraphics[width=0.15\textwidth]{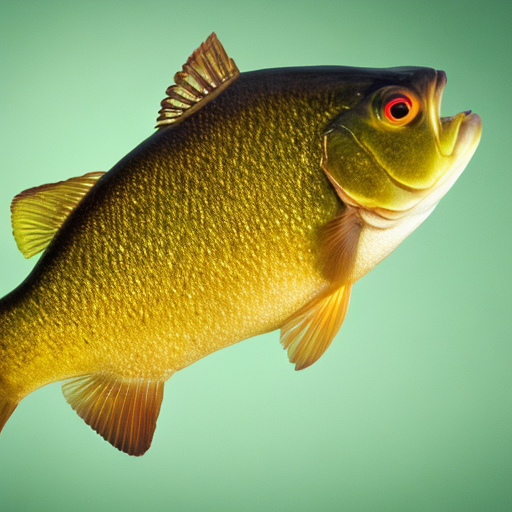}
    & \rotatebox{90}{\parbox{1.64cm}{\centering \textbf{\hspace{0pt} Tench}}}
    & \includegraphics[width=0.15\textwidth]{figures/dit/fish4_ori.png}
    & \includegraphics[width=0.15\textwidth]{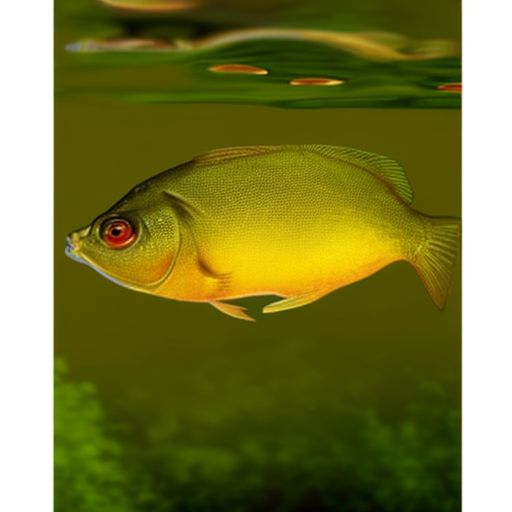}
    & \includegraphics[width=0.15\textwidth]{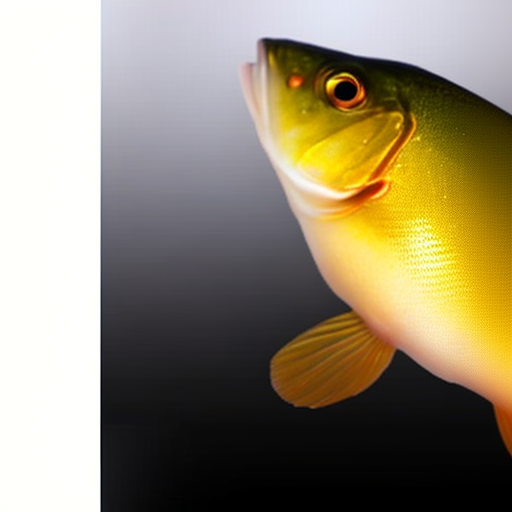} \\
  & \rotatebox{90}{\parbox{1.64cm}{\centering \textbf{\hspace{0pt} Wall Clock}}}
    & \includegraphics[width=0.15\textwidth]{figures/dit/wallclock3_ori.png}
    & \includegraphics[width=0.15\textwidth]{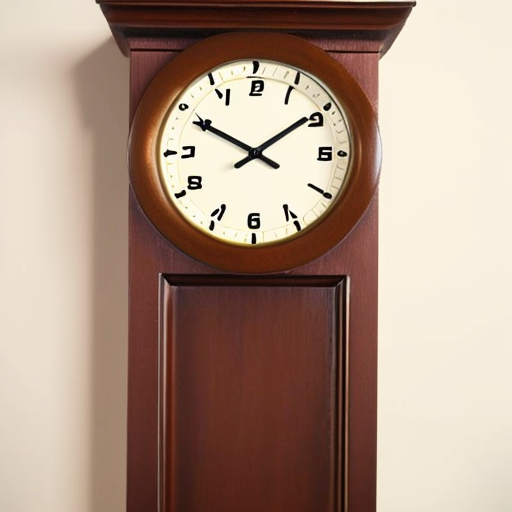}
    & \includegraphics[width=0.15\textwidth]{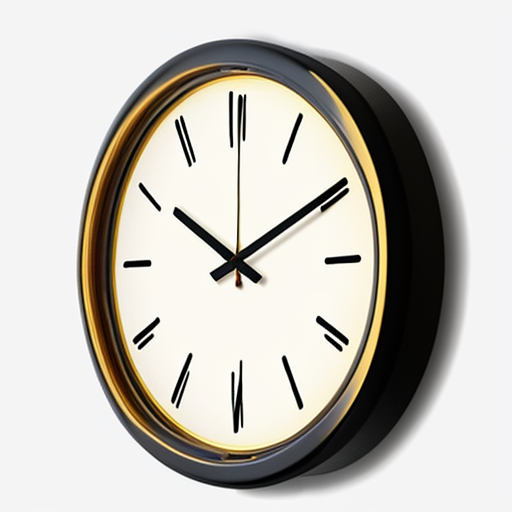}
    & \rotatebox{90}{\parbox{1.64cm}{\centering \textbf{\hspace{0pt} Wall Clock}}}
    & \includegraphics[width=0.15\textwidth]{figures/dit/wallclock2_ori.png}
    & \includegraphics[width=0.15\textwidth]{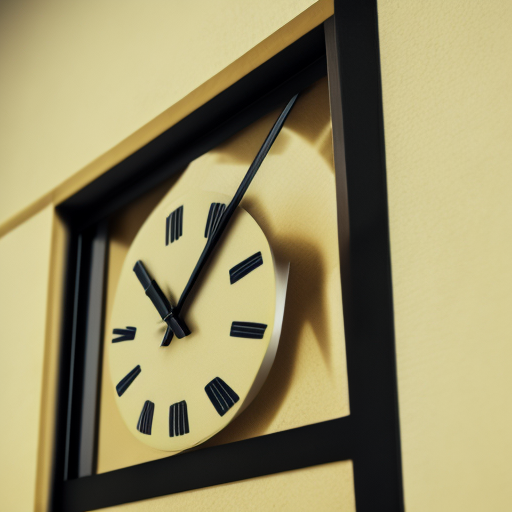}
    & \includegraphics[width=0.15\textwidth]{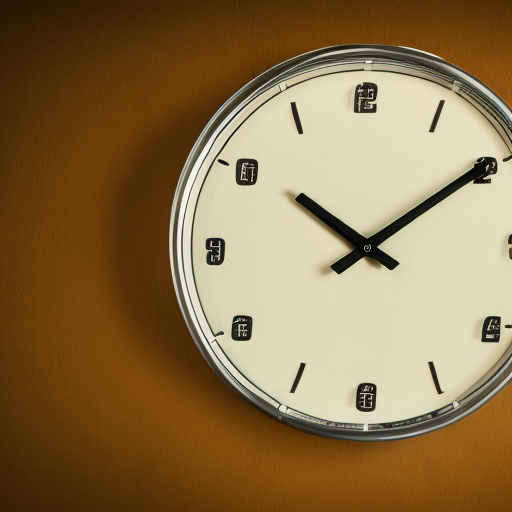} \\
  & \rotatebox{90}{\parbox{1.64cm}{\centering \textbf{\hspace{0pt} Airliner}}}
    & \includegraphics[width=0.15\textwidth]{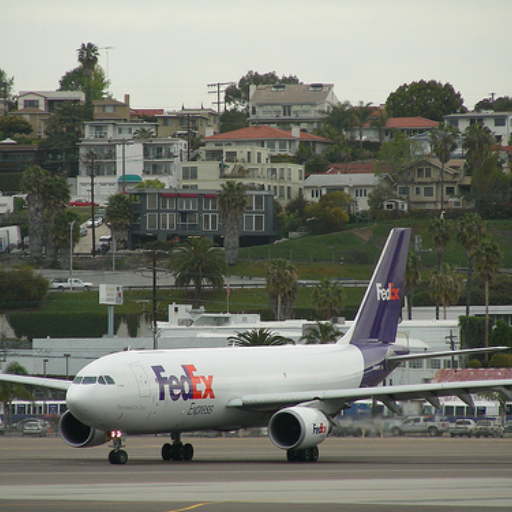}
    & \includegraphics[width=0.15\textwidth]{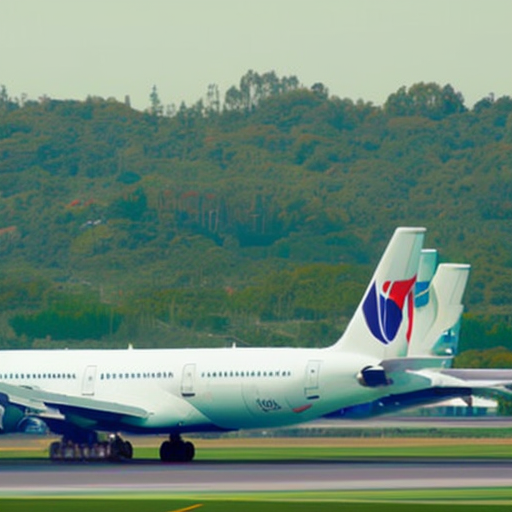}
    & \includegraphics[width=0.15\textwidth]{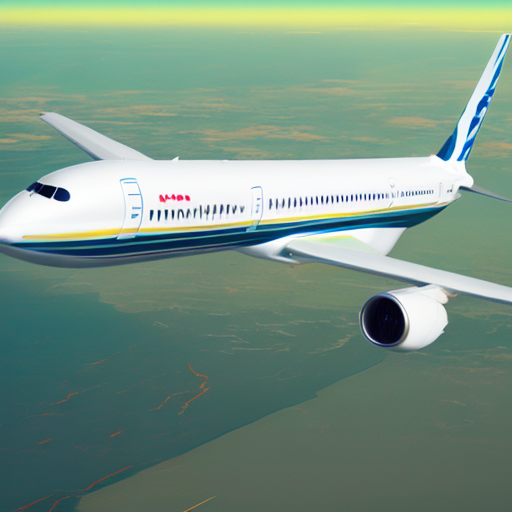}
    & \rotatebox{90}{\parbox{1.64cm}{\centering \textbf{\hspace{0pt} Wine Bottle}}}
    & \includegraphics[width=0.15\textwidth]{figures/dit/bottle_ori.png}
    & \includegraphics[width=0.15\textwidth]{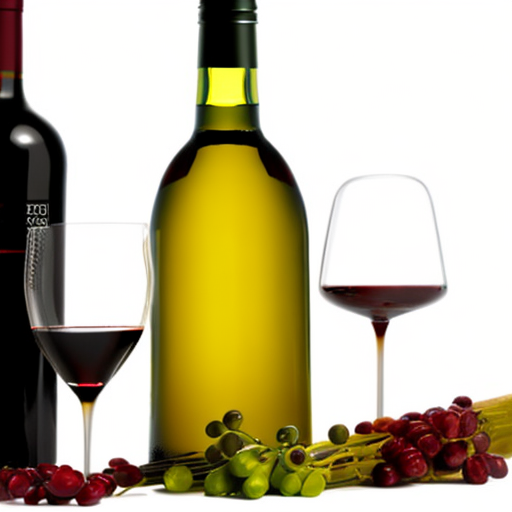}
    & \includegraphics[width=0.15\textwidth]{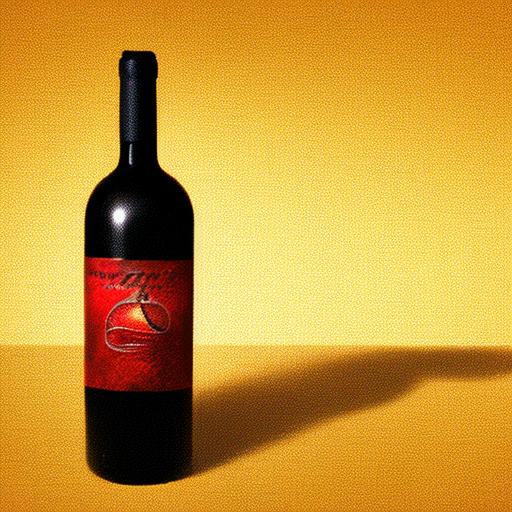} \\
  \end{tabular}
    }
  \captionof{figure}{Comparison of classifier-free guidance (CFG) and \MethodName{} on an ImageNet $256\times256$ DiT \cite{dit} and Stable Diffusion 2.1 (SD) \cite{ldm}. We use the following prompt template for SD: \textit{a photo of <Class>} \cite{diff-classifier-first}. \MethodName{} focuses on identifying \Cano{}s to preserve the core class information, yielding \Canoimg{}s that provide an interpretable summary of the essential class semantics, whereas CFG aims at finding high-likelihood images. Two \Canoimg{}s from the \textit{Tench} and \textit{Wall Clock} class are presented to show that \Cano{}s do not collapse to a single constant vector. All "Original" images are taken from ImageNet. More visual results are in Section~\ref{sec:supp-more-visual-results} in the Appendix.
  }
  \label{fig:visual-result-main}
  \vspace{-15pt}
\end{table*}

Scaling up, we demonstrate the qualitative effect of \MethodName{} for two pre-trained CDMs, a class-conditioned DiT \cite{dit} trained on the ImageNet $256\times256$ dataset and a Stable Diffusion 2.1 model \cite{ldm} trained on a subset of LAION-5B \cite{laion5b}. We use DDIM \cite{ddim} as the sampler and use 100 diffusion steps for both generation and inversion. Visual results are shown in Figure~\ref{fig:visual-result-main}. We adopt classifier-free guidance after removing \ExtraDir{}s in \MethodName{} to ensure the data fidelity, and compare the results against pure CFG. The \Cano{}s visualized as \Canoimg{}s show that our method preserves the core class information in the original images, while CFG focuses on increasing the class-conditional likelihood. We provide more visual results in Section~\ref{sec:supp-more-visual-results} in the Appendix. Occasionally, \MethodName{} can select suboptimal $t_e$ and $k$, leading to artefacts in the generated images. The failure cases are shown in Figure~\ref{fig:fail-t}, \ref{fig:fail-k} in the Appendix.

\vspace{-8pt}
\subsection{On the generalization of \MethodName{}}
\vspace{-5pt}
While we focus on class-conditional DMs to develop the \MethodName{} framework, it also extends naturally to text-conditioned models \textit{e.g.} Stable Diffusion \cite{ldm}, as shown in Figure~\ref{fig:visual-result-main} and Section~\ref{sec:supp-cafol-generalization} in the Appendix. Text prompts span a far richer semantic space than one-hot class labels, giving finer control over where the \Cano{}s lie.
We show examples of the effect using different prompts on the same input in Section~\ref{sec:supp-finegrained-text-control} in the Appendix. Understanding how this semantic structure shapes the located \Cano{}s is an interesting direction to explore. \MethodName{} is also compatible with different DM samplers, as shown in Section~\ref{sec:supp-edm-cafol} in the Appendix.

\vspace{-8pt}
\section{The application of \Cano{}s}
\label{sec:canodistill-aug}
\vspace{-5pt}
\begin{figure*}
    \centering
    \includegraphics[width=\linewidth, trim={90 60 90 50},clip]{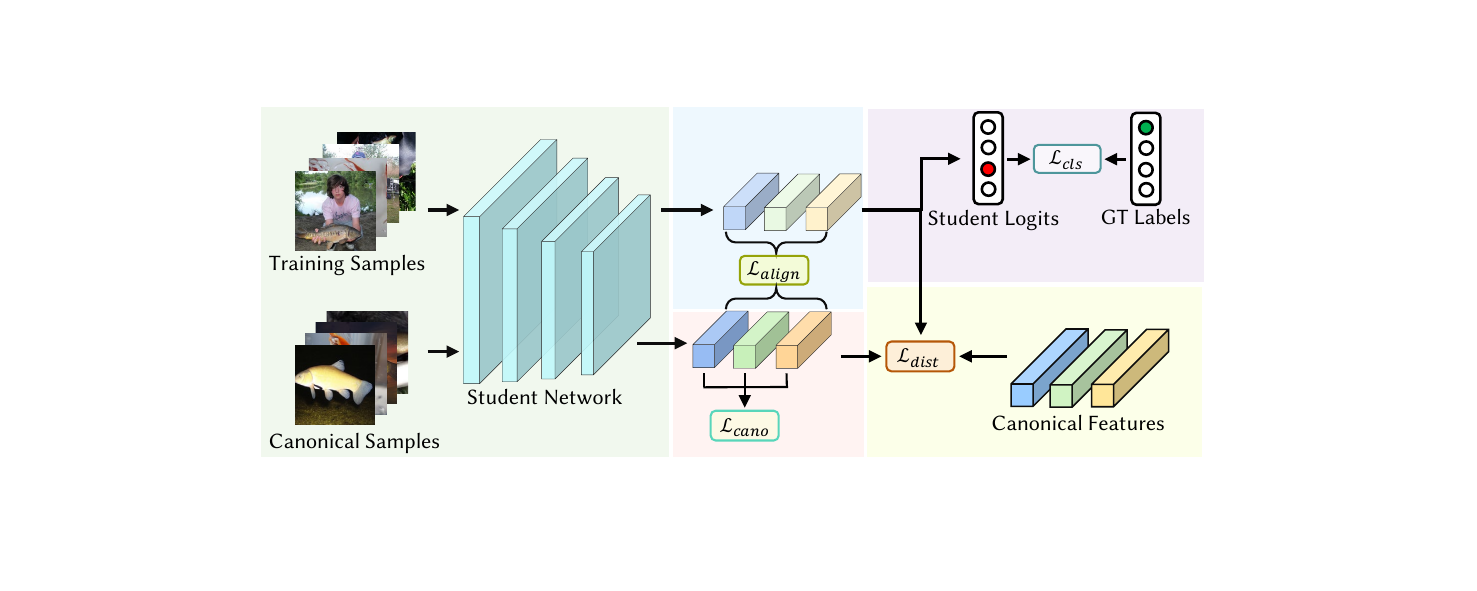}
    \caption{Overview of \textit{\textbf{\CanoDistill{}}}. We align the student's features of training images to those of \Canoimg{}s using $\mathcal{L}_{align}$. The student is trained to discriminate between \Canoimg{}s in different classes by optimizing $\mathcal{L}_{cano}$. The CDM, as a teacher, provides \Canofeat{}s for feature distillation with $\mathcal{L}_{dist}$. Finally, the student is supervised on the ground-truth labels via $\mathcal{L}_{cls}$. }
    \label{fig:cano-distill-aug}
    \vspace{-10pt}
\end{figure*}
As demonstrated in Section~\ref{sec:cafol}, \Cano{}s correspond to the essential class information learned by the CDMs. Building on this insight, we design a feature distillation framework for \Cano{}s, termed \textit{\textbf{\CanoDistill{}}}, as illustrated in Figure~\ref{fig:cano-distill-aug}. Given a training batch $B=\{(\bm{x}_{1}, \bm{c}_{1}), (\bm{x}_{2}, \bm{c}_{2}), ..., (\bm{x}_{b}, \bm{c}_{b}) \}$ where $\bm{x}_i$ is an image and $\bm{c}$ is the class label, we first \textbf{randomly} select a \Cano{} $\tilde{\bm{x}}_i$ corresponding to the category of each sample. The random sampling relaxes the constraint of the size equivalence between the training dataset and the set of \Cano{}s. Let $P_i=\{j \in \{1,2,...,b \} \mid \bm{c}_j=\bm{c}_i \}$ be the set of indices of samples in the batch with class label $\bm{c}_i$. We then compute student features $\bm{z}_i = g_\phi(\bm{x}_i) \in \mathbb{R}^d, \tilde{\bm{z}}_i = g_\phi(\tilde{\bm{x}}_i) \in \mathbb{R}^d$, where $g$ denotes the student network and $\phi$ is the set of its training parameters.
The first component in \textbf{\textit{\CanoDistill{}}}, $\mathcal{L}_{align}$, is designed to pull each training image feature towards all same-class \Canoimg{}s, and push away the ones from other classes. It is given in Eq.~\ref{eq:alignment-canodistill}.
\vspace{-5pt}
\begin{equation}
    \mathcal{L}_{align}=-\frac{1}{b}
        \sum_{i=1}^b
        \frac{1}{|P_i|}
        \sum_{j\in P_i}
        \log
        \frac{
          \exp\!\bigl(\bm{z}_i \cdot \tilde{\bm{z}}_j / \tau\bigr)
        }{
          \sum_{k=1}^B \exp\!\bigl(\bm{z}_i \cdot \tilde{\bm{z}}_k  / \tau\bigr)
        }\,.
    \label{eq:alignment-canodistill}
\end{equation}
$\tau$ is a temperature hyperparameter. 
Second, we encourage all \Canoimg{}s of the same class to cluster in the student's feature space, and separate those of different classes. It is achieved by minimizing $\mathcal{L}_{cano}$ in Eq.~\ref{eq:cano-canodistill}. When a \Canoimg{} has no same-class positives, we simply optimize the denominator. See Section~\ref{sec:supp-lcano-ce-nopos-discussion} in the Appendix for more details. 
\begin{equation}
    \mathcal{L}_{cano}=-\frac{1}{b}
        \sum_{i=1}^b
        \frac{1}{|P_i| - 1}
        \sum_{j\in P_i, j\neq i}
        \log
        \frac{
          \exp\!\bigl(\tilde{\bm{z}}_i \cdot \tilde{\bm{z}}_j / \tau\bigr)
        }{
          \sum_{k \ne i} \exp\!\bigl(\tilde{\bm{z}}_i \cdot \tilde{\bm{z}}_k  / \tau\bigr)
        }\,.
    \label{eq:cano-canodistill}
\end{equation}
We perform an ablation study on the design of $\mathcal{L}_{cano}$ by replacing it with a classification loss on the \Canoimg{}s in Section~\ref{sec:supp-lcano-ce-nopos-discussion} in the Appendix, demonstrating the advantage of the current $\mathcal{L}_{cano}$.
For feature distillation, we transfer the structures of \Canofeat{}s to the student via minimizing $\mathcal{L}_{dist}$. Specifically, we use the Centered Kernel Alignment (CKA) \cite{cka} metric to align the student’s representations of both the training images and the \Canoimg{}s with the \Canofeat{}s extracted from the CDM. Maximizing CKA, \textit{i.e.} minimizing $\mathcal{L}_{dist}$, aligns the linear subspace spanned by the student’s feature vectors with that of the teacher, effectively distilling the teacher’s class-discriminative structure into the student. Denote the student feature matrices of training images and \Canoimg{}s as $\bm{Z} \in \mathbb{R}^{b \times d}$ and $\tilde{\bm{Z}} \in \mathbb{R}^{b \times d}$, respectively, and the \Canofeat{} matrix as $\mathcal{A} \in \mathbb{R}^{b \times d^\prime}$. The \Canofeat{}s are extracted from a frozen CDM for all $\tilde{x}_i$ in a training batch.
\vspace{-3pt}
\begin{equation}
    \mathcal{L}_{dist}= \lambda_{cka} \log (1- \text{CKA}(\bm{Z}, \mathcal{A})) + (1-\lambda_{cka} ) \log (1- \text{CKA}(\tilde{\bm{Z}}, \mathcal{A})).
    \label{eq:dist-canodistill}
\end{equation}
We find that using CKA for structural alignment outperforms existing diffusion-based feature distillation methods \cite{dreamteacher,rkd, fitnet-hint,dm-as-representation-learner} and refer the readers to Section~\ref{sec:supp-ldist-hint-rkd-cka} in the Appendix for more details.
Finally, the student is trained with the standard cross-entropy loss $\mathcal{L}_{cls}$ on ground-truth labels.
The final loss function for \textbf{\textit{\CanoDistill{}}} is given in Eq.~\ref{eq:canodist-final}. 
\begin{equation}
    \mathcal{L}_{CaDistill} = \mathcal{L}_{cls} + \lambda_{cs} ( \lambda_{cf} \mathcal{L}_{align} + (1-\lambda_{cf}) \mathcal{L}_{cano} ) + \lambda_{dist} \mathcal{L}_{dist}
    \label{eq:canodist-final}
\end{equation}

\vspace{-10pt}
\subsection{\Cano{}s in practice}
\vspace{-5pt}
\label{sec:cano-quant-result}

\begin{table*}[t]
\centering
\caption{Quantitative comparisons of \textbf{\textit{\CanoDistill{}}} and baselines on
CIFAR-10 \cite{cifar10} (ResNet-18) and ImageNet \cite{deng2009imagenet} (ResNet-50). $\mathcal{D}$: Dataset.
Adversarial robustness benchmarks: PGD \cite{pgd}, CW \cite{cw}, APGD-DLR / APGD-CE \cite{autoattack};
Evaluations of generalization ability : Corruption (CIFAR10-C and ImageNet-C) \cite{imagenet-c},
ImageNet-A \cite{imagenet-a}, ImageNet-ReaL \cite{im-real}.  $\text{Data}_{\text{DM}}$ is the portion of data for which the DM acts as teacher.  Higher is better.
Values lower than the vanilla model are in \textcolor{venetianred}{red}. $^\dagger$: the model relies on unconditional DMs, or the training cannot be performed, see Section~\ref{sec:supp-repfusion-detail} in the Appendix.}
\label{tab:main-quant}
\resizebox{\linewidth}{!}{
\begin{tabular}{l l c | c c c c c | c c c}
\toprule
$\mathcal{D}$ & Model & $\text{Data}_{\text{DM}}$ & Clean & PGD & CW & APGD-DLR & APGD-CE & Corruption & IM-A & IM-ReaL\\
\midrule
 \multirow{6}{*}[-5pt]{\rotatebox{90}{\parbox{2cm}{\centering \textbf{\hspace{0pt} CIFAR10}}}} & Vanilla & /        & 92.4 & 33.4 & 20.9 & 34.2 & 32.0 & 76.1 & -- & --\\
& SupCon  \cite{supcon}                   & /      & 92.7 & \textcolor{venetianred}{29.1} & \textcolor{venetianred}{16.8} & 34.8 & \textcolor{venetianred}{29.9} & 76.9 & -- & --\\
\cmidrule(lr){2-11} 
& RepFusion$^\dagger$ \cite{dm-as-representation-learner}       & 100\%  & 92.7 & \textcolor{venetianred}{30.3} & \textcolor{venetianred}{17.2} & \textcolor{venetianred}{32.1} & \textcolor{venetianred}{29.2} & \textcolor{venetianred}{75.3} & -- & --\\
& \DMFit{}                  & 100\%  & 92.9 & 41.3 & 32.8 & 38.7 & 36.0 & 76.7 & -- & --\\
\cmidrule(lr){2-11} 
& \DMDistill{}              & 10\%   & 92.9 & 43.7 & 37.3 & 40.9 & 39.0 & 76.6 & -- & --\\
\rowcolor{gray!20} \cellcolor{white} & \textit{\textbf{\CanoDistill{}}} & 10\% & \textbf{93.1} & \textbf{47.9} & \textbf{43.1} & \textbf{44.1} & \textbf{43.3} & \textbf{77.7} & -- & --\\
\midrule
\midrule
\multirow{5}{*}[-5pt]{\rotatebox{90}{\parbox{2cm}{\centering \textbf{\hspace{0pt} ImageNet}}}} & Vanilla & /        & 75.9 & 15.6 & 13.7 & 17.2 & 16.7 & 45.9 & 6.3 & 82.8\\
\cmidrule(lr){2-11} 
& DiffAug \cite{diffaug}                & 100\% & \textbf{76.0} & 15.9 & \textcolor{venetianred}{13.1} & 17.2 & 17.0 & \textbf{47.2} & \textcolor{venetianred}{4.8} & 83.1\\
& \DMFit{}                & 100\% & \textcolor{venetianred}{75.7} & 15.7 & 14.1 & \textcolor{venetianred}{17.0} & 16.7 & \textcolor{venetianred}{43.6} & \textcolor{venetianred}{5.0} & 82.8\\
\cmidrule(lr){2-11} 
& \DMDistill{}            & 10\%  & \textcolor{venetianred}{75.7} & 20.8 & 20.3 & 20.8 & 21.4 & \textcolor{venetianred}{45.6} & \textcolor{venetianred}{6.0} & \textcolor{venetianred}{82.7}\\
\rowcolor{gray!20} \cellcolor{white} & \textit{\textbf{\CanoDistill{}}} & 10\% & 75.9 & \textbf{21.9} & \textbf{21.7} & \textbf{22.5} & \textbf{22.3} & 46.1 & \textbf{6.7} & \textbf{83.1}\\
\bottomrule
\end{tabular}
}
\vspace{-5pt}
\end{table*}

We conduct experiments of \textbf{\textit{\CanoDistill{}}} on CIFAR10 \cite{cifar10} using a pre-trained UNet-based CDM \cite{ddpm,dm-discriminative3-ddae} and on ImageNet using the ImageNet $256\times256$-trained Diffusion Transformer (DiT) \cite{dit}. We choose two different DM architectures to demonstrate the applicability of our method. The CDM on CIFAR10 does not have an unconditional branch. The DMs are trained on the same dataset as the student models, hence no extra data is considered \cite{diffaug}. We follow this principle to choose the baselines, comparing our method with the SOTA diffusion-based feature distillation \cite{dm-as-representation-learner} and data augmentation \cite{diffaug} methods. In addition, we design two important baselines. 
The first one, \DMFit{}, is to distill the raw space structure in CDMs using $\mathcal{L}_{dist}$ on all training samples, which represents the current mainstream idea of using DMs as teachers in feature distillation. It outperforms existing diffusion-based feature distillation losses \cite{dreamteacher,rkd,fitnet-hint,dm-as-representation-learner} (Section~\ref{sec:supp-ldist-hint-rkd-cka} in the Appendix).
Second, for the \DMDistill{} baseline, we train the student model using the same framework of \textbf{\textit{\CanoDistill{}}}, except that the images and features are obtained using CFG. For the CDM on CIFAR10, we sample new images for \DMDistill{} as this CDM lacks the unconditional branch and cannot perform CFG. We fix $t_r$, \textit{i.e.} the feature extraction time step, for all methods that do not adaptively change it \cite{dm-as-representation-learner}. For the student network, we use ResNet18 \cite{resnet} on CIFAR10 and ResNet50 \cite{resnet} on ImageNet, two well-established convolutional neural network baselines \cite{dm-as-representation-learner}. We conduct PGD \cite{pgd}, CW \cite{cw}, APGD-DLR \cite{autoattack}, and APGD-CE \cite{autoattack} attacks to test the adversarial robustness of the student. On CIFAR10, we report the Top-1 accuracy on the CIFAR10-C dataset \cite{imagenet-c} as the metric for evaluating generalization. On ImageNet, we show the Top-1 accuracy on ImageNet-C (IM-C) \cite{imagenet-c}, ImageNet-A (IM-A) \cite{imagenet-a}, and ImageNet-ReaL (IM-ReaL) \cite{im-real}. We refer readers to Section~\ref{sec:supp-training-detail} in the Appendix for more training and evaluation procedure details.
\begin{wraptable}{r}{0.5\linewidth}
    \centering
    \caption{Results on the Backgrounds Challenge \cite{inbg-challenge}. Higher is better. See Section~\ref{sec:supp-inbg} in the Appendix for details. }
    \label{tab:imagenet-inbg}
     \resizebox{\linewidth}{!}{
    \begin{tabular}{c|cc|cc}
       Model & Original & BG-Same & BG-Rand & Only-FG \\
       \midrule
        Vanilla & 96.0 & 88.0 & 81.1 & 87.6 \\
        \midrule
        \midrule
        DiffAug \cite{diffaug} & 96.1 & 87.5 & 80.3 & 87.4 \\
        \DMFit{} & \textbf{96.3} & 88.0 & 80.6 & 84.5 \\
        \midrule
        \DMDistill{} & \textbf{96.3} & \textbf{89.0} & 82.6 & 87.8 \\
        \rowcolor{gray!20} \textit{\textbf{\CanoDistill{}}} & \textbf{96.3} & \textbf{89.0} & \textbf{83.6} & \textbf{88.5} \\
    \end{tabular}
    }
\end{wraptable}
\hspace{-4.5pt}The results are shown in Table~\ref{tab:main-quant}. \textbf{\textit{\CanoDistill{}}} consistently improves the adversarial robustness and generalization ability of the student, while the main effect of SOTA methods is improving the clean accuracy or a single aspect of the robustness. A more detailed discussion of the results, including the difference between CIFAR10 and ImageNet, is in Section~\ref{sec:discussion-quant-result} in the Appendix. Moreover, we consider an interesting and challenging baseline that is capable of extracting class semantics, \textit{i.e.,} using the class token in an ImageNet-pretrained ViT \cite{deit, deit3} as the teacher. We show the results in Section~\ref{sec:supp-vit-cls-token-compare} in the Appendix. \textbf{\textit{\CanoDistill{}}} is effective when the student has transformer-based vision backbones, as shown in Section~\ref{sec:generalization-swin} in the Appendix. 

The student, trained with \textbf{\textit{\CanoDistill{}}}, focuses more on the core class signal. We demonstrate this by a test on the Backgrounds Challenge \cite{inbg-challenge}, where the background of an image that is irrelevant to the class identity is either removed or shuffled. We provide visualization of the images in Section~\ref{sec:supp-inbg} in the Appendix. The results are given in Table~\ref{tab:imagenet-inbg}. \textbf{\textit{\CanoDistill{}}} improves the student performance on BG-Rand and Only-FG while maintaining the accuracy on the Original and BG-Same splits, indicating a mitigation in the model’s dependence on the spurious background cues for classification. A more detailed discussion of the results is in Section~\ref{sec:supp-inbg} in the Appendix.



\textbf{Ablation studies.} Section~\ref{sec:supp-ablation} include the ablation analysis on \textbf{\textit{\CanoDistill{}}}, including: 
\begin{itemize}[leftmargin=*]
    \item \textbf{Number of \Cano{}s}. We demonstrate that 10\% of \Cano{}s is sufficient for achieving competitive performance, implying the low-dimensionality property of the class manifolds in CDMs.
    \item \textbf{Necessity of $\mathcal{L}_{align}$, $\mathcal{L}_{cano}$, $\mathcal{L}_{dist}$, and their balance.} We empirically conclude the effects of all loss functions and choose the optimal weighting schemes, including $\lambda_{cs},\lambda_{cf},\lambda_{dist},\lambda_{cka}$.
    \item \textbf{CFG magnitude.} We perform CFG after obtaining \Cano{}s, and show a proper choice of its magnitude. Importantly, a higher CFG magnitude does not contribute to better performance, indicating that \textbf{\textit{\CanoDistill{}}} is not merely providing a converging prior on the student features. We also provide a discussion on this topic in Section~\ref{sec:supp-limit-nmi-not-converging} in the Appendix.
\end{itemize}

\section{Limitation}
\label{sec:limitation}
\MethodName{} has certain limitations. It can occasionally select a suboptimal projection time step, $t_e$, or the total number of \ExtraDir{}s considered, $n$, as discussed in Section~\ref{sec:qualitative-cafol} and Section~\ref{sec:supp-te-k} in the Appendix.
%
%
Calculating the singular vectors of the Jacobian of CDMs is computationally intensive. 
Whether the application of \Cano{}s can be effective on larger-scale problems, \textit{e.g.,} ImageNet22K, remains a question. 




\section{Conclusion}
We introduce \FullMethodNameBold{} (\MethodName{}), a principled method to uncover the core categorical information encoded in pre‑trained conditional diffusion models (CDMs). 
By removing \ExtraDir{}s from a sample’s latent code, \MethodName{} produces \textit{\Cano{}}, whose internal representation—\Canofeat{}—distills the class-defining semantics of each category
%
Decoding \Cano{}s yields \Canoimg{}s that offer an interpretable and compact summary of the class. Quantitatively, \Canofeat{}s form more compact and easily separable clusters in CDM feature space than the original inputs.
%
%
Building on \Cano{}s, we have proposed \textbf{\textit{\CanoDistill{}}}, a diffusion-based feature distillation framework. The teacher CDM transfers core class semantics to the student only via \Cano{}s, which is equivalent to merely 10\% of the original data, while the student is being trained on the full training set. The student achieves strong adversarial robustness and generalization on CIFAR‑10 and ImageNet, focusing more on the true class signal instead of spurious background cues than the original model. Together, \MethodName{} and \textbf{\textit{\CanoDistill{}}} demonstrate that CDMs can be transformed from black‑box generators into compact, interpretable teachers for robust representation learning.

\clearpage
\bibliographystyle{plainnat}
\bibliography{neurips_2025}

\clearpage

\appendix
\section*{Appendix: Table of Contents}
\startcontents[sections]
\printcontents[sections]{}{1}{}
\clearpage

\section{An overview of the paper}
\begin{figure*}[ht!]
    \centering
    \includegraphics[width=\linewidth, trim={240 200 210 200},clip]{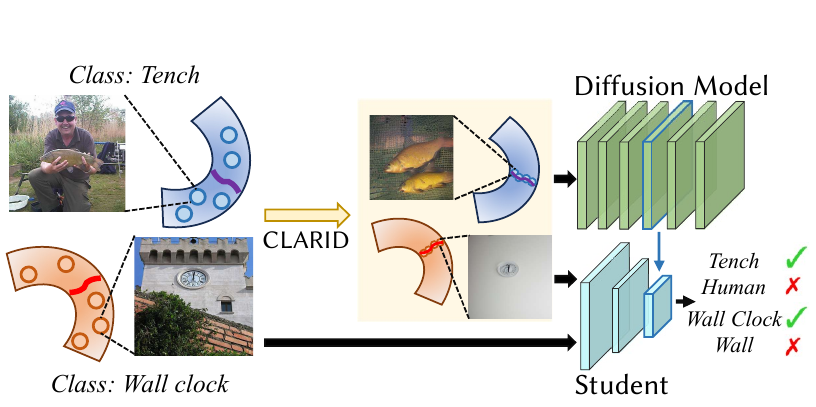}
    \caption{An overview of our paper. We identify the \textit{\Cano{}s} via \FullMethodNameBold{} (\MethodName{}) inside conditional diffusion models (CDMs) as a compact family of latent codes that contain the core class information with minimal class-irrelevant signals. These prototypes lie on low-dimensional manifolds ( red \textcolor{myred}{\large$\boldsymbol{\sim}$} and purple \textcolor{mypurple}{\large$\boldsymbol{\sim}$} tildes) within the latent space of CDMs. 
    Leveraging the \Cano{}s, we design a diffusion-based \textcolor{myblue}{feature distillation} paradigm, improving the adversarial robustness and generalization of downstream models in image classification.}
    \label{fig:supp-teaser}
\end{figure*}

We outline the paper in Figure~\ref{fig:supp-teaser}. 

\section{Detailed Preliminaries}
\label{sec:supp-preliminary}
\subsection{Denoising diffusion probabilistic model (DDPM)}
DDPM \cite{ddpm} models the generation process as an inversion of a fixed forward Gaussian diffusion $q(\bm{x}_{1:T}\mid \bm{x}_0)
:=\prod_{t=1}^T q(\bm{x}_t\mid \bm{x}_{t-1})$. The forward kernel $q(\bm{x}_t\mid \bm{x}_{t-1})$ is described in Eq.~\ref{eq:ddpm}.
\begin{equation}
q(\bm{x}_t\mid \bm{x}_{t-1})
= \mathcal{N}\bigl(\bm{x}_t;\,\sqrt{1-\beta_t}\,\bm{x}_{t-1},\,\beta_t \bm{I}\bigr)
= 
:=\mathcal{N}\Bigl(\sqrt{\tfrac{\alpha_t}{\alpha_{t-1}}}\,\bm{x}_{t-1},\,(1-\tfrac{\alpha_t}{\alpha_{t-1}})\bm{I}\Bigr),
\label{eq:ddpm}
\end{equation}
where $\{\beta_t\}^{T}_{t=1}$ is the variance schedule, $\alpha_t=\prod^t_{k=1} (1-\beta_k)$.
The inversion process is defined as $p_{\theta}(\bm{x}_{0:T}):=p(\bm{x}_{T})\prod_{t=1}^{T}p_{\theta}(\bm{x}_{t-1}\mid \bm{x}_{t})$, where $\bm{x}_{T}\sim\mathcal{N}(\bm{0},\bm{I})$. $\theta$ denotes the parameter set of a trainable noise predictor $\bm{f}_\theta$. A single reverse step is formalized in Eq.~\ref{eq:ddpminv}.
\begin{equation}
\bm{x}_{t-1}
=\frac{1}{\sqrt{1-\beta_{t}}}
\Bigl(
\bm{x}_{t}
-\frac{\beta_{t}}{\sqrt{\alpha_{t}}}\,\bm{f}_{\theta,t}(\bm{x}_{t})
\Bigr)
+\sqrt{\beta_t}\,\bm{\epsilon}_{t},
\quad
\bm{\epsilon}_{t}\sim\mathcal{N}(\bm{0},\bm{I}).
\label{eq:ddpminv}
\end{equation}
$\bm{f}_{\theta,t}$ means that the noise predictor receives $t$ as a conditional input.

\subsection{Denoising diffusion implicit models (DDIM)}
DDIM \cite{ddim} proposes a non-Markovian forward diffusion process, implying the parametrization described in Eq.~\ref{eq:ddim}.
\begin{equation}
    q_{\xi}(\bm{x}_{t-1}\mid \bm{x}_t, \bm{x}_0)
    = \mathcal{N}\Bigl(
    \sqrt{\alpha_{t-1}}\,\bm{x}_0 
    + \sqrt{\,1-\alpha_{t-1}-\xi_t^2\,}\,
    \frac{\bm{x}_t - \sqrt{\alpha_t}\,\bm{x}_0}{\sqrt{1-\alpha_t}}
    ,\;
    \xi_t^2\,\bm{I}
    \Bigr),
    \label{eq:ddim}
\end{equation}
where $\xi_t = \eta \sqrt{\frac{1 - \alpha_{t-1}}{\,1 - \alpha_t\,}} \;\sqrt{1 - \frac{\alpha_t}{\alpha_{t-1}}}$. The reverse step is described in Eq.~\ref{eq:ddiminv}.
\begin{equation}
    \bm{x}_{t-1}
    =
    \Bigl(
    \frac{\bm{x}_t - \sqrt{1-\alpha_t}\,\bm{f}_\theta(\bm{x}_t)}{\sqrt{\alpha_t}}
    \Bigr)
    \;+\;\sqrt{\,1-\alpha_{t-1}-\xi_t^2\,}\,\bm{f}_\theta(\bm{x}_t)
    \;+\;
    \xi_t\,\bm{\epsilon}_t.
    \label{eq:ddiminv}
\end{equation}
DDIM and other parametrizations of the diffusion process \cite{edm,edm2} can perform inversion on the input sample, retaining certain semantic information of it at $t=T$ \cite{goldennoise}.

\section{Random samples from \textit{Tench} class generated by DiT}
\label{sec:supp-tench-exp-supp}
\begin{figure}[]
    \centering
    \includegraphics[width=\linewidth, trim={80 80 80 80},clip]{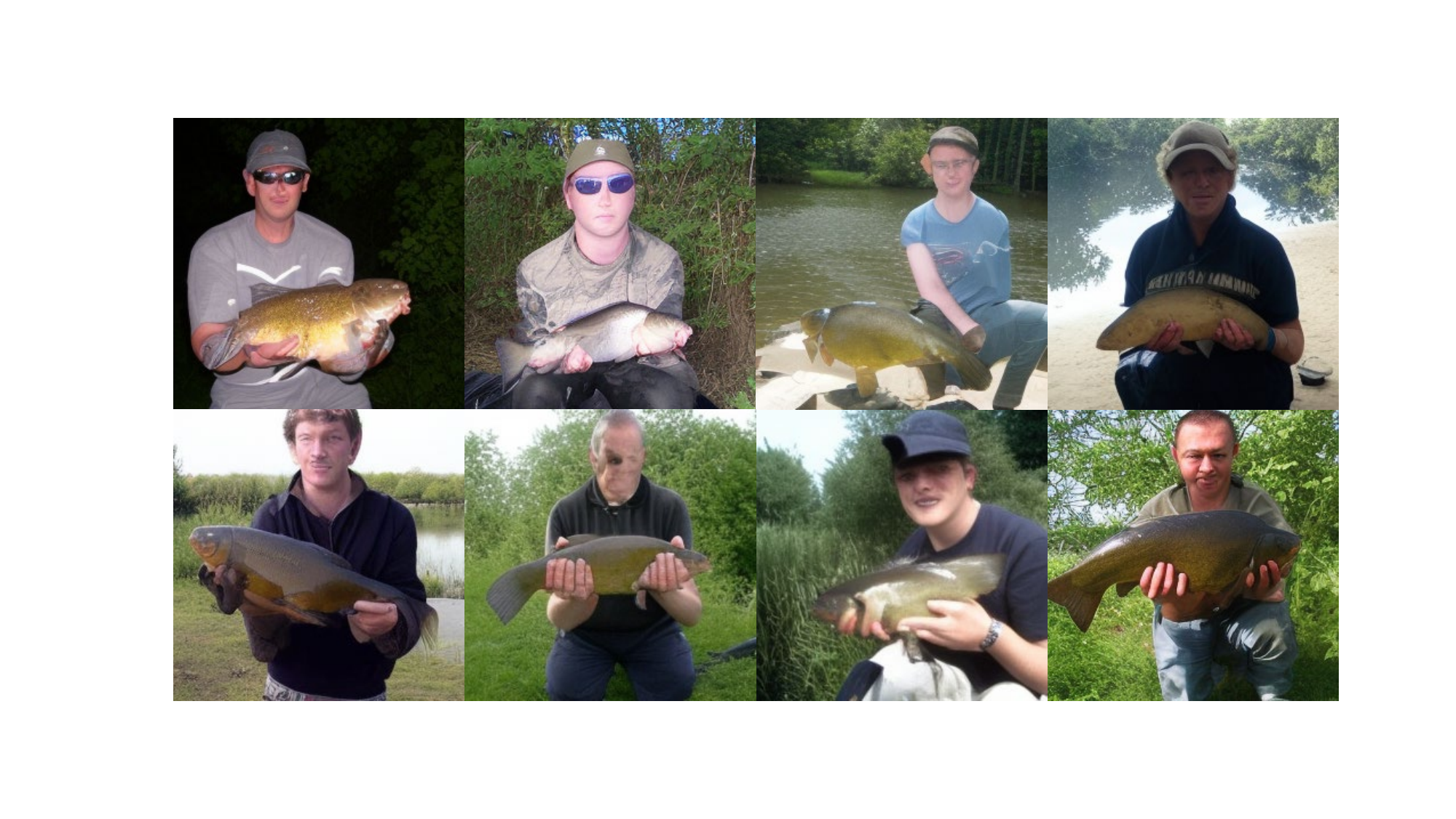}
    \caption{Random samples generated by DiT $256 \times 256$ \cite{dit} when conditioned on the \textit{Tench} class (0th class in ImageNet).}
    \label{fig:tench-exp-supp}
\end{figure}

We sample random images using DiT-XL $256 \times 256$ \cite{dit} from the \textit{Tench} class in ImageNet. We use a classifier-free guidance scale of 1.5, which is the one used in the original paper that achieves the best generation quality. The results are shown in Figure~\ref{fig:tench-exp-supp}. Most images that we observe contain an angler. 

\section{Class identity preservation when moving along \ExtraDir{}s}
\label{sec:supp-extra-dir-class-preserve}
\begin{table}[]
    \centering
    \begin{tabular}{c|ccc}
       Original  & $E_1$ & $E_2$ & $E_3$ \\
       \midrule
        \includegraphics[width=0.2\textwidth]{figures/dit/cat1_ori.png} & \includegraphics[width=0.2\textwidth]{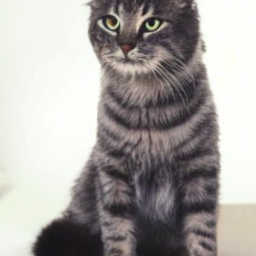} & \includegraphics[width=0.2\textwidth]{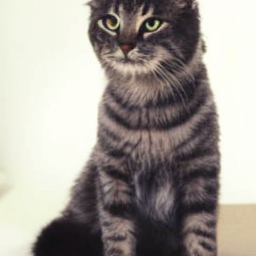} & \includegraphics[width=0.2\textwidth]{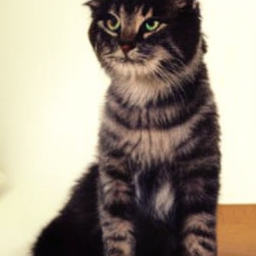} \\
        \includegraphics[width=0.2\textwidth]{figures/dit/wallclock3_ori.png} & \includegraphics[width=0.2\textwidth]{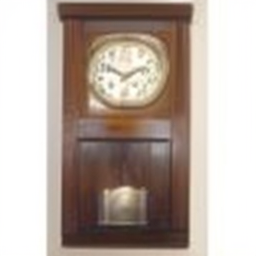} & \includegraphics[width=0.2\textwidth]{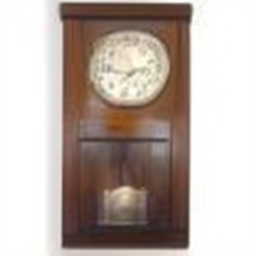} & \includegraphics[width=0.2\textwidth]{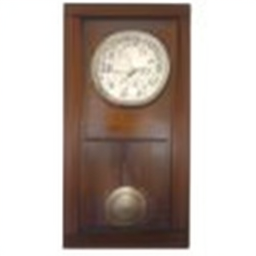} \\
        \includegraphics[width=0.2\textwidth]{figures/dit/fish2_ori.png} & \includegraphics[width=0.2\textwidth]{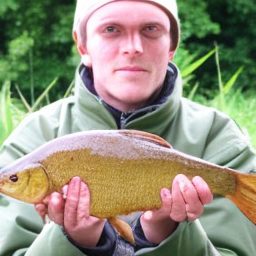} & \includegraphics[width=0.2\textwidth]{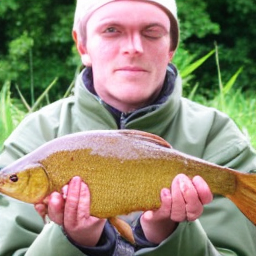} & \includegraphics[width=0.2\textwidth]{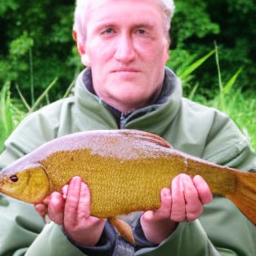}
    \end{tabular}
    \captionof{figure}{Moving along \ExtraDir{}s (Section~\ref{sec:cafol-overall-pipeline}) will alter the appearance of the image while preserving the class identity. $E_i$ represents the editing strength. }
    \label{fig:extra-dir-class-preserve}
\end{table}

We show that the \ExtraDir{}s (Section~\ref{sec:cafol-overall-pipeline}) carry semantics that are not related to the class identity in Figure~\ref{fig:extra-dir-class-preserve}. We adopt the method proposed by \citet{riemannian-diffusion-edit}. The editing focuses mostly on the background and preserves the class identity, as long as the movement does not orthogonalize the latent code and the editing direction. This phenomenon motivates our experiments.

\section{The layer index for Jacobian calculation}
\label{sec:supp-jac-layer-index}
\begin{figure}
    \centering
    \includegraphics[width=1.0\linewidth, trim={0 170 0 170},clip]{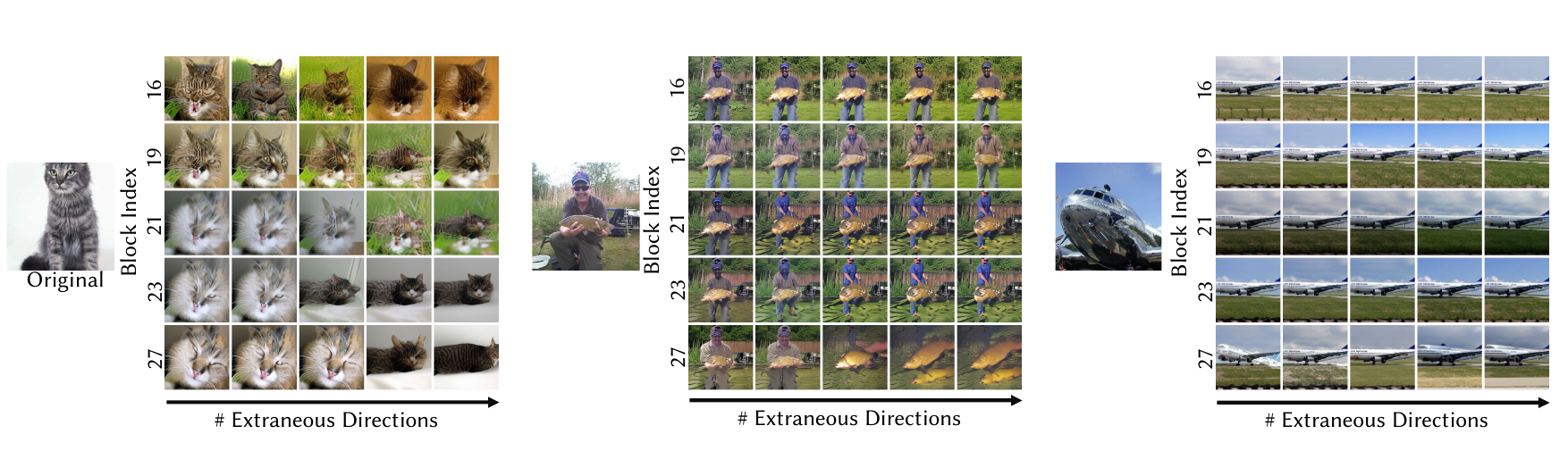}
    \caption{\Cano{}s when computing \ExtraDir{}s at different layers in a DiT \cite{dit}. We use $t_e=0.8T$. Note that we do not use CFG after the \ExtraDir{}s projection, to present a more straightforward comparison between different layers. We choose $l=27$, \textit{i.e.} the last layer, to ensure an adequate change in the input.}
    \label{fig:dit-layer-idx}
\end{figure}
In Section~\ref{sec:cafol-overall-pipeline}, we treat the CDM as a feature extractor for calculating the Jacobian. We visualize the effect of selecting different $l$, \textit{i.e.} the layer index for computing the Jacobian, in Figure~\ref{fig:dit-layer-idx}. While in some cases different layers can yield similar effects, we choose the last layer, namely $l=27$ (the minimum $l$ is 0), to ensure an adequate change in the input, or the background can remain unchanged in certain cases.

For Stable Diffusion \cite{ldm}, we follow the practice in \citet{riemannian-diffusion-edit} to select the layer index. Concretely, we extract the features after the middle block to ensure that the \ExtraDir{}s are semantically meaningful \cite{h-space-semantic,dm-already-semantic}. We adopt the same strategy for all UNet-based CDMs, including the one used in our CIFAR10 experiments in Section~\ref{sec:cano-quant-result} and the CDM in the EDM framework \cite{edm,edm2}. 

\section{The generalization of \MethodName{}}
\label{sec:supp-cafol-generalization}
\subsection{Fine-grained control of \Cano{}s with text conditioning}
\label{sec:supp-finegrained-text-control}
\begin{figure}[htp]
  \centering
  \begin{subfigure}[b]{0.24\textwidth}
    \centering
    \includegraphics[width=\textwidth]{figures/dit/fish2_ori.png}
    \caption{}%
    \label{fig:fish2-ori-text-cond-flexible-control}
  \end{subfigure}
  \begin{subfigure}[b]{0.24\textwidth}
    \centering
    \includegraphics[width=\textwidth]{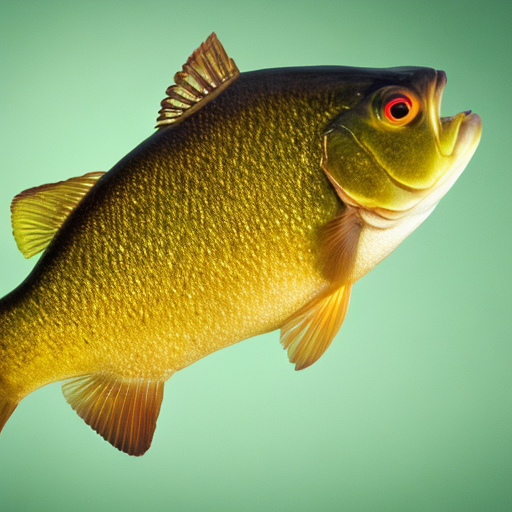}
    \caption{}%
    \label{fig:fish2-jacnormal-text-cond-flexible-control}
  \end{subfigure}
  \begin{subfigure}[b]{0.24\textwidth}
    \centering
    \includegraphics[width=\textwidth]{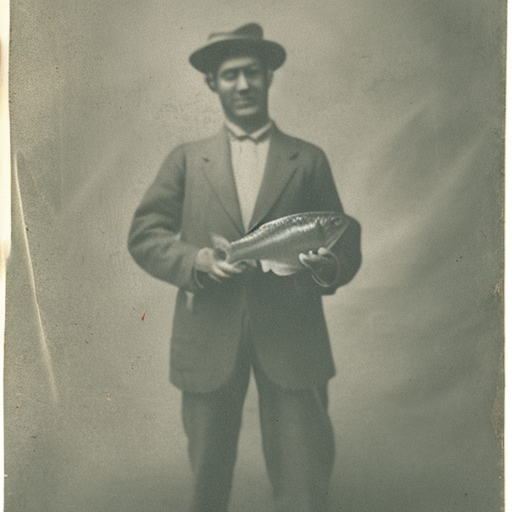}
    \caption{}%
    \label{fig:fish2-jacnew-text-cond-flexible-control}
  \end{subfigure}
  \begin{subfigure}[b]{0.24\textwidth}
    \centering
    \includegraphics[width=\textwidth]{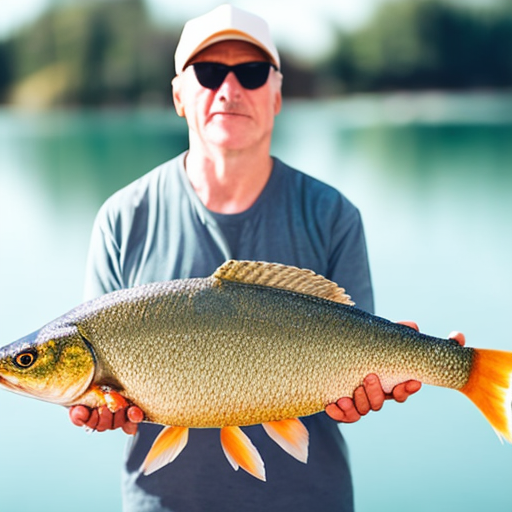}
    \caption{}%
    \label{fig:fish2-cfgnew-text-cond-flexible-control}
  \end{subfigure}
  \caption{(a): The original image from the class \textit{Tench}. (b): The \Cano{} obtained with prompt: \textit{a photo of tench}. (c): The \Cano{} obtained with prompt: \textit{a photo of a man holding a fish}. (d): An image generated with CFG using the same prompt as in (c), using the same starting noise as in (b) and (c).}
  \label{fig:fish-text-cond-flexible-control}
\end{figure}

\begin{figure}[htp]
  \centering
  \begin{subfigure}[b]{0.24\textwidth}
    \centering
    \includegraphics[width=\textwidth]{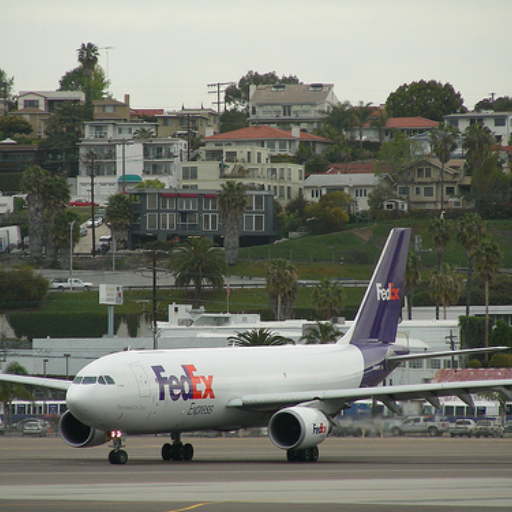}
    \caption{}%
    \label{fig:air-ori-text-cond-flexible-control}
  \end{subfigure}
  \begin{subfigure}[b]{0.24\textwidth}
    \centering
    \includegraphics[width=\textwidth]{appendix/figures/airplane_cfgjac_sd_show.png}
    \caption{}%
    \label{fig:air-jacnormal-text-cond-flexible-control}
  \end{subfigure}
  \begin{subfigure}[b]{0.24\textwidth}
    \centering
    \includegraphics[width=\textwidth]{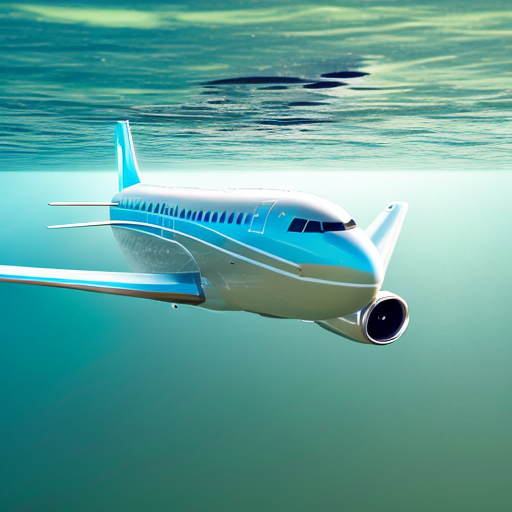}
    \caption{}%
    \label{fig:air-jacnew-text-cond-flexible-control}
  \end{subfigure}
  \begin{subfigure}[b]{0.24\textwidth}
    \centering
    \includegraphics[width=\textwidth]{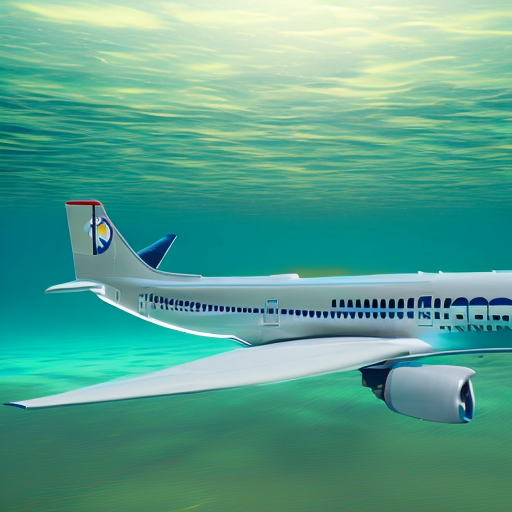}
    \caption{}%
    \label{fig:air-cfgnew-text-cond-flexible-control}
  \end{subfigure}
  \caption{(a): The original image from the class \textit{Airliner}. (b): The \Cano{} obtained with prompt: \textit{a photo of airliner}. (c): The \Cano{} obtained with prompt: \textit{an airplane flying under water}. (d): An image generated with CFG using the same prompt as in (c), using the same starting noise as in (b) and (c).}
  \label{fig:air-text-cond-flexible-control}
\end{figure}

\MethodName{} naturally extends to text-conditioning CDMs. Text-conditioning offers a more flexible control over where \Cano{}s lie than one-hot label conditioning. We show visual results in Figure~\ref{fig:fish-text-cond-flexible-control} and \ref{fig:air-text-cond-flexible-control}. The used CDM, a Stable Diffusion 2.1 \cite{ldm}, successfully adapts to different text prompts on the same image, which is in line with previous findings \cite{riemannian-diffusion-edit}. In Figure~\ref{fig:air-text-cond-flexible-control}, the CDM finds a \Cano{} that does not exist in the real world, but all the components in it are real, \textit{e.g.} the water and the airplane. The results demonstrate that it is possible to perform a more fine-grained control over where \Cano{}s lie. We believe investigating the relationship between the \Cano{}s and the text conditioning on the same image can reveal the image understanding capability of CDMs, which we leave as a promising future direction. 

\subsection{\MethodName{} is compatible with the EDM sampler and UViT architecture}
\label{sec:supp-edm-cafol}
The main idea behind \MethodName{} is to identify the latent vectors that carry non-discriminative information and render the latent code of a \DDIMinv{}-inverted sample orthogonal to them. Such a formulation does not require any knowledge about the sampler or the architecture of the model, as long as the DM has enough capability to model the data distribution. Here, we demonstrate preliminary results on the generalization of \MethodName{}. Specifically, we test the main idea behind \MethodName{} on the EDM sampler \cite{edm,edm2} with a UNet-based CDM, and a UViT \cite{uvit} model using the same DDIM sampler as in the main paper. All CDMs are trained on ImageNet. We implement the inversion sampler of EDM. Here, we aim at showing the effectiveness of the identification of \ExtraDir{}s but not on the feature quality. Hence, we do not perform the same analysis as in Section~\ref{sec:optimal-kt}. We present some visual results to demonstrate the intuitive summary of the categorical semantics, as shown in Figure~\ref{fig:visual-result-edm} and \ref{fig:visual-result-uvit}. We believe investigating the relationship between the performance of the DM in generative tasks and it as teacher in \textbf{\textit{\CanoDistill{}}} is promising, as previous works have found \cite{gen-data-robust3-adv,dm-discriminative3-ddae}, and leave it as a future work. 

\begin{table}[]
  \centering
  \resizebox{\textwidth}{!}{
  \begin{tabular}{
    c|ccc||ccc                            
  }
  & Class & Original & \MethodName{} & Class & Original & \MethodName{} \\
  \midrule
  \multirow{3}{*}[5pt]{\rotatebox{90}{\parbox{1.64cm}{\centering \textbf{\hspace{0pt} EDM \cite{edm,edm2}}}}} & \rotatebox{90}{\parbox{2.5cm}{\centering \textbf{\hspace{0pt} Tench}}}
    & \includegraphics[width=0.2\textwidth]{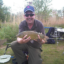}
    & \includegraphics[width=0.2\textwidth]{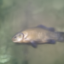}
    & \rotatebox{90}{\parbox{2.5cm}{\centering \textbf{\hspace{0pt} Wine Bottle}}}
    & \includegraphics[width=0.2\textwidth]{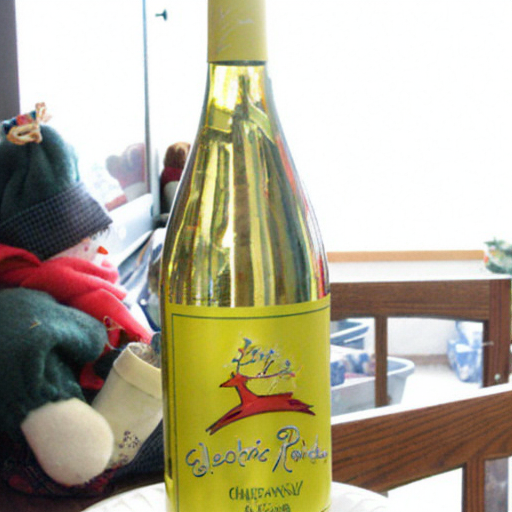}
    & \includegraphics[width=0.2\textwidth]{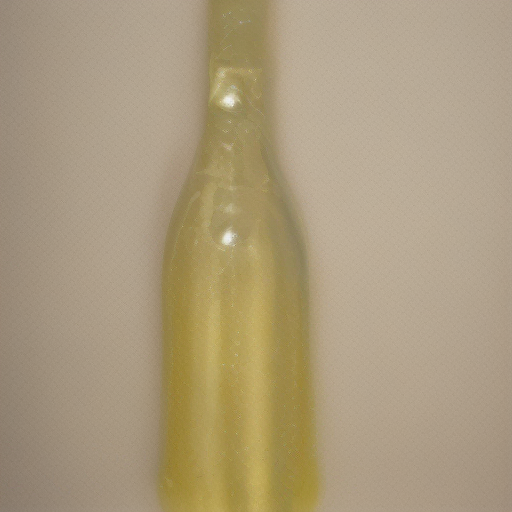} \\
    & \rotatebox{90}{\parbox{2.5cm}{\centering \textbf{\hspace{0pt} Airliner}}} & \includegraphics[width=0.2\textwidth]{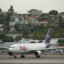}
    & \includegraphics[width=0.2\textwidth]{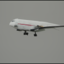}
    & \rotatebox{90}{\parbox{2.5cm}{\centering \textbf{\hspace{0pt} Screen}}}
    & \includegraphics[width=0.2\textwidth]{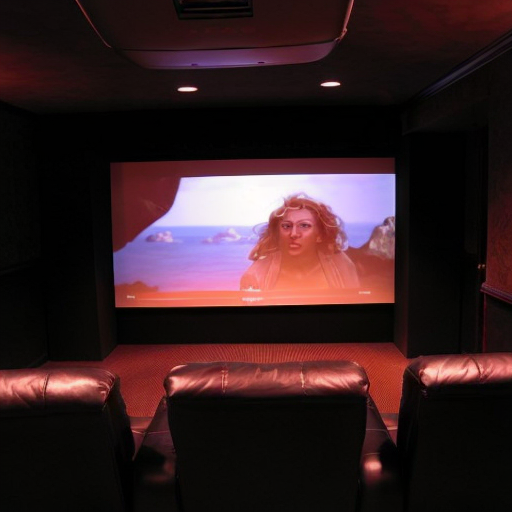}
    & \includegraphics[width=0.2\textwidth]{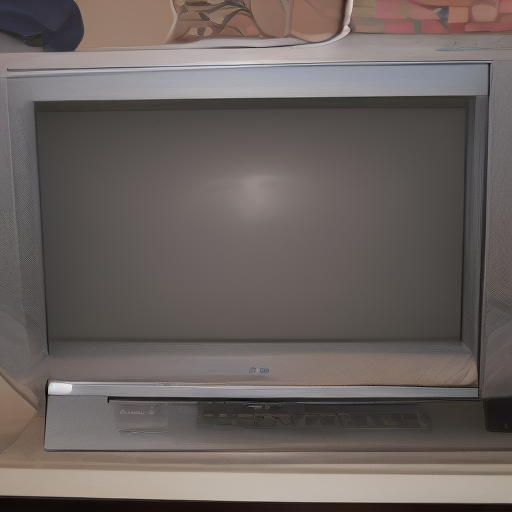} \\
  \end{tabular}
    }
  \captionof{figure}{Preliminary results on \Canoimg{}s generated by the EDM framework \cite{edm,edm2}. The left column is an EDM trained on ImageNet $64\times64$, the right one is on ImageNet $512\times 512$. We implement the inversion sampler of EDM. It indicates that the main idea behind \MethodName{} is also effective when facing inputs with different resolutions. Note that we do not use CFG or Autoguidance \cite{autoguidance} to improve the visual quality, to provide a more straightforward insight into the effectiveness of \MethodName{} on EDM.}
  \label{fig:visual-result-edm}
\end{table}

\begin{table}[]
  \centering
  \resizebox{\textwidth}{!}{
  \begin{tabular}{
    c|ccc||ccc                            
  }
  & Class & Original & \MethodName{} & Class & Original & \MethodName{} \\
  \midrule
  \multirow{3}{*}[5pt]{\rotatebox{90}{\parbox{1.64cm}{\centering \textbf{\hspace{0pt} EDM \cite{edm,edm2}}}}} & \rotatebox{90}{\parbox{2.5cm}{\centering \textbf{\hspace{0pt} Robin}}}
    & \includegraphics[width=0.2\textwidth]{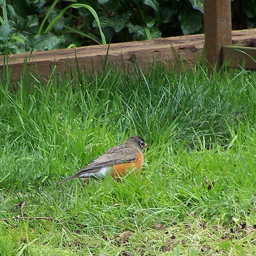}
    & \includegraphics[width=0.2\textwidth]{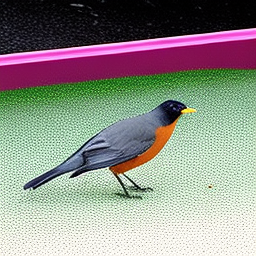}
    & \rotatebox{90}{\parbox{2.5cm}{\centering \textbf{\hspace{0pt} Goldfish}}}
    & \includegraphics[width=0.2\textwidth]{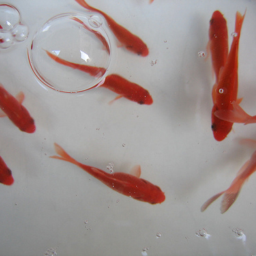}
    & \includegraphics[width=0.2\textwidth]{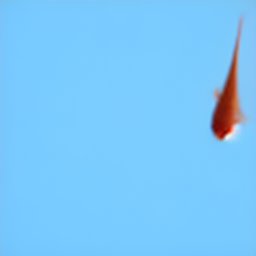} \\
    & \rotatebox{90}{\parbox{2.5cm}{\centering \textbf{\hspace{0pt} Jeep}}} & \includegraphics[width=0.2\textwidth]{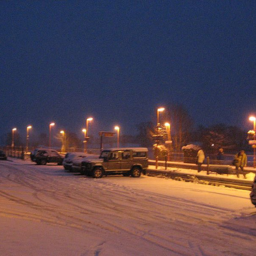}
    & \includegraphics[width=0.2\textwidth]{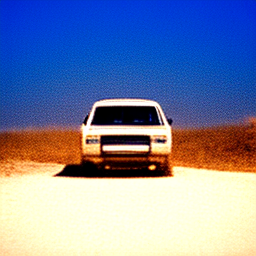}
    & \rotatebox{90}{\parbox{2.5cm}{\centering \textbf{\hspace{0pt} Wall Clock}}}
    & \includegraphics[width=0.2\textwidth]{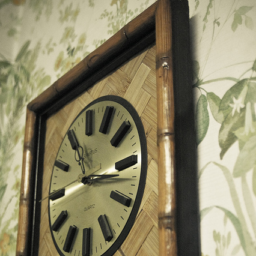}
    & \includegraphics[width=0.2\textwidth]{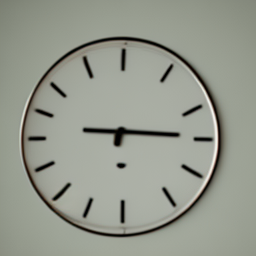} \\
  \end{tabular}
    }
  \captionof{figure}{Preliminary results on \Canoimg{}s generated by a ImageNet-256 UViT \cite{uvit} using DDIM \cite{ddim} sampler. It indicates that the main idea behind \MethodName{} is effective on different architectures of CDMs. Note that we do not use CFG to improve the visual quality, to provide a more straightforward insight into the effectiveness of \MethodName{} on UViT. }
  \label{fig:visual-result-uvit}
\end{table}

\section{Details of the ImageNet20 experiment}
\label{sec:imagenet20}
\begin{figure}
    \centering
    \includegraphics[width=1.0\linewidth,trim={0 0 90 65}, clip]{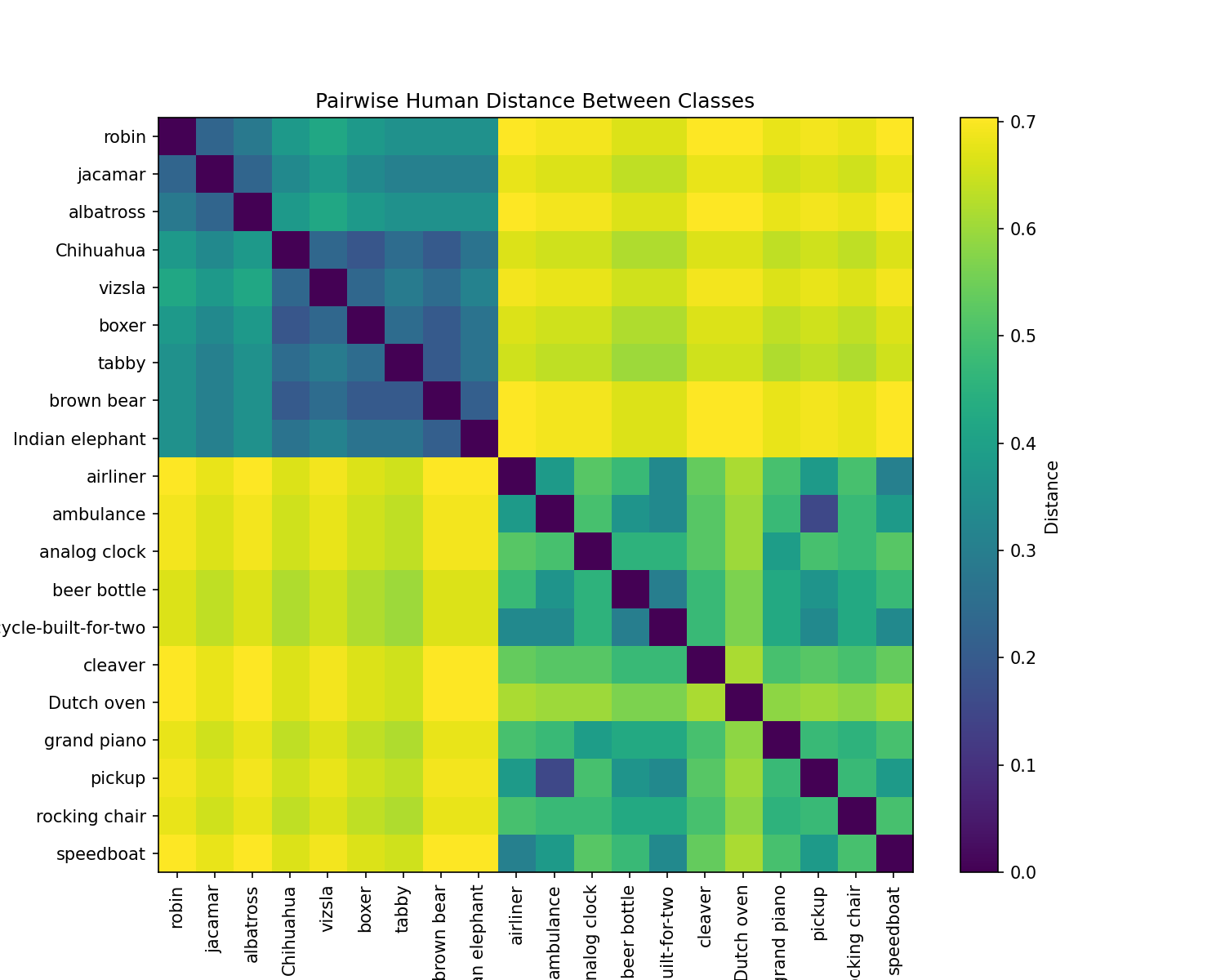}
    \caption{Wu-Palmer (WUP) distances (1.0-WUP similarity \cite{wup-sim}) between the classes in ImageNet20. The class relationships are structured, hence appropriate for analyzing the class separability in our case.}
    \label{fig:wup-imagenet20}
\end{figure}

To develop the \MethodName{} framework, we perform quantitative experiments on a 20-class subset of ImageNet \cite{deng2009imagenet}. This choice balances between computational costs and statistical significance. The indices of the 20 classes are: [15, 95, 146, 151, 211, 242, 281, 294, 385, 404, 407, 409, 440, 444, 499, 544, 579, 717, 765, 814]. The corresponding class names are: ['robin', 'jacamar', 'albatross', 'Chihuahua', 'vizsla', 'boxer', 'tabby', 'brown bear', 'Indian elephant', 'airliner', 'ambulance', 'analog clock', 'beer bottle', 'bicycle-built-for-two', 'cleaver', 'Dutch oven', 'grand piano', 'pickup', 'rocking chair', 'speedboat']. We plot the pair-wise Wu-Palmer (WUP) distances \cite{wup-sim} between the selected classes in Figure~\ref{fig:wup-imagenet20}. The selected classes can be similar or dissimilar to each other, demonstrating certain structures. This is appropriate for analyzing the class separability in our case. We choose 50 images from the ImageNet20, building a 1000-sample dataset for the following analysis.

\subsection{Selecting the optimal layer and time step for feature extraction}
\label{sec:imagenet20-layer-timestep}
\begin{figure}
    \centering
    \begin{minipage}{0.49\textwidth}
        \centering
        \includegraphics[width=\textwidth]{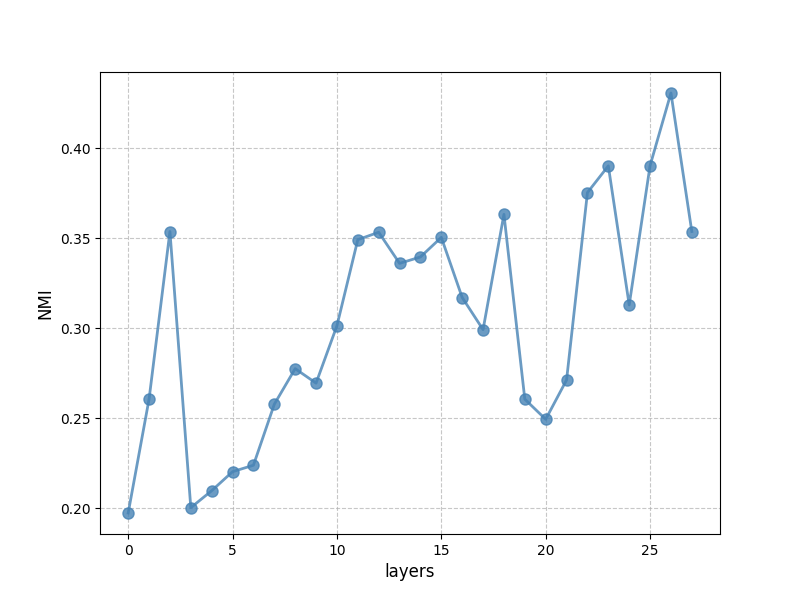}
        \caption{NMI v.s. layers on ImageNet20 in a DiT \cite{dit}, fixing $t_r=0.1T$. We choose the penultimate layer, \textit{i.e.} the 27th layer (the figure has a 0-start index), in all our experiments.}
        \label{fig:layer-nmi-imagenet20}
    \end{minipage}\hfill
    \hspace{1pt}
    \begin{minipage}{0.49\textwidth}
        \centering
        \includegraphics[width=\textwidth]{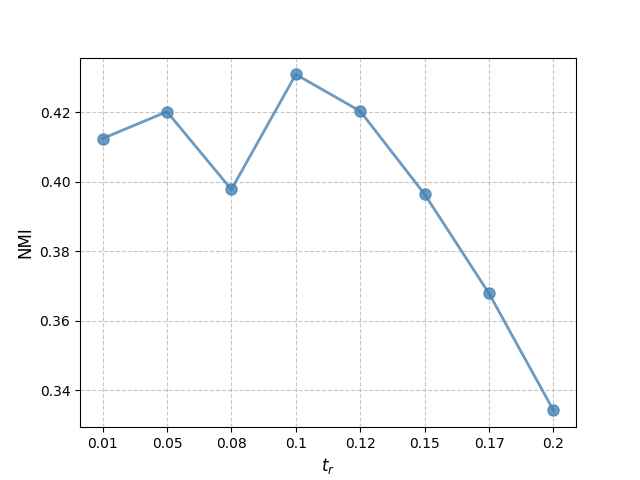}
        \caption{NMI v.s. feature extraction time step ($t_r$) on ImageNet20 using a DiT \cite{dit}, fixing the layer index to be 27. We choose $t_r=0.1T$ in all our experiments.}
        \label{fig:time-nmi-imagenet20}
    \end{minipage}
\end{figure}
We choose an ImageNet $256\times 256$-pretrained DiT-XL \cite{dit} model for the quantitative analysis. We consider the outputs of all 28 ViT blocks in it. For the time step, we select $t_r=\{0.01,0.05,0.08,0.1,0.12,0.15,0.17,0.2\}$. We perform K-means clustering on the feature map after average pooling, and compute the normalized mutual information between the cluster assignments and the ground truth class labels. The average pooling reduces noise in the feature map, making the cluster more accurate and compact. The results for different layers and time steps are shown in Figure~\ref{fig:layer-nmi-imagenet20}, \ref{fig:time-nmi-imagenet20}. Our conclusion on the time step for feature extraction is consistent with previous studies \cite{dm-beat-gan-classification,dm-as-representation-learner} that adopt linear probing on the features for quantifying the feature quality, albeit with different DM architectures. Such a result provides \textbf{evidence on the validity of our metric, \textit{i.e.} normalized mutual information.}

\subsection{On using NMI for measuring feature quality and the effectiveness of \textbf{\textit{\CanoDistill{}}}}
\label{sec:supp-limit-nmi-not-converging}
While NMI is valid for self-evaluation of the feature quality within the \MethodName{} framework, it is not valid for a direct comparison between the quality of features obtained via different methods, such as CFG. For example, when using CFG on the same 1000 samples after \DDIMinv{} and extracting features at the same $t_r$ as in \MethodName{}, it can yield a higher NMI (0.6108 in \MethodName{} v.s. 0.7808 with CFG magnitude being 4.0). Adopting CFG after performing \ExtraDir{} projection can also improve the metric (0.7762 in \MethodName{} with CFG being 3.0). However, NMI only examines the compactness and separability of each cluster, ignoring the low-dimensional manifold structure of the data. For example, a line-shaped manifold and a circle-shaped manifold can yield the same NMI, while they capture different characteristics of the data. In an extreme case where the features are constant for all samples belonging to the same class, the NMI will be 1.0, while the features are not meaningful in this case. \textbf{In other words, we do not want the student to just learn a "converging" prior on the features belonging to the same classes, but to learn actual class semantics}. This claim is validated in our extended ablation studies in Section~\ref{sec:supp-ablation}, where the student does not perform well when the CFG magnitude is high. These results undermine the notion that the student only mimics the CDM’s label-conditioning embeddings: those embeddings behave as a trivial collapsing prior rather than conveying the encoded semantic structure. A promising future direction is to develop new feature quality metrics or adapt existing ones into \MethodName{} that consider the structure of the data. This metric is also invalid on datasets with simple data structures, such as CIFAR10 \cite{cifar10}. The CDM features of the original samples are already perfectly separable and achieve an NMI of 1.0. However, as shown in Table~\ref{tab:main-quant} in the main paper, \MethodName{} can still find more semantically meaningful samples than the original ones, and convey the knowledge to the student via \textbf{\textit{\CanoDistill{}}}. While the simple structure of the data prevents the utilization of NMI, it is significantly easier to process than real image datasets. Either calculating the \ExtraDir{}s or training with \textbf{\textit{\CanoDistill{}}} requires far less computational resources than large, real-image datasets such as ImageNet \cite{deng2009imagenet}. Therefore, we simply brute-force search the best hyperparameters in \MethodName{} on CIFAR10. The hyperparameters on CIFAR10 are: $t_e=0.8,n=10,t_r=0.13T,$ layer $= up.0$. We also only use a 10\% subset for obtaining \Cano{}s, as shown in Table~\ref{tab:main-quant}.
\begin{figure}
    \centering
    \includegraphics[width=1.0\linewidth]{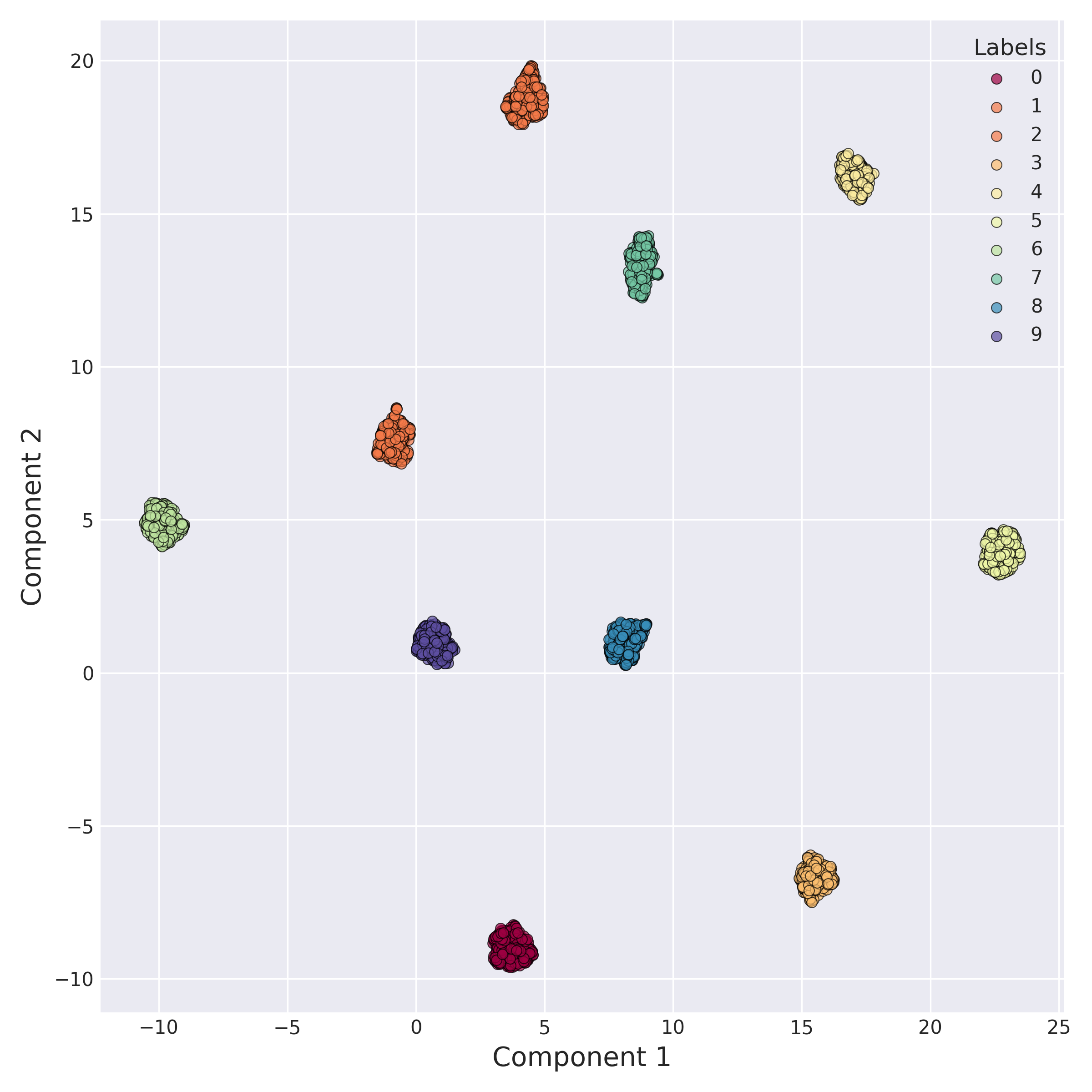}
    \caption{The CDM features of the original samples in CIFAR10 \cite{cifar10} are already separable and achieve an NMI of 1.0. However, \MethodName{} is still effective in this case, as shown in Table~\ref{tab:main-quant}.}
    \label{fig:umap-cifar-ori}
\end{figure}

\section{Choosing hyperparameters $t_e$ and $n$ for \MethodName{}}
\label{sec:supp-te-k}
\subsection{Finding the optimal $t_e$}
\label{sec:supp-te}
\begin{figure}[htp]
  \centering
  \begin{subfigure}[b]{0.49\textwidth}
    \centering
    \includegraphics[width=\textwidth]{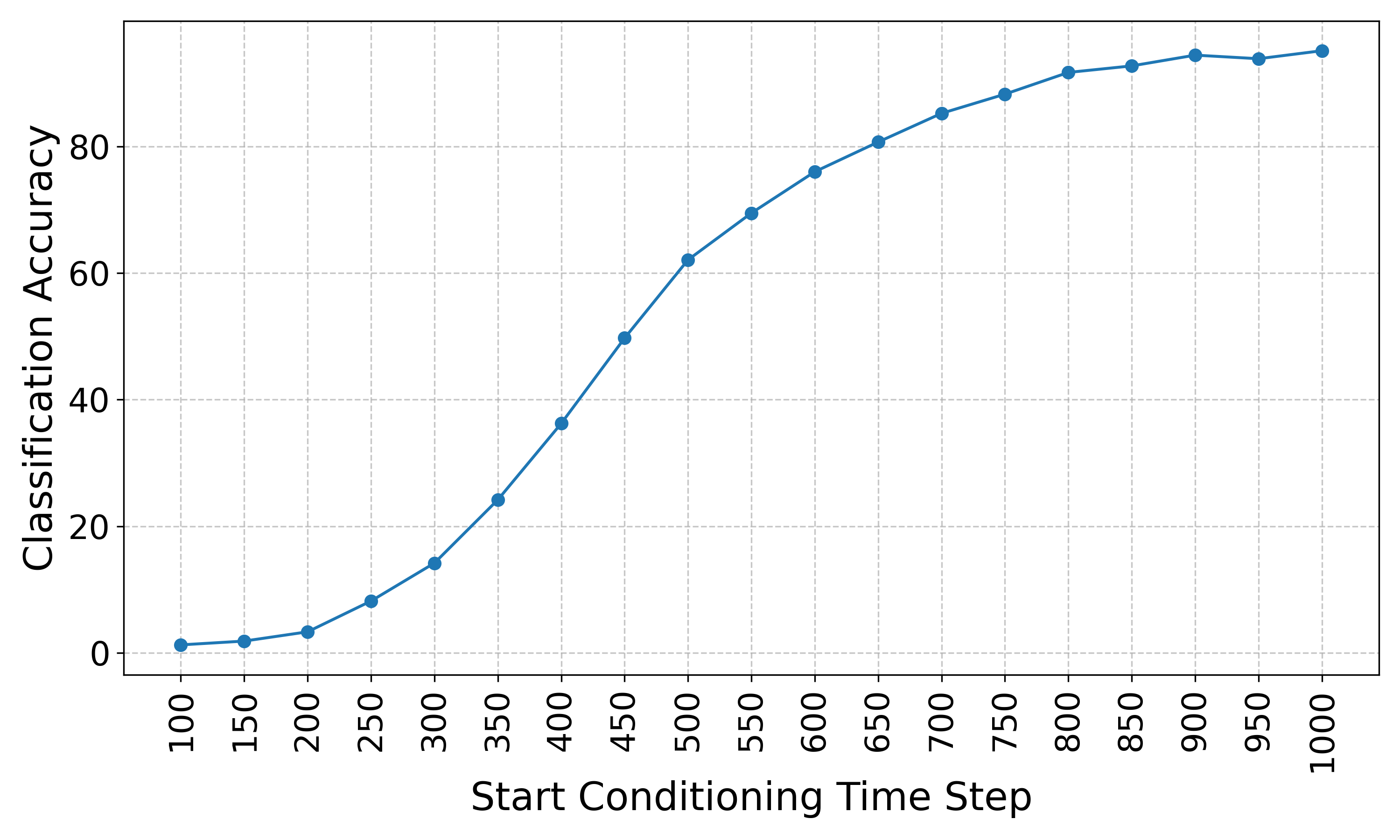}
    \caption{}%
    \label{fig:dit-acc}
  \end{subfigure}
  \begin{subfigure}[b]{0.49\textwidth}
    \centering
    \includegraphics[width=\textwidth]{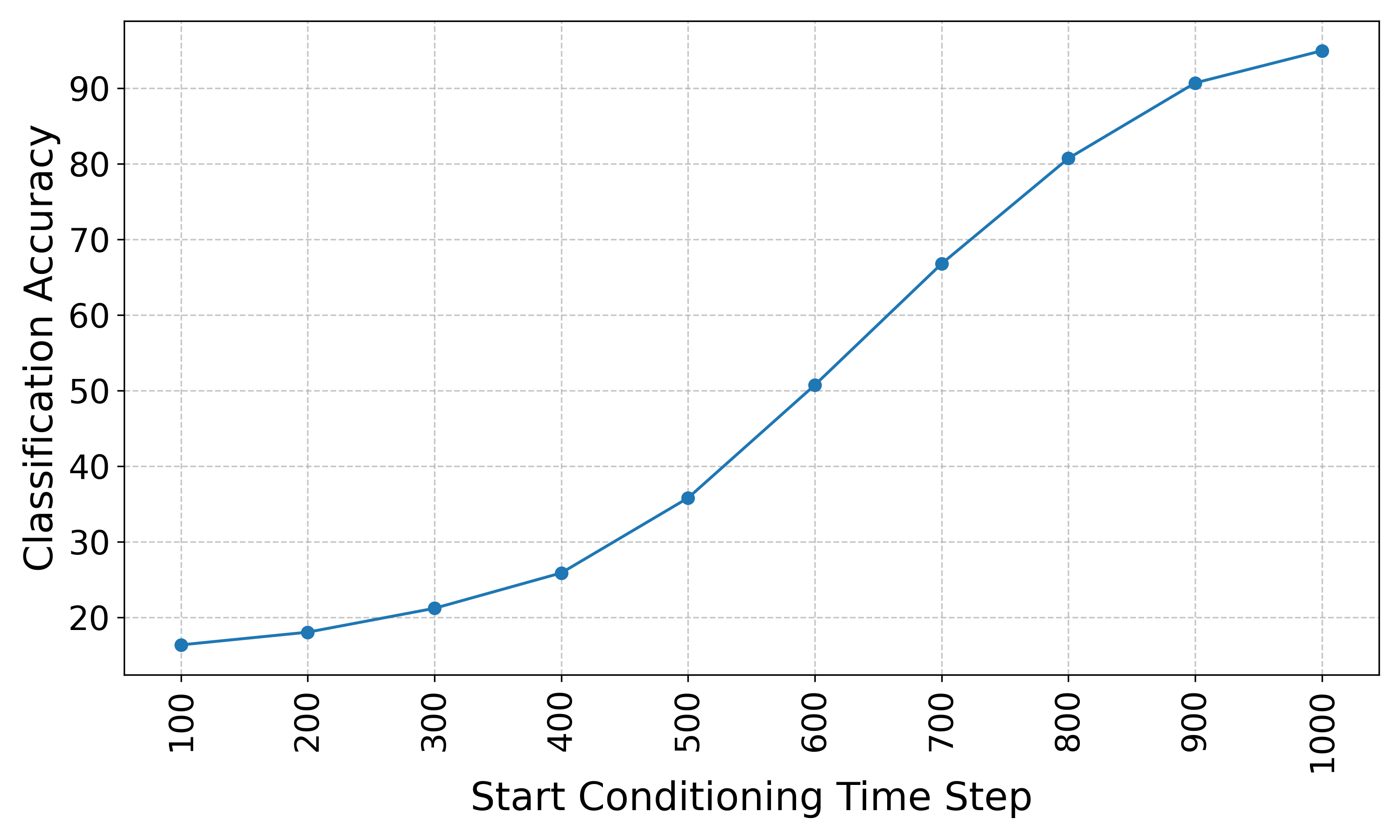}
    \caption{}%
    \label{fig:sd-acc}
  \end{subfigure}
  \caption{The classification accuracy on samples generated by the two-stage strategy in Section~\ref{sec:cafol-overall-pipeline}, using \textbf{(a)} DiT \cite{dit} and \textbf{(b)} Stable Diffusion \cite{ldm}. The accuracy is averaged over 3 classifiers.}
  \label{fig:supp-sd-dit-acc}
\end{figure}
\begin{figure}[htp]
  \centering
  \begin{subfigure}[b]{0.99\textwidth}
    \centering
    \includegraphics[width=\textwidth, trim={0 140 0 150}, clip]{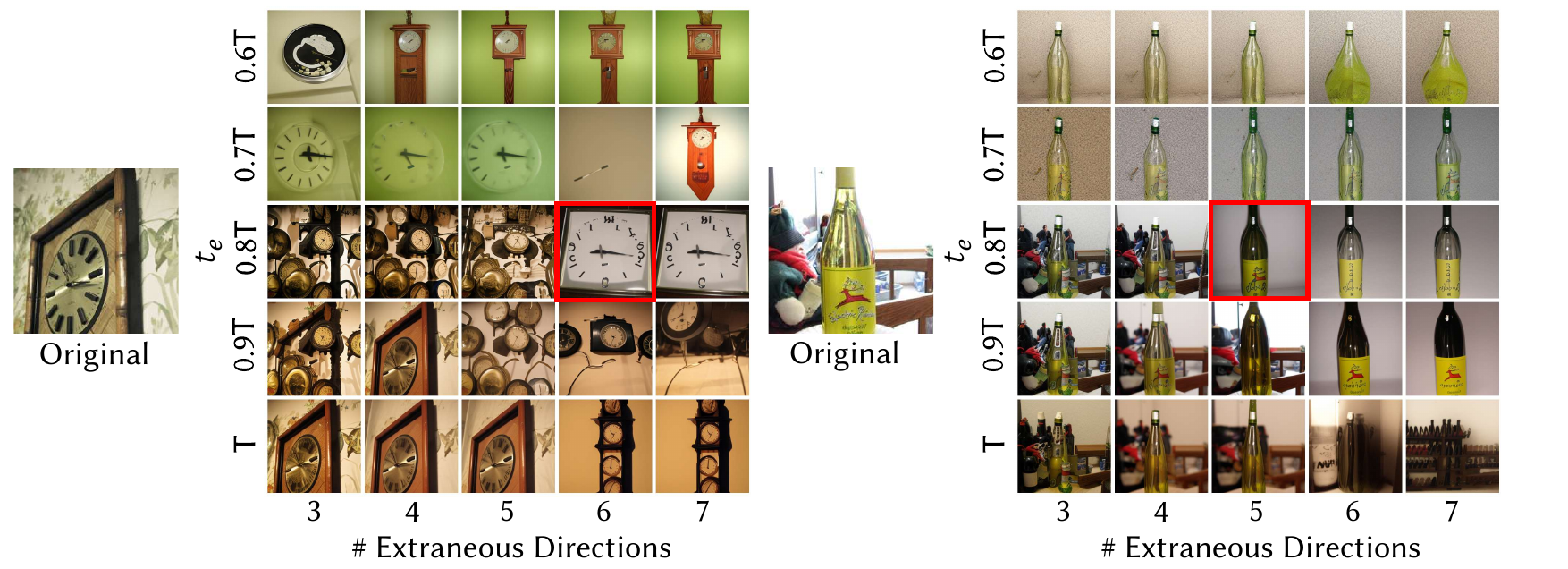}
    \caption{Results of DiT.}%
    \label{fig:dit-jac-kt-compare}
  \end{subfigure}
  \begin{subfigure}[b]{0.99\textwidth}
    \centering
    \includegraphics[width=\textwidth, trim={0 140 0 150}, clip]{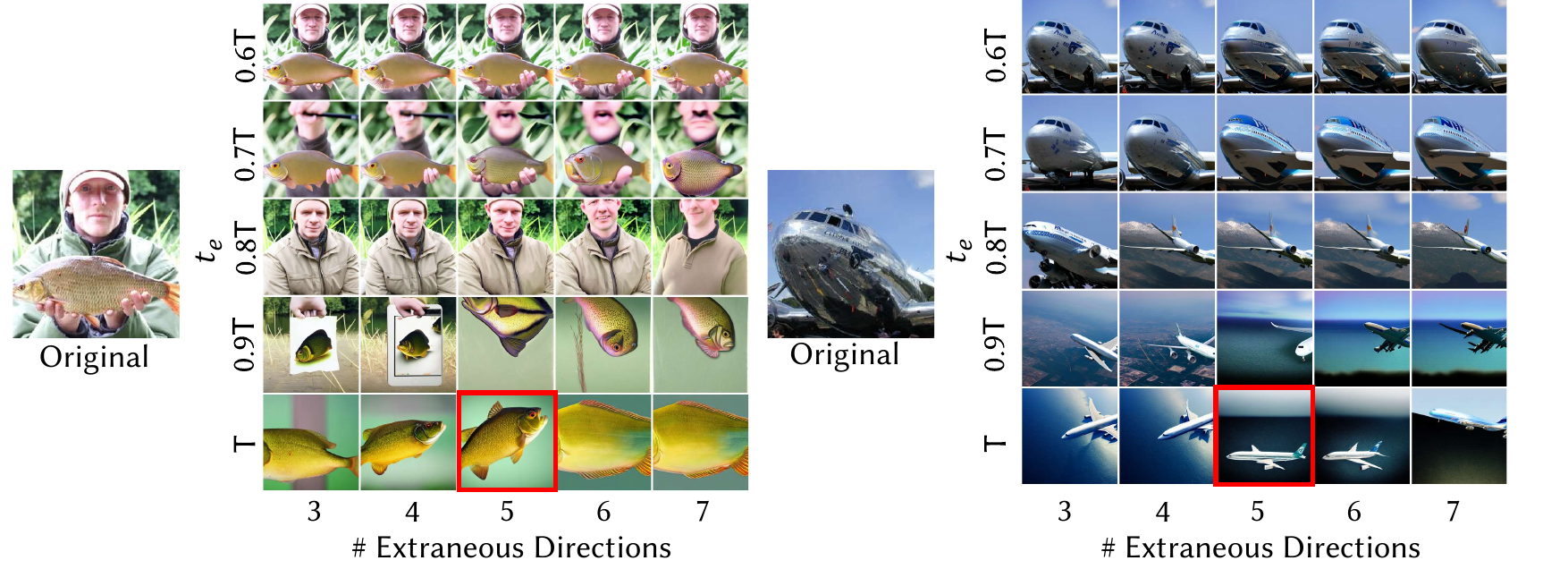}
    \caption{Results of Stable Diffusion.}%
    \label{fig:sd-jac-kt-compare}
  \end{subfigure}
  \caption{Visual comparisons between the images generated by selecting different $t_e$ and $k$ in \MethodName{}. Red boxes indicate the one automatically selected by our method.}
  \label{fig:jac-kt-compare}
\end{figure}

In Section~\ref{sec:cafol-overall-pipeline}, we decide $t_e$ by finding the saturation point of classification accuracy on samples generated by our two-stage strategy. We use 3 classifiers and compute the average accuracy of them: 
\begin{enumerate}
    \item A ViT-Large pre-trained on ImageNet12K \cite{deng2009imagenet}. The input size is 224.
    \item An ImageNet22k-pre-trained Swin V2 \cite{swinv2}. The input size is 256.
    \item An ImageNet22k-pre-trained ConvNeXt V2 \cite{convnextv2}. The input size is 384. 
\end{enumerate}
All model weights are downloaded from the PyTorch Image Model library (timm) \cite{timm}. The accuracy curve for DiT is shown in Figure~\ref{fig:supp-sd-dit-acc}. The maximum time step $T$ is 1000 for both DiT \cite{dit} and Stable Diffusion \cite{ldm}. We use DDIM as the sampler and set the total diffusion time step to 50. For each class, we generate 5 images. Hence $m=5$ in Section~\ref{sec:cafol-overall-pipeline}. The prompt template for Stable Diffusion is: \textit{a photo of <class name>}, where the class name is the WordNet name of each ImageNet class \cite{deng2009imagenet}. We identify $t_e=800$ ($0.8T$) for DiT and $t_e=1000$ ($T$) for Stable Diffusion. We provide visual comparisons between the images generated by selecting different $t_e$ in Figure~\ref{fig:jac-kt-compare}. Qualitatively, a small $t_e$ might lead to insufficient changes in the input image, while a large $t_e$ in DiT can fail to make the model aware of the class conditioning, resulting in less meaningful \ExtraDir{}s. 

\subsection{Choosing the total number of \ExtraDir{}s $n$ for adaptively selecting $k$}
\label{sec:supp-k}
\begin{figure}[htp]
  \centering
  \begin{subfigure}[b]{0.49\textwidth}
    \centering
    \includegraphics[width=\textwidth, trim={220 220 290 220}, clip]{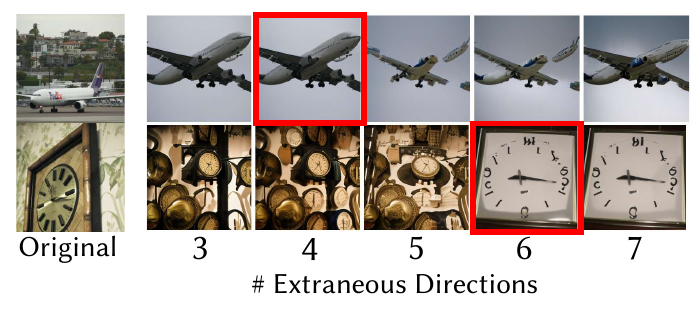}
    \caption{Results of DiT without CFG for a more straightforward comparison.}%
    \label{fig:dit-jac-different-k}
  \end{subfigure}
  \begin{subfigure}[b]{0.49\textwidth}
    \centering
    \includegraphics[width=\textwidth, trim={220 220 290 220}, clip]{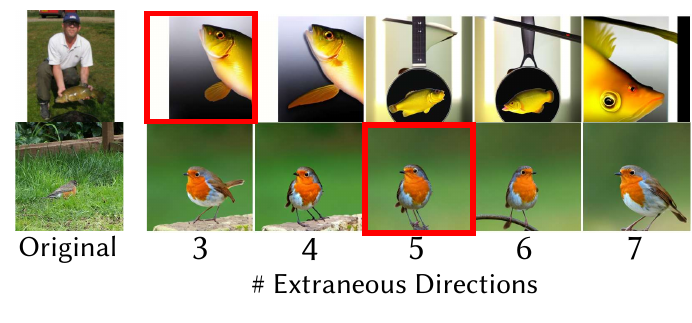}
    \caption{Results of Stable Diffusion with CFG magnitude being 7.5.}%
    \label{fig:sd-jac-different-k}
  \end{subfigure}
  \caption{Visual comparisons between the images generated by selecting different $k$ in \MethodName{}. Red boxes indicate the one automatically selected by our method.}
  \label{fig:jac-different-k}
\end{figure}

Our key observation on selecting $k$, \textit{i.e.} the number of \ExtraDir{}s to be projected, is that the effect caused by projecting \ExtraDir{}s is not smooth with varying $k$. That is, projecting one more \ExtraDir{}s can cause significant changes in the input. It is because \ExtraDir{}s are orthogonal to each other by design, hence there is no guarantee that their semantics are correlated. Importantly, projecting more \ExtraDir{}s does not necessarily mean a more separable set of \Canofeat{}s. It can lead to a loss of the class-defining cues, as shown in the fish image in Figure~\ref{fig:sd-jac-different-k}. We show the visual effects of selecting different $k$ in Figure~\ref{fig:jac-kt-compare}, \ref{fig:jac-different-k}. Our method \MethodName{} selects $k$ by adaptively choosing the elbow point on the explained variance ratio (EVR) sequence with total number of \ExtraDir{}s being $n$. The algorithm for finding the elbow point is presented in Algorithm~\ref{alg:knee-elbow}.

\begin{algorithm}[H]
\caption{Find Elbow via Knee Method(Sequence $S[0\ldots n-1]$)}
\begin{algorithmic}[1]
\STATE $n \leftarrow |S|$
\STATE $P_{\mathrm{start}} \leftarrow (0,\,S[0])$
\STATE $P_{\mathrm{end}}   \leftarrow (n-1,\,S[n-1])$
\STATE $v \leftarrow P_{\mathrm{end}} - P_{\mathrm{start}}$
\STATE $u \leftarrow v / \|v\|$
\FOR{$i = 0$ \TO $n-1$}
  \STATE $w \leftarrow (i,\,S[i]) - P_{\mathrm{start}}$
  \STATE $\mathrm{projLen} \leftarrow w \cdot u$
  \STATE $\mathrm{projVec} \leftarrow \mathrm{projLen}\times u$
  \STATE $\mathrm{perpVec} \leftarrow w - \mathrm{projVec}$
  \STATE $d[i] \leftarrow \|\mathrm{perpVec}\|$
\ENDFOR
\STATE $k \leftarrow \arg\max_{0\le i < n} d[i]$
\RETURN $k$
\end{algorithmic}
\label{alg:knee-elbow}
\end{algorithm}

\begin{figure}
    \centering
    \includegraphics[width=1.0\linewidth]{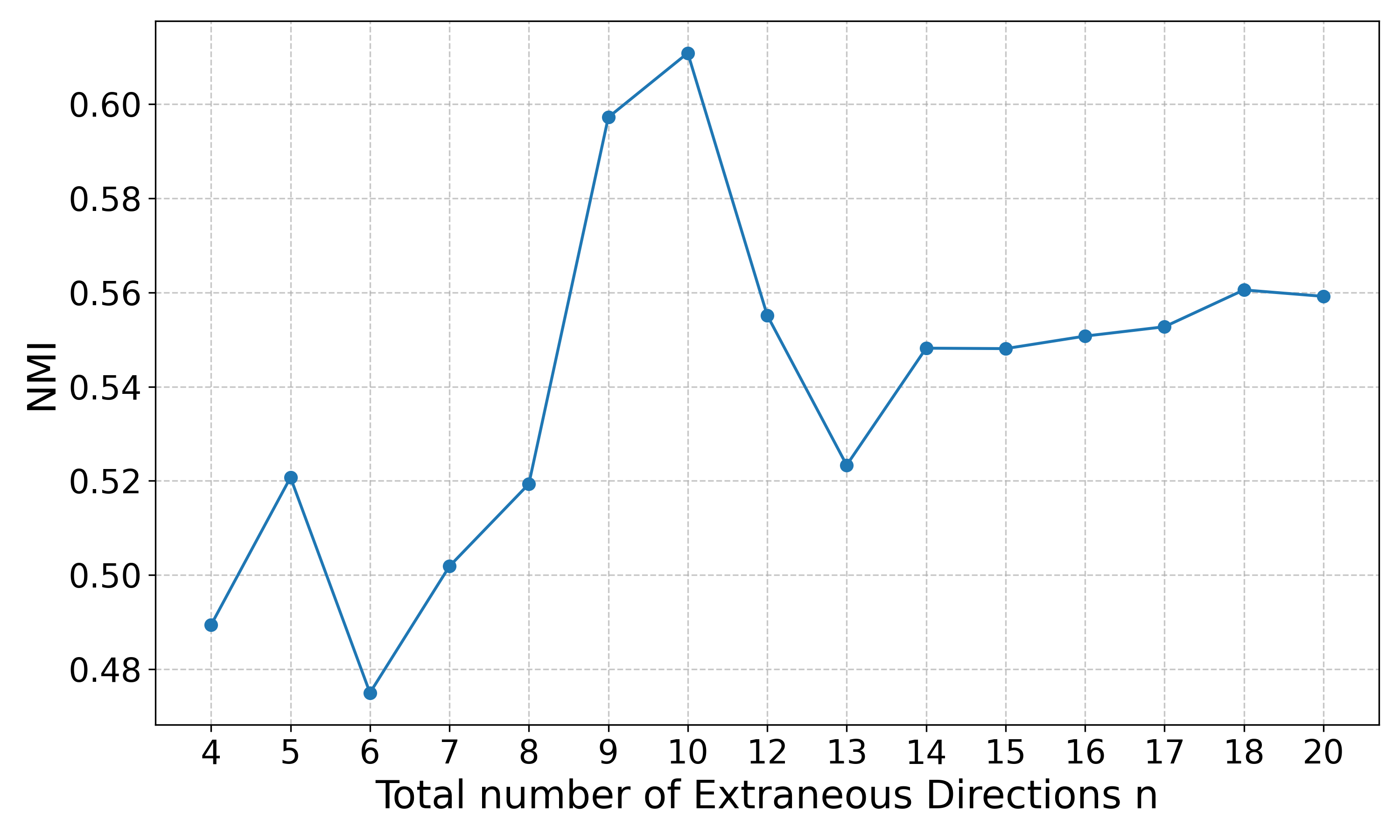}
    \caption{NMI on ImageNet20 v.s. total number of \ExtraDir{}s $n$ on DiT. A large $n$ can diminish the discriminative power of the input images, whereas a small one cannot change the inputs too much. Neither case is desired. Hence, we choose $n=10$ for DiT in our experiment. Note that this is a self-evaluation within the \MethodName{} framework, hence NMI is valid in this case.}
    \label{fig:elbow-num-different}
\end{figure}

\begin{figure}
    \centering
    \includegraphics[width=\linewidth]{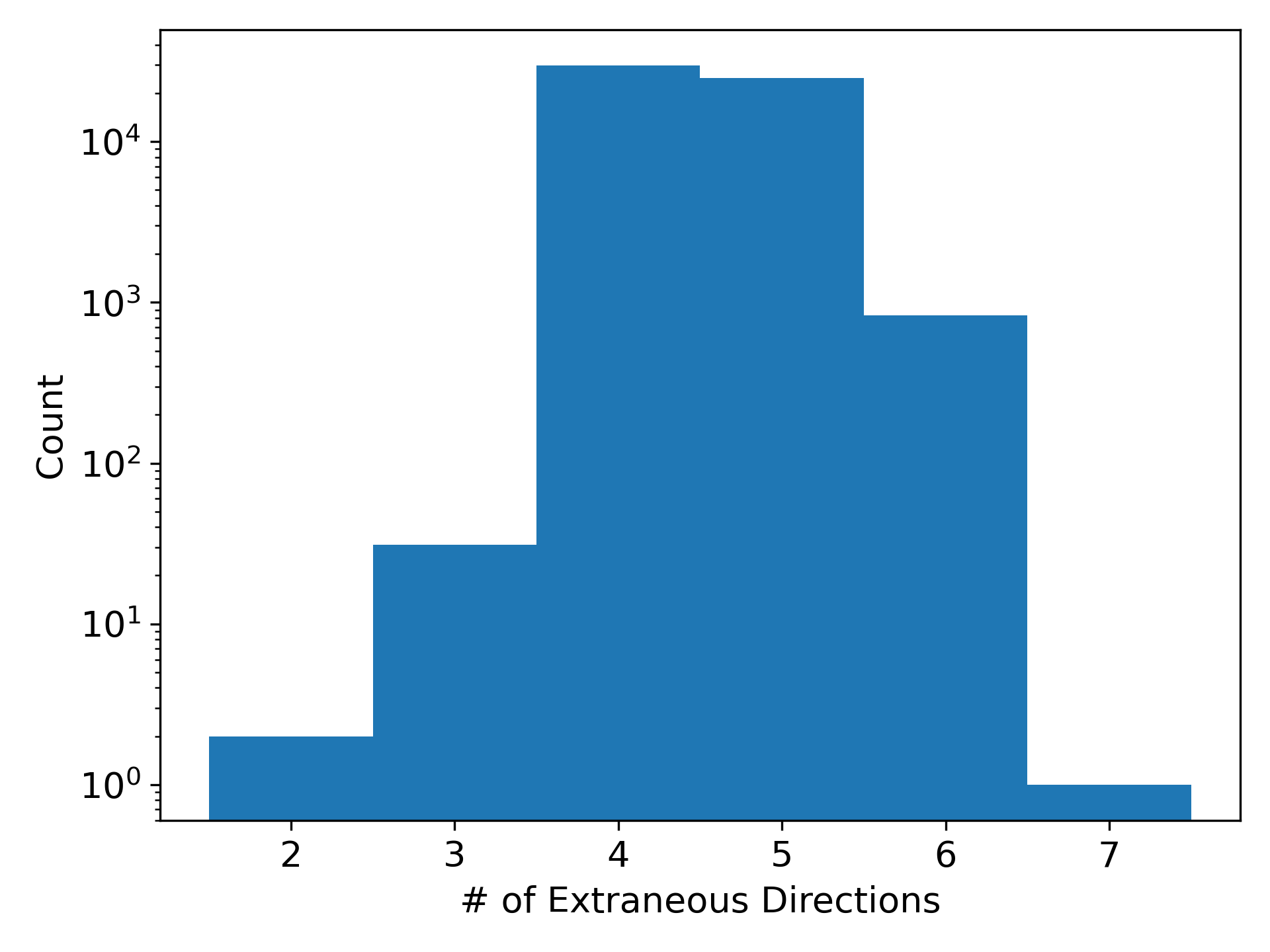}
    \caption{Histogram of $k$ when fixing $n=10$ on 78000 images from ImageNet100. We use this dataset in our experiments in Section~\ref{sec:supp-ablation}.}
    \label{fig:hist-jac}
\end{figure}

We show a quantitative comparison between different $n$ in Figure~\ref{fig:elbow-num-different} on ImageNet20. Too large $n$ tends to select a larger $k$, leading to eliminating too many components in the input image and diminishing discriminative power. A small $n$ will lead to a small $k$ and cannot change the inputs too much. We select $n=10$ for DiT, and $n=11$ for SD in our experiment. Note that this $n$ is tailored to specific CDMs. 
We plot the histogram of $k$ when fixing $n=10$ on 78000 images from ImageNet100, as we used in our experiments in Section~\ref{sec:supp-ablation}, in Figure~\ref{fig:hist-jac}. The selection process does not converge to a single $k$, supporting the effectiveness of our method. \MethodName{} has certain fault tolerance capacity, \textit{i.e.} slightly changing the number of projected \ExtraDir{}s can still result in desired images. For example, in Figure~\ref{fig:jac-different-k}, selecting $k=3$ or $k=4$ for the airplane image on DiT, or selecting $k=3\sim7$ for the bird image on Stable Diffusion, can all result in desired outputs. 

\subsubsection{Stable Diffusion 2.1}
\label{sec:supp-sd-quant-imagenet20}
\begin{figure}
    \centering
    \begin{minipage}{0.49\textwidth}
        \centering
        \includegraphics[width=\textwidth]{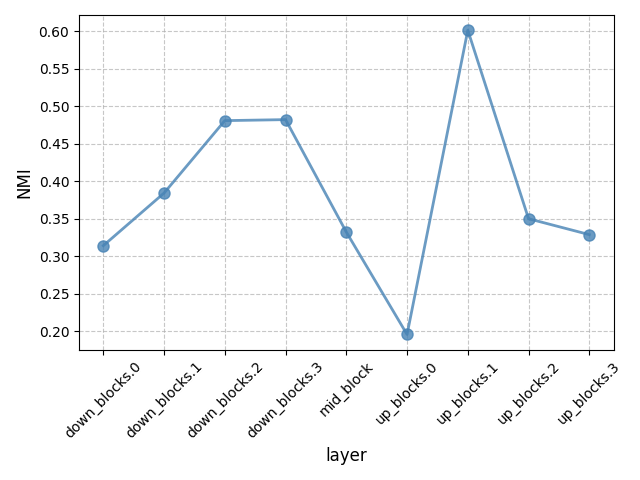}
        \caption{NMI v.s. layers on ImageNet20 in a Stable Diffusion 2.1 \cite{ldm}, fixing $t_r=0.13T$. We choose the up\_blocks.1 layer in all our experiments.}
        \label{fig:layer-nmi-imagenet20-sd}
    \end{minipage}\hfill
    \hspace{1pt}
    \begin{minipage}{0.49\textwidth}
        \centering
        \includegraphics[width=\textwidth]{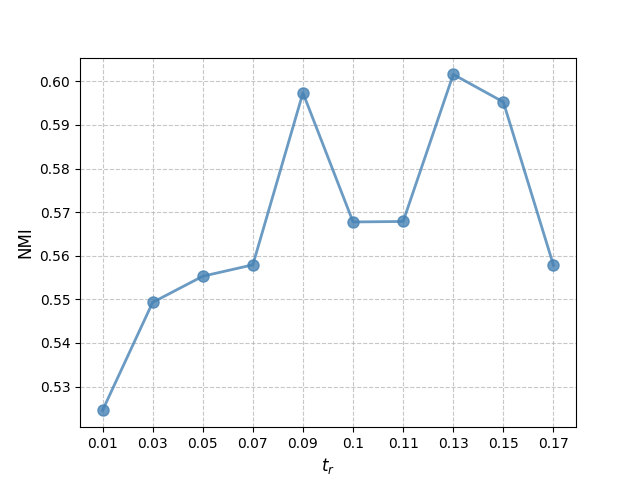}
        \caption{NMI v.s. feature extraction time step ($t_r$) on ImageNet20 using a Stable Diffusion 2.1 \cite{ldm}, fixing the layer index to be up\_blocks.1.}
        \label{fig:time-nmi-imagenet20-sd}
    \end{minipage}
\end{figure}

\begin{figure}
    \centering
    \includegraphics[width=1.0\linewidth]{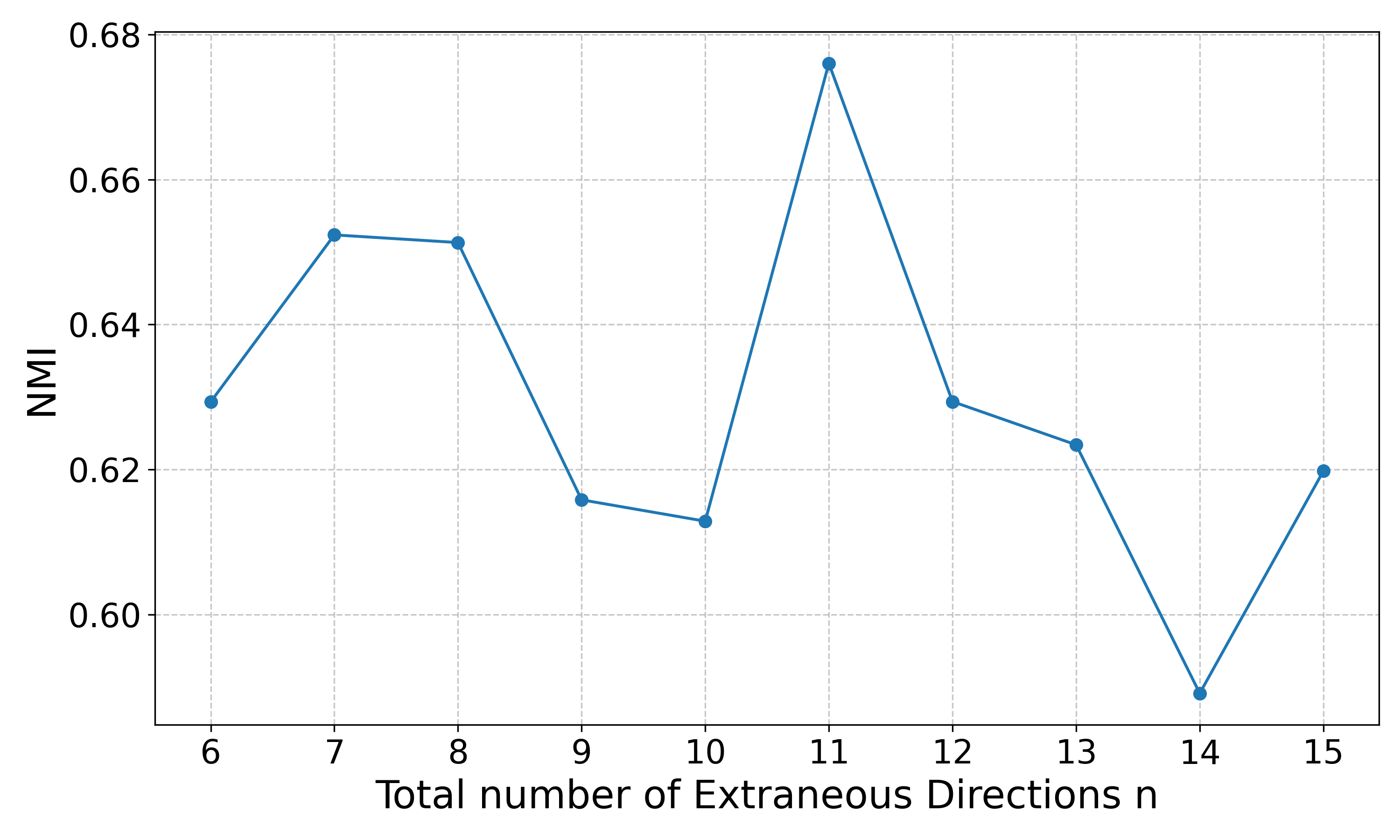}
    \caption{NMI on ImageNet20 v.s. total number of \ExtraDir{}s $n$ on a Stable Diffusion 2.1 \cite{ldm}. A large $n$ can diminish the discriminative power of the input images, whereas a small one cannot change the inputs too much. Neither case is desired. Hence, we choose $n=11$ for Stable Diffusion in our experiment. Note that this is a self-evaluation within the \MethodName{} framework, hence NMI is valid in this case. }
    \label{fig:elbow-num-different-sd}
\end{figure}

We provide additional results on Stable Diffusion 2.1 model \cite{ldm}. We first choose $t_r=0.13T$ and the layer to be up\_blocks.1, according to the results in Figure~\ref{fig:layer-nmi-imagenet20-sd}, \ref{fig:time-nmi-imagenet20-sd}. We then perform the same analysis as done in Section 3.2.1 in the main paper and Section H in the Appendix. Figure~\ref{fig:nmi-sd} shows the results. Observe that $t_e=T$, which is $t_e=0.999$ in the figure, yields the best results when performing fixed-$k$ projection. This validates our saturation-point-based $t_e$ selection, as described in Section 3.2.1 in the main paper and Section H in the Appendix. Again, our \MethodName{} produces the highest NMI among all other methods. In Figure~\ref{fig:elbow-num-different-sd}, we show the different NMI results achieved by selecting different $n$ for deciding $k$ adaptively, and choose $n=11$. 

\begin{figure}
    \centering
    \includegraphics[width=\textwidth]{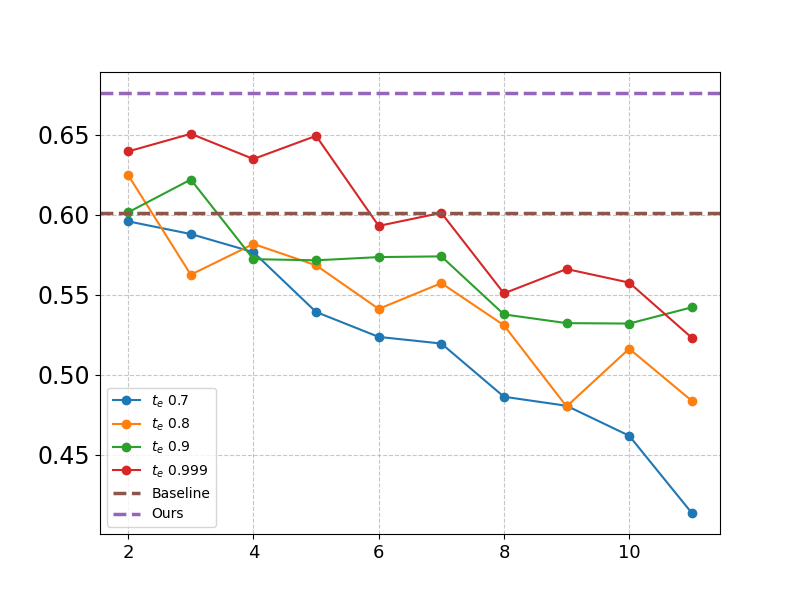}
    \caption{The normalized mutual information (NMI, higher is better) between cluster assignments of Stable Diffusion (SD) features using a Stable Diffusion 2.1 \cite{ldm} and the ground truth labels. \MethodName{} achieves the highest NMI. Baseline is the original SD features.}
    \label{fig:nmi-sd}
\end{figure}

\begin{figure}
    \centering
    \begin{minipage}{0.49\textwidth}
        \centering
        \includegraphics[width=\textwidth,trim={140 220 140 210},clip]{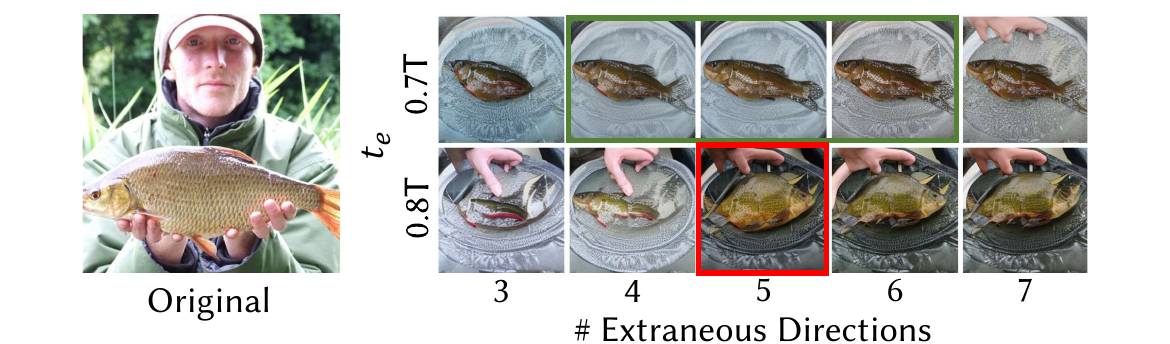}
        \caption{A Failure case of \MethodName{} on selecting $t_e$. Green boxes are the optimal choice, qualitatively. Red boxes are the ones \MethodName{} selects.}
        \label{fig:fail-t}
    \end{minipage}\hfill
    \hspace{1pt}
    \begin{minipage}{0.49\textwidth}
        \centering
        \includegraphics[width=\textwidth,trim={60 220 60 230},clip]{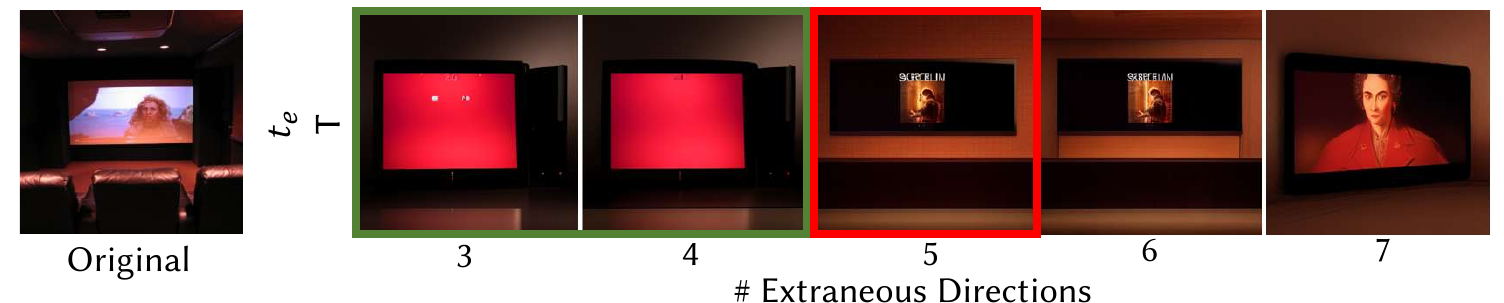}
        \caption{A Failure case of \MethodName{} on selecting $k$ when $n=10$. Green boxes are the optimal choice, qualitatively. Red boxes are the ones \MethodName{} selects.}
        \label{fig:fail-k}
    \end{minipage}
\end{figure}

We identify some promising future directions. First, we fix $t_e$ for all samples, which can be suboptimal on certain images. We only perform \MethodName{} in a single time step instead of selecting multiple $t_e$. A series of projecting away \ExtraDir{}s has the potential of discarding more class-irrelevant information. Regarding the number of \ExtraDir{}s, we only perform experiments on cumulative projection, \textit{i.e.} projecting away the top-$k$ \ExtraDir{}s. A careful selection of the \ExtraDir{}s can contribute to better results. Due to the limits in computational resources, we leave them as future work. Occasionally, \MethodName{} can also select suboptimal $k$, leading to artefacts in the generated images. We show failure cases for $t_e$ and $k$ in Figure~\ref{fig:fail-t}, \ref{fig:fail-k}.

\section{Details of the toy experiment}
\label{sec:supp-toy-detail}
\begin{figure}[htp]
  \centering
  \begin{subfigure}[b]{0.35\textwidth}
    \centering
    \includegraphics[width=\textwidth,trim={70 0 80 0},clip]{figures/toy-data.pdf}
    \caption{}%
    \label{fig:supp-toy-data}
  \end{subfigure}
  \begin{subfigure}[b]{0.35\textwidth}
    \centering
    \includegraphics[width=\textwidth,trim={70 0 80 0},clip]{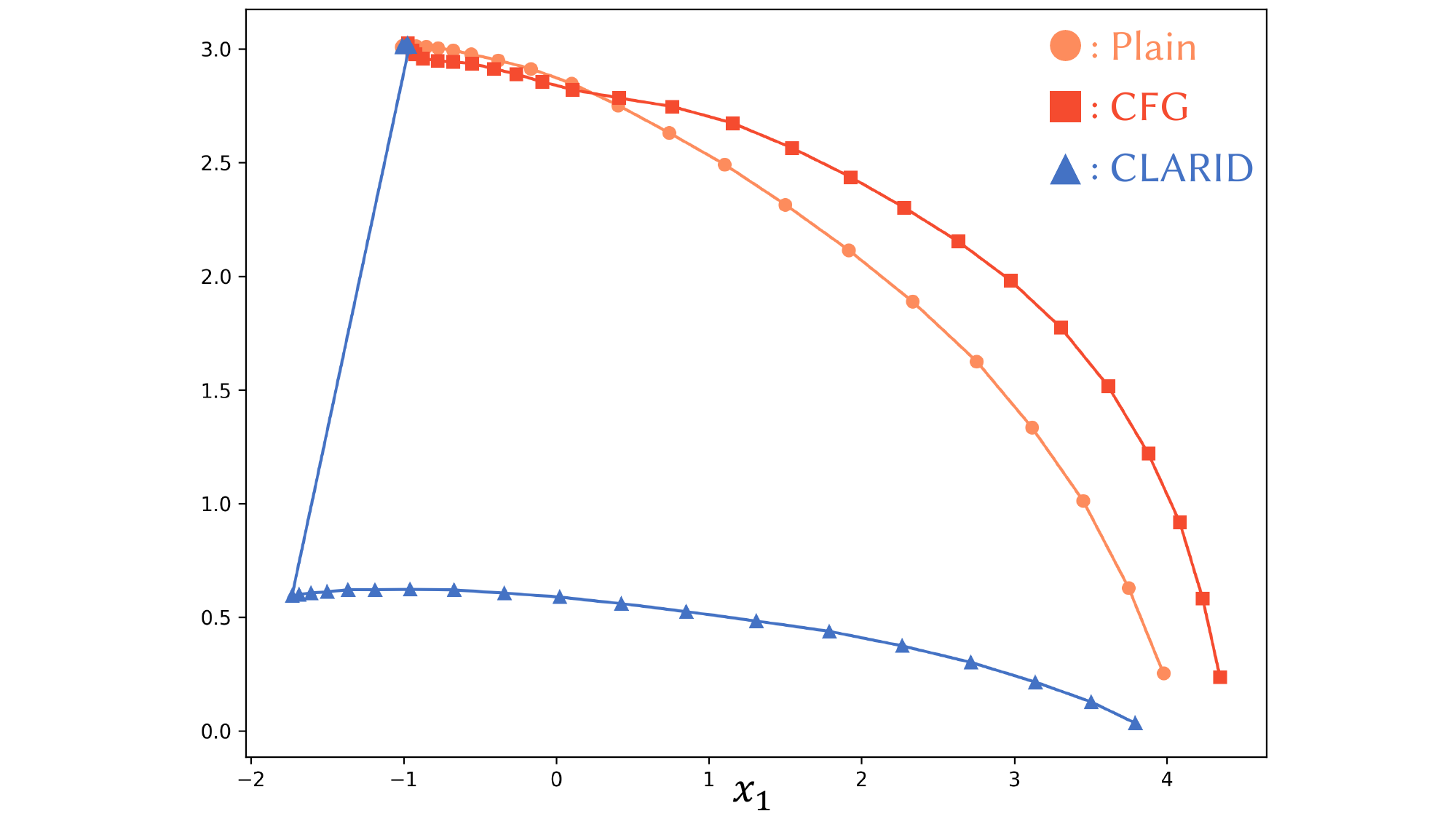}
    \caption{}%
    \label{fig:supp-toy-traj}
  \end{subfigure}

  \begin{subfigure}[b]{0.22\textwidth}
    \centering
    \includegraphics[width=\textwidth]{figures/toy-ori.pdf}
    \caption{}%
    \label{fig:supp-toy-ori}
  \end{subfigure}
  \hspace{5pt}
  \begin{subfigure}[b]{0.22\textwidth}
    \centering
    \includegraphics[width=\textwidth]{figures/toy-cfg.pdf}
    \caption{}%
    \label{fig:supp-toy-cfg}
  \end{subfigure}
  \hspace{5pt}
  \begin{subfigure}[b]{0.22\textwidth}
    \centering
    \includegraphics[width=\textwidth]{figures/toy-jac.pdf}
    \caption{}%
    \label{fig:supp-toy-jac}
  \end{subfigure}
  \caption{
  A toy example of \MethodName{}. \textbf{(a)}: The samples of \textcolor{gray}{class 0} and \textcolor{toyorange}{class 1}; \textbf{(b)}: Sampling trajectory of \textcolor{toyorange}{plain conditioning (Plain)}, \textcolor{toyred}{classifier-free guidance (CFG)}, and \textcolor{toyblue}{\Cano{} (\MethodName{})} starting from \DDIMinv{}($[4.0,0.2]$). \textcolor{toyblue}{\Cano{} (\MethodName{})} orthogonalize the data latent code and class-irrelevant components encoded in the latent code in the CDM, yielding \Canoimg{}s; \textbf{(c,d,e)}: The generated samples of \textcolor{toyorange}{Plain}, \textcolor{red}{CFG}, and \textcolor{toyblue}{\MethodName{}}, respectively. \Cano{}s converge to a 1D manifold inside class 1, capturing the core class information.
  }
  \label{fig:supp-toy-all}
\end{figure}
We adopt a hierarchical generative process described in Eq.~\ref{eq:toy-gen-model} to generate 2D data points from two classes. 
\begin{equation}
\begin{gathered}       
p(\bm{x})
  = p(y)\,p(\bm{x}_{\text{core}}\mid y)\,p(\bm{x}_{\text{var}}\mid \bm{x}_{\text{core}}), \\
Y \sim \operatorname{Bernoulli}\!\bigl(\tfrac12\bigr),\quad
U \sim \mathcal{U}(-0.1,0.1),\quad
\bm\varepsilon=(\varepsilon_x,\varepsilon_y)^{\mathsf T}
               \sim \mathcal N\!\bigl(\bm 0,0.01 \bm{I}_{2}\bigr), \\
\bm{s}(Y)=
  \begin{cases}
    (0,0)^{\mathsf T}, & Y=0,\\
    (4,0)^{\mathsf T}, & Y=1,
  \end{cases}\qquad
\bm{x}_{\text{core}}=(U,0)^{\mathsf T}+\bm{s}(Y),\qquad
\bm{x}_{\text{var}}=\bm\varepsilon, \\
\bm{x}= \bm{x}_{\text{core}} + \bm{x}_{\text{var}}
        + \bigl(3\lvert\varepsilon_y\rvert,0\bigr)^{\mathsf T}.
\end{gathered}
\label{eq:toy-gen-model}
\end{equation}
In our toy model, we have two classes. We simulate the core class information by formulating the $p(\bm{x}_{\text{core}}\mid y)$, so that all data points sampled from $p(\bm{x}_{\text{core}}\mid \text{class} 1)$ lie on the line $L=\{ (x_1,0)|3.9 \le x_1 \le 4.1 \}$. We design the class-irrelevant variations to be $\bm\varepsilon$, sampled from a Gaussian distribution. We shift the $x$-axis value according to the $y$-axis variation to increase the complexity of the distribution. We design a simple 3-layer multilayer-perceptron-based (MLP-based) diffusion model to model this 2D distribution. We set the hidden dimensionality to 80, and the dimensionality for both label and time step is 16. We use sinusoidal embeddings for uniquely encoding the 1000 time steps \cite{vit}. We train the diffusion model using the standard DDPM loss \cite{ddpm}. The hyperparameters of training are given in Table~\ref{tab:param-toy}. The resulting data points of \MethodName{}, \textit{i.e.} \Cano{}s, lie exactly on the line $L$. This is the reason why we claim that the underlying structure revealed by \Cano{}s corresponds to one of the true generative processes for the observed data. 
\begin{table}[]
    \centering
    \caption{The hyperparameters used in our toy experiment in Section~\ref{sec:cafol-toy}. }
    \label{tab:param-toy}
    \begin{tabular}{c|c|c|c|c|c|c}
      \# data points & Epoch & Optimizer & Batch size & Learning rate & Weight decay & Label drop rate \\
      \midrule
       1000  & 1000 & Adam \cite{adam} & 128 & 1e-3 & 0 & 0.1
    \end{tabular}
\end{table}

\section{Training and evaluation details}
\label{sec:supp-training-detail}

We perform all the experiments using the PyTorch platform. We provide the training hyperparameters in Table~\ref{tab:param-main}. The temperature parameter $\tau$ in \textbf{\textit{\CanoDistill{}}} is fixed to $0.1$ in all cases, as in \cite{supcon}. On CIFAR10, we use random crop (\verb|torchvision.transforms.RandomCrop(32,padding=4)|) and random horizontal flip (\verb|torchvision.transforms.RandomHorizontalFlip|) for data augmentation. Note that the SupCon \cite{supcon} baseline uses a different data augmentation strategy and is trained for much longer epochs (1000). We follow the official code implementation (\href{https://github.com/HobbitLong/SupContrast}{link}) to reproduce this baseline on CIFAR10. On ImageNet and ImageNet100, we adopt the data augmentation used in \citet{supcon}, to improve the generalization performance of the trained model so that the baselines have meaningful results on the used generalization benchmarks. For example, the ResNet50 trained with random resized crop and random horizontal flip, as in \citet{resnet}, will have 0\% accuracy on ImageNet-A \cite{imagenet-a}, while the baseline ResNet50 (Vanilla) in our experiments achieves 6.3\%. The difference in data augmentation results in the performance difference on the ImageNet validation set between our baseline model and the one trained in \citet{resnet}. However, we are not focusing on the clean performance in our settings. We use a single Nvidia A100 GPU for the CIFAR10 experiments, and four A100 GPUs for the ImageNet and ImageNet100 experiments. Training one model on CIFAR10 takes around 0.5 $\sim$ 1 hour. Training on ImageNet takes around 40 hours for a model.

\begin{table}[]
    \centering
    \caption{The hyperparameters used in our experiments. We train a ResNet18 \cite{resnet} on CIFAR10 and a ResNet50 on ImageNet. SGDM is SGD with momentum=0.9. \textbf{\textit{\CanoDistill{}}} is effective with different training settings, as shown in Section~\ref{sec:supp-different-aug} and \ref{sec:generalization-swin}. }
    \label{tab:param-main}
    \resizebox{1.0\linewidth}{!}{
    \begin{tabular}{c|c|c|c|c|c|c|c}
      $\mathcal{D}$ & Epoch & Optimizer & Batch size & Learning rate & Weight decay & LR scheduler & LR decay rate \\
      \midrule
      CIFAR10 & 200 & SGDM & 128 & 0.1 & 5e-4 & Step 100,150 & 0.1 \\
      ImageNet100 & 100 & SGDM & 256 & 0.1 & 1e-4 & Cosine & / \\
      ImageNet & 100 & SGDM & 512 & 0.1 & 1e-4 & Cosine & / 
    \end{tabular}
    }
\end{table}

Our \CanoDistill{} involves the usage of CDM features, \textit{i.e.}, \Canofeat{}s. Instead of forwarding the CDM during training, which will lead to a large computational cost, we pre-compute the \Canofeat{}s and load them during training, contributing to an efficient training pipeline. All \Canofeat{}s are 1D vectors, which are the results of performing average pooling on the original feature map, as done in a previous work \cite{dm-as-representation-learner}.

We examine the adversarial robustness of different student models using four adversarial attacks, PGD \cite{pgd}, CW \cite{cw}, APGD-DLR \cite{autoattack}, and APGD-CE \cite{autoattack}. The detailed settings are given in Table~\ref{tab:attack-cifar10}, \ref{tab:attack-imagenet},and \ref{tab:attack-imagenet100} for CIFAR10, ImageNet, and ImageNet100 (AutoAttack), respectively. We choose PGD \cite{pgd} since it is the most popular method for examining adversarial robustness \cite{adanca}. Moreover, we want to examine whether the drawback of cross-entropy loss can lead to false robustness \cite{autoattack}. The step size in PGD can also largely affect the result. Hence, we choose the Auto-PGD family \cite{autoattack} to automatically decide the step size and incorporate the new Difference of Logits Ratio (DLR) loss, resulting in APGD-CE and APGD-DLR, respectively. We also want to include an optimization-based adversarial attack and thus select the CW \cite{cw} attack. The hyperparameters of different attacks are chosen to ensure a meaningful comparison between different models, avoiding the case in which all models have 0\% accuracy after the attack. Our choices ensure a thorough test of the adversarial robustness in a white-box setting, revealing the multifacetedness of the adversarial robustness. Information about the generalization benchmarks is given in Section~\ref{sec:dataset-info}. 

\begin{table}[]
    \caption{Hyperparameters in different adversarial attacks on CIFAR10.\\}
    \label{tab:attack-cifar10}
    \centering
    \begin{tabular}{c|cccc}
        \toprule
                      & PGD \cite{pgd} & CW \cite{cw} & APGD-DLR \cite{autoattack} & APGD-CE \cite{autoattack}  \\
        \midrule
        Max magnitude & 2.0/255                & /               & 2.0/255                 & 2.0/255 \\
        Steps         & 5                  & 5                & 5                   & 5   \\
        Step size         & 0.5                  & /                & /                   & /   \\
        $\kappa$         & /                   & 0.0                 & /                   & /   \\
        $c$         & /                   & 0.2                & /                   & /   \\
        $\rho$         & /                   & /                 & 0.75                   & 0.75   \\
        EOT         & /                   & /                 & 1                   & 1   \\
        \bottomrule
    \end{tabular}
\end{table}

\begin{table}[]
    \caption{Hyperparameters in different adversarial attacks on ImageNet.\\}
    \label{tab:attack-imagenet}
    \centering
    \begin{tabular}{c|cccc}
        \toprule
                      & PGD \cite{pgd} & CW \cite{cw} & APGD-DLR \cite{autoattack} & APGD-CE \cite{autoattack}  \\
        \midrule
        Max magnitude & 0.33/255                & /               & 0.33/255                 & 0.33/255 \\
        Steps         & 5                  & 5                & 5                   & 5   \\
        Step size         & 0.5                  & /                & /                   & /   \\
        $\kappa$         & /                   & 0.0                 & /                   & /   \\
        $c$         & /                   & 0.1               & /                   & /   \\
        $\rho$         & /                   & /                 & 0.75                   & 0.75   \\
        EOT         & /                   & /                 & 1                   & 1   \\
        \bottomrule
    \end{tabular}
\end{table}

\begin{table}[]
    \caption{Hyperparameters of all attacks used in AutoAttack \cite{autoattack} in our ablation studies on ImageNet100.}
    \label{tab:attack-imagenet100}
    \centering
    \begin{tabular}{c|cccc}
        \toprule
                      & PGD \cite{pgd} & CW \cite{cw} & APGD-DLR \cite{autoattack} & APGD-CE \cite{autoattack}  \\
        \midrule
        Max magnitude & 0.5/255                & /               & 0.5/255                 & 0.5/255 \\
        Steps         & 5                  & 10                & 5                   & 5   \\
        Step size         & 0.5                  & /                & /                   & /   \\
        $\kappa$         & /                   & 0.0                 & /                   & /   \\
        $c$         & /                   & 0.1               & /                   & /   \\
        $\rho$         & /                   & /                 & 0.75                   & 0.75   \\
        EOT         & /                   & /                 & 1                   & 1   \\
        \bottomrule
    \end{tabular}
\end{table}

Due to the limits in computational resources, we do not report error bars in our experiments. During testing, we find that different runs of adversarial attacks result in similar performance. We test each adversarial attack 3 times with different seeds and find that the resulting performance has standard deviations all smaller than 0.05. 

\section{More results on ImageNet}

\subsection{The performance of $\mathcal{L}_{dist}$ alone in feature distillation}
\label{sec:supp-ldist-hint-rkd-cka}
In Section~\ref{sec:cano-quant-result}, we design a baseline experiment that distills the structure of the raw diffusion features into the representation space of the student network, which is based on $\mathcal{L}_{dist}$. Our design of $\mathcal{L}_{dist}$ differs from all previous works on diffusion-based feature distillation. We use a CKA \cite{cka} metric for measuring the linear subspace alignment between the feature vectors of the student and the teacher. CKA is invariant to isotropic scaling as well as orthonormal transformations. Such an invariance lets us transfer the class-discriminative structure encoded in \Canofeat{}s without over-constraining the student’s own feature basis. Previous works focus on using three classical feature distillation losses: (1) FitNet \cite{dreamteacher,fitnet-hint,dm-as-representation-learner}, which is the L2 distance between student features and the teacher's; (2) Attention transfer (AT) \cite{dreamteacher,dm-as-representation-learner,attention-transfer}, which distills the saliency structure of the activation map to the student; (3) Relational knowledge distillation (RKD) \cite{rkd,dm-as-representation-learner}, which aligns the relational representations of the samples between the teacher and the student. However, in our case, these loss functions do not significantly contribute to the student's performance, as shown in Table~\ref{tab:l-dist-abl-other-loss}. Note that in these experiments, the student features and the teacher ones are one-to-one matching to mimic the typical feature distillation framework, instead of the random strategy as we designed in \textbf{\textit{\CanoDistill{}}}.

\begin{table}[]
    \centering
    \caption{Comparison between our CKA-based \cite{cka} $\mathcal{L}_{dist}$ and other loss functions in diffusion-based feature distillation.}
    \label{tab:l-dist-abl-other-loss}
    \begin{tabular}{c|cc}
       $\mathcal{L}_{cano}$  & Clean & AutoAttack \cite{autoattack} \\
       \midrule
       Vanilla \cite{resnet}  & 86.5 & 15.9 \\
       \midrule
       FitNet \cite{dreamteacher,fitnet-hint,dm-as-representation-learner}  & 86.7 & 16.4 \\
       AT \cite{dreamteacher,dm-as-representation-learner,attention-transfer}  & 86.6 & 16.3 \\
       RKD \cite{rkd,dm-as-representation-learner}  & 86.2 & 16.4 \\
       \rowcolor{gray!20} Ours & \textbf{87.3} & \textbf{18.8}
    \end{tabular}
\end{table}
Our CKA-based \cite{cka} feature distillation loss outperforms all previous designs without introducing additional parameters during training. This is a novel loss function used in a diffusion-based feature distillation framework, inspired by previous works \cite{it-align,cka-distill2,cka-distill1}. Performing knowledge transfer with the teacher and/or the student being a ViT is still an open question \cite{vit-vit-distill-bad,cnn-vit-distill-bad}, and can lead to a performance drop in the student. Our $\mathcal{L}_{dist}$, however, achieves good performance in the diffusion-based settings.

\subsection{\textbf{\textit{\CanoDistill{}}} is effective with different data augmentation strategies}
\label{sec:supp-different-aug}
We show that \textbf{\textit{\CanoDistill{}}} is effective when the data augmentation strategy is different, demonstrating the generalization of the paradigm. Specifically, the training lasts 120 epochs, and the data augmentations are: 
\begin{itemize}
    \item \verb|torchvision.transforms.RandomResizedCrop(224)|,
    \item \verb|torchvision.transforms.RandomHorizontalFlip()|,
    \item \verb|torchvision.transforms.ColorJitter(0.3, 0.3, 0.3)|,
\end{itemize}
The learning rate is 0.2, and the decay happens every 30 epochs with a decay rate of 0.1. The results are in Table~\ref{tab:imagenet-quant-different-aug}. \textbf{\textit{\CanoDistill{}}} yields a student that outperforms the vanilla model on all benchmarks, proving the effectiveness of our method in this case and implying its generalization capability.

\begin{table}[]
    \centering
    \caption{Quantitative comparisons between \textbf{\textit{\CanoDistill{}}} and baselines on ImageNet \cite{deng2009imagenet} with a ResNet50 \cite{resnet}, using a different training setting (Section~\ref{sec:supp-different-aug}) from the one in Section~\ref{sec:supp-training-detail}. Higher is better. }
    \label{tab:imagenet-quant-different-aug}
    \resizebox{\linewidth}{!}{
    \begin{tabular}{c|c|c|cccc|cccc}
       Model & $\text{Data}_{\text{DM}}$ & Clean & PGD \cite{pgd} & CW \cite{cw} & APGD-DLR \cite{autoattack} & APGD-CE \cite{autoattack} & IM-C \cite{imagenet-c} & IM-A \cite{imagenet-a}  & IM-ReaL \cite{im-real}  \\
       \midrule
        Vanilla & / & 76.6 & 17.3 & 13.5 & 18.8 & 17.7 & 40.6 & 3.4 & 83.3 \\
        \rowcolor{gray!20} \textit{\textbf{\CanoDistill{}}} & 10\% &  76.7 &  \textbf{21.3} & \textbf{21.3} & \textbf{22.6} & \textbf{21.3} &\textbf{41.2} & \textbf{4.2} & \textbf{83.3} 
    \end{tabular}
    }
\end{table}

\subsection{\textbf{\textit{\CanoDistill{}}} improves the student's black-box adversarial robustness}
\label{sec:supp-blackbox}
In Table~\ref{tab:main-quant}, we demonstrate that \textbf{\textit{\CanoDistill{}}} improves the student's white-box robustness. Here, we show that \textbf{\textit{\CanoDistill{}}} improves the student performance when facing black-box adversarial attacks. Specifically, we test all models on ImageNet using the Square attack \cite{square}, which is a black-box adversarial attack algorithm. We use $L_{inf}$ metric with attacking budget $4.0/255.0$, and the query number is 1000. To reduce computational costs, we randomly select 2000 samples from ImageNet to perform the evaluation. The result is given in Table~\ref{tab:black-box-imagenet}.

\begin{table}[]
    \centering
    \caption{Quantitative comparisons between \textbf{\textit{\CanoDistill{}}}, and baselines on ImageNet \cite{deng2009imagenet} with a ResNet50 \cite{resnet}, on black-box adversarial robustness. Higher is better. \textcolor{venetianred}{Red} is lower than the vanilla model. $\text{Data}_{\text{DM}}$: the portion of the subset on which the diffusion model serves as the teacher. \DMFit{}: Feature distillation by $\mathcal{L}_{dist}$ on the whole dataset using a DiT model; \DMDistill{}: Using the framework of \textbf{\textit{\CanoDistill{}}}, but replace \Cano{}s by samples with CFG from the CDM. }
    \label{tab:black-box-imagenet}
    \begin{tabular}{c|c|c|c}
       Model & $\text{Data}_{\text{DM}}$ & Clean & Square \cite{square} \\
       \midrule
        Vanilla & / & 75.9 & 23.5 \\
        \midrule
        \midrule
        DiffAug \cite{diffaug} & 100\% & 76.0 & \textcolor{venetianred}{20.7} \\
        \DMFit{} & 100\% & \textcolor{venetianred}{75.7} & \textcolor{venetianred}{22.9}\\
        \midrule
        \midrule
        \DMDistill{} & 10\% & 75.7 & \textcolor{venetianred}{23.3}  \\
        \rowcolor{gray!20} \textit{\textbf{\CanoDistill{}}} & 10\% &  75.9 &  \textbf{25.6} 
    \end{tabular}
\end{table}


\subsection{Extracting class semantics using class tokens in vision transformer}
\label{sec:supp-vit-cls-token-compare}
Our main claim on the application of \Cano{}s is that they distill the core class semantics of each category. Here, we investigate another mainstream approach to distill such information, which is the class token in vision transformer (ViT) \cite{vit,deit,deit3}. The class token performs the attention operation \cite{nlp-transformer} to all spatial tokens, collecting the discriminative signals inside the feature map for classification. We adopt a challenging baseline network, DeiT-III-Huge \cite{deit3}, which is a ViT model solely trained on ImageNet with the number of parameters (DeiT: 632.1M; DiT: 675M) and FLOPS (DeiT: 167.4G; DiT: 118.6G) matching the DiT \cite{dit} used in our experiments. It is challenging because the DeiT-III-Huge model is trained with advanced data augmentation techniques and performs well on ImageNet classification tasks \cite{deit3}, whereas DiT is trained with a plain horizontal flip augmentation and is not good at classification \cite{diff-classifier-first} (85.2 v.s. 77.5 Top-1 accuracy). We use our CKA-based $\mathcal{L}_{dist}$ to align the representations of the student network to the class token in DeiT-III-Huge, termed $\text{DeiT}_{dist}$. All the settings are the same as used in Section~\ref{sec:cano-quant-result} and \ref{sec:supp-training-detail}. The results are given in Table~\ref{tab:deit-compare-imagenet-quant}. Notably, the student trained with \textbf{\textit{\CanoDistill{}}} outperforms the one trained with $\text{DeiT}_{dist}$ in terms of clean accuracy and generalization. Achieving good performance in feature distillation between ViT and CNNs is still an open problem in the field \cite{cnn-vit-distill-bad}. Despite this, our experiments control the architecture (both teachers are ViTs), the number of parameters, and the FLOPS. Moreover, the $\text{DeiT}_{dist}$ can yield a student that outperforms \DMFit{} in terms of all adversarial attack benchmarks, which demonstrates the effectiveness of the method and validity of our experiments. We believe investigating the difference between the mechanisms of how discriminative models and generative ones encode class information is an interesting future direction. 

\begin{table}[]
    \centering
    \caption{Quantitative comparisons between \textbf{\textit{\CanoDistill{}}}, and baselines on ImageNet \cite{deng2009imagenet} with a ResNet50 \cite{resnet}. Higher is better. \textcolor{venetianred}{Red} is lower than the vanilla model. $\text{Data}_{\text{DM}}$: the portion of the subset on which the diffusion model serves as the teacher. $\text{DeiT}_{dist}$: Feature distillation by $\mathcal{L}_{dist}$ on the whole dataset using the class token in an ImageNet-pretrained DeiT-III-Huge model \cite{deit3}; \DMFit{}: Feature distillation by $\mathcal{L}_{dist}$ on the whole dataset using a DiT model; \DMDistill{}: Using the framework of \textbf{\textit{\CanoDistill{}}}, but replace \Cano{}s by samples with CFG from the CDM.}
    \label{tab:deit-compare-imagenet-quant}
    \resizebox{\linewidth}{!}{
    \begin{tabular}{c|c|c|cccc|cccc}
       Model & $\text{Data}_{\text{DM}}$ & Clean & PGD \cite{pgd} & CW \cite{cw} & APGD-DLR \cite{autoattack} & APGD-CE \cite{autoattack} & IM-C \cite{imagenet-c} & IM-A \cite{imagenet-a}  & IM-ReaL \cite{im-real}  \\
       \midrule
        Vanilla & / & 75.9 & 15.6 & 13.7 & 17.2 & 16.7 & 45.9 & 6.3 & 82.8 \\
        \midrule
        \midrule
        $\text{DeiT}_{dist}$ \cite{deit3} & 100\% & \textcolor{venetianred}{75.1} & 19.7 & 18.4 & 20.8 & 20.3 & \textcolor{venetianred}{45.2} & \textcolor{venetianred}{5.4} & \textcolor{venetianred}{82.5} \\
        \DMFit{}                & 100\% & \textcolor{venetianred}{75.7} & 15.7 & 14.1 & \textcolor{venetianred}{17.0} & 16.7 & \textcolor{venetianred}{43.6} & \textcolor{venetianred}{5.0} & 82.8\\
        \midrule
        \midrule
        \DMDistill{} & 10\% & 75.7 & 20.8 & 20.3 & 20.8 & 21.4 & \textcolor{venetianred}{45.6} & \textcolor{venetianred}{6.0} & \textcolor{venetianred}{82.7} \\
        \rowcolor{gray!20} \textit{\textbf{\CanoDistill{}}} & 10\% &  75.9 &  \textbf{21.9} & \textbf{21.7} & \textbf{22.5} & \textbf{22.3} &\textbf{46.1} & \textbf{6.7} & \textbf{83.1} 
    \end{tabular}
    }
\end{table}

\subsection{Details of the Background Challenge}
\label{sec:supp-inbg}
\begin{figure}[htp]
  \centering
  \begin{subfigure}[b]{0.23\textwidth}
    \centering
    \includegraphics[width=\textwidth]{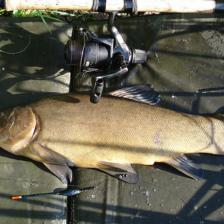}
    \caption{Original.}%
    \label{fig:bg-ori-fish}
  \end{subfigure}
  \begin{subfigure}[b]{0.23\textwidth}
    \centering
    \includegraphics[width=\textwidth]{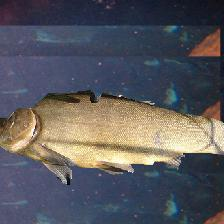}
    \caption{BG-Same.}%
    \label{fig:bg-mix-same-fish}
  \end{subfigure}
  \begin{subfigure}[b]{0.23\textwidth}
    \centering
    \includegraphics[width=\textwidth]{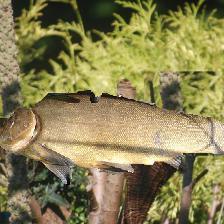}
    \caption{BG-Rand.}%
    \label{fig:bg-mix-rand-fish}
  \end{subfigure}
  \begin{subfigure}[b]{0.23\textwidth}
    \centering
    \includegraphics[width=\textwidth]{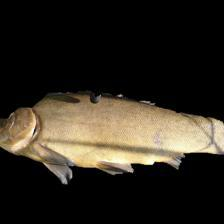}
    \caption{Only-FG.}%
    \label{fig:bg-only-fg-fish}
  \end{subfigure}
  \caption{Samples from the Backgrounds Challenge \cite{inbg-challenge}. (a) Original: The original image. (b) BG-Same: Put a random background from the same class onto the image. (c) BG-Rand: Put a random background from a different class onto the image. (d) Only-FG: Discard the background and make it black.}
  \label{fig:bg-vis}
\end{figure}

In Section \ref{sec:cano-quant-result}, we test the student model on the Backgrounds Challenge \cite{inbg-challenge}. Figure \ref{fig:bg-vis} illustrates its three variants: BG-Same re-uses a single background per class; BG-Rand pairs each foreground with randomly chosen backgrounds from other classes; Only-FG removes the background entirely. These operations preserve the foreground object while removing background cues. Hence, the performance in this test quantifies a model’s ability to rely on true class signals rather than spurious background correlations.

As shown in Table \ref{tab:imagenet-inbg}, \textbf{\textit{\CanoDistill{}}} achieves the highest accuracy across all splits. \DMDistill{} matches \textbf{\textit{\CanoDistill{}}} on the Original and BG-Same sets, but \textbf{\textit{\CanoDistill{}}} outperforms it on BG-Rand and Only-FG. This gap indicates that the student trained with \DMDistill{} still uses background information shared within each class; when those backgrounds are shuffled or removed, its accuracy declines. The evidence suggests that CFG introduces label-correlated yet non-essential background signals into the training data, whereas \textbf{\textit{\CanoDistill{}}} suppresses those signals and encourages the student to focus on the foreground object.


\subsection{On the reproduction of RepFusion}
\label{sec:supp-repfusion-detail}
RepFusion \cite{dm-as-representation-learner} proposes a novel framework for diffusion-based feature distillation. It uses a neural network for adaptively selecting the time step of feature extraction from the teacher DM. This neural network is trained using the REINFORCE \cite{REINFORCE} algorithm, using the task performance as the reward. The task performance is the classification accuracy. In this case, the neural network is non-linear and can directly decode the label conditioning in the CDM to maximize the reward, leading to a failed training. Hence, we reproduce the method on an unconditional DM. Despite this, we provide a strong baseline using CDM, \DMFit{}, which uses a feature distillation loss that can outperform all the loss functions used in RepFusion, as shown in Section~\ref{sec:supp-ldist-hint-rkd-cka}. 

\subsection{Ablation studies}
\label{sec:supp-ablation}
We conduct ablation studies on ImageNet100, a 100-class subset of ImageNet. Previous studies \cite{multilinear-network,imagenet100-3,adanca,imagenet100-1,imagenet100-2} have shown that ImageNet100 serves as a representative subset of ImageNet1K. Hence, we can obtain representative results for the self-evaluation of the model while efficiently using our computational resources. All the ablations are based on a ResNet50 \cite{resnet} model. Note that we still use an ImageNet-pretrained DiT \cite{dit} as the teacher. We report the test accuracy on the ImageNet100 validation set as the clean accuracy (\textcolor{customblue}{Clean}), and the adversarial accuracy (\textcolor{customorange}{Adv}) under AutoAttack \cite{autoattack} that consists of PGD \cite{pgd}, CW \cite{cw}, APGD-DLR \cite{autoattack}, and APGD-CE \cite{autoattack}. 
\begin{figure}
    \centering
    \begin{minipage}{0.49\textwidth}
        \centering
        \includegraphics[width=\textwidth]{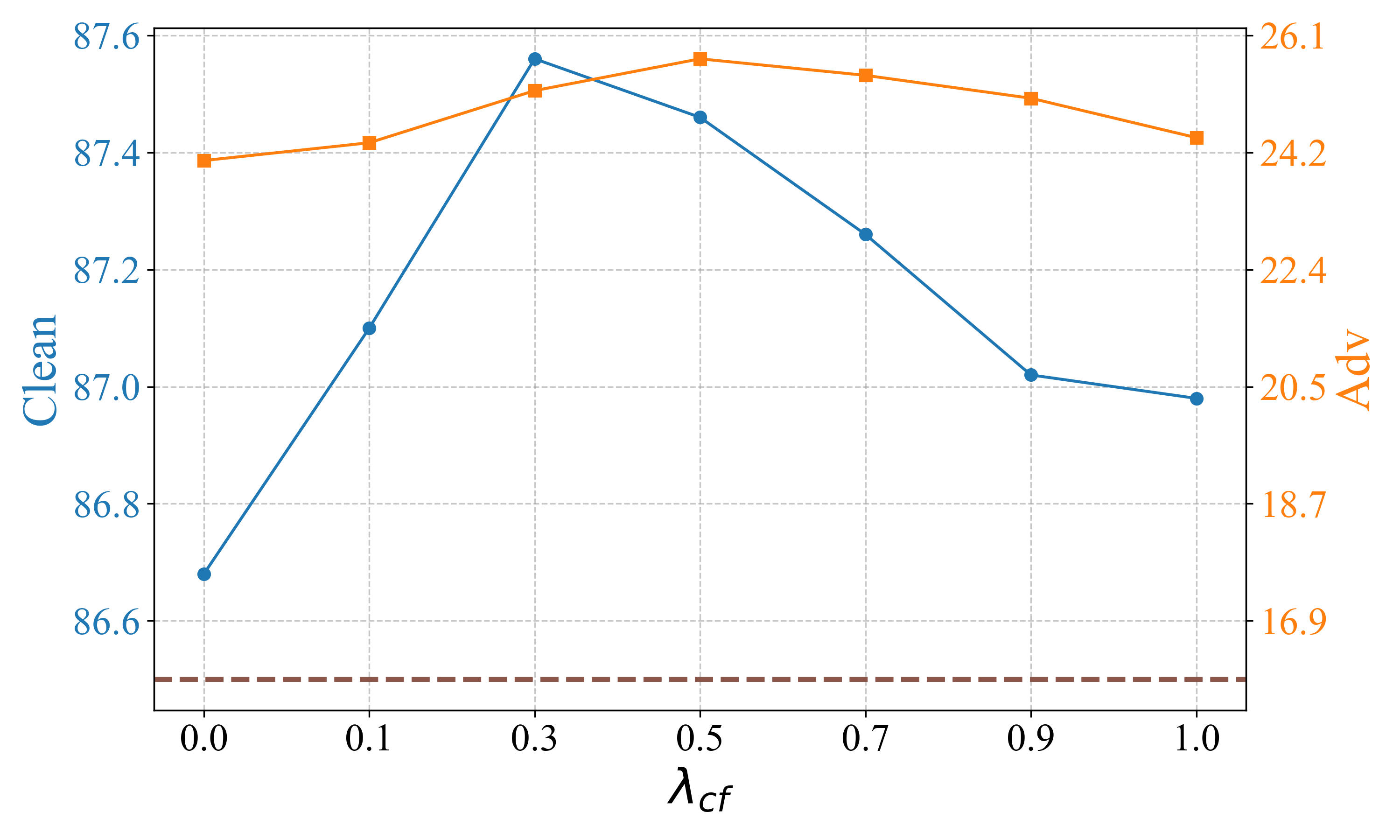}
        \caption{Ablation on $\lambda_{cf}$. An effective training requires a trade-off between $\mathcal{L}_{align}$ and $\mathcal{L}_{cano}$, and necessitates both of them. We choose $\lambda_{cf}=0.5$ to balance between the \textcolor{customblue}{Clean} accuracy and \textcolor{customorange}{robustness}. \textcolor{custombrown}{Brown} is the baseline of both \textcolor{customblue}{Clean} and \textcolor{customorange}{Adv}.}
        \label{fig:proto-cf}
    \end{minipage}\hfill
    \hspace{1pt}
    \begin{minipage}{0.49\textwidth}
        \centering
        \includegraphics[width=\textwidth]{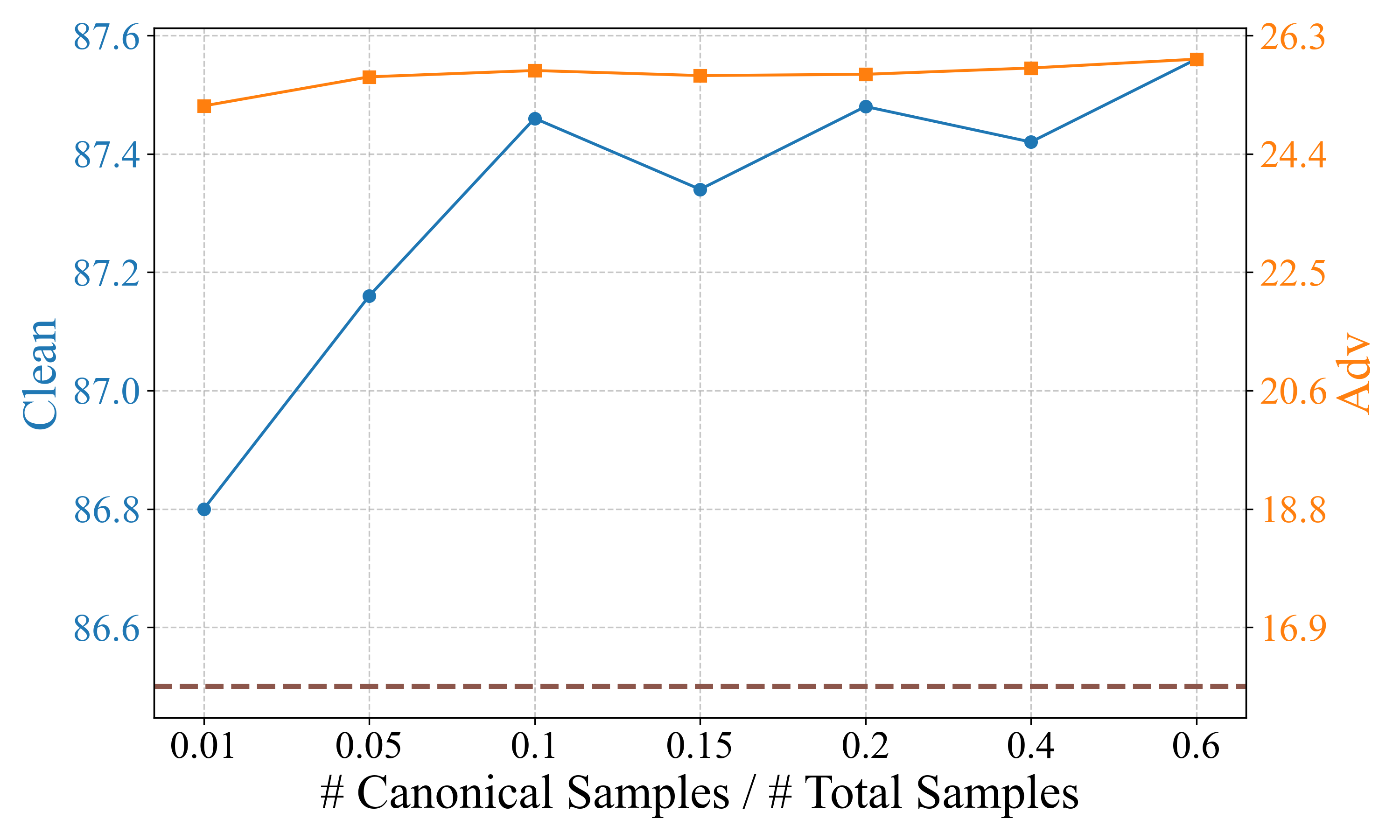}
        \caption{Ablation on the number of \Cano{}s.  A small amount of \Cano{}s, \textit{e.g.} 10\%, is sufficient for achieving competitive performance. It implies the low-dimensionality property of the class manifolds inside CDMs, which is in line with previous findings \cite{diffusion-low-dim}. }
        \label{fig:trunc}
    \end{minipage}
\end{figure}

\subsubsection{Number of \Cano{}s}
\label{sec:ablation-num-cano}
Figure~\ref{fig:trunc} shows the student performance when trained with different numbers of \Cano{}s. Remarkably, training with as little as 10\% of the available \Cano{}s already yields near‑optimal performance. The result suggests that a small data subset is enough to capture the core class semantics in CDMs, because those semantics lie on a low‑dimensional manifold, which is consistent with earlier findings \cite{diffusion-low-dim}.

\subsubsection{The necessity of $\mathcal{L}_{align}$ and $\mathcal{L}_{cano}$ and their balance}
\label{sec:ablation-align-cano}
Our design includes on two complementary objectives: the alignment loss, $\mathcal{L}_{align}$, which pulls each sample towards \Cano{}s from its class, and the \Cano{} separation loss, $\mathcal{L}_{cano}$, which drives the \Cano{}s of different classes apart. Figure~\ref{fig:proto-cf} demonstrates the trade-off between the two. If $\mathcal{L}_{align}$ is omitted ($\lambda_{cf} = 0$), samples remain distant from their canonical counterparts, preventing the student from learning the core semantics of each class. Conversely, dropping $\mathcal{L}_{cano}$ ($\lambda_{cf} = 1$) can lead to \Cano{}s that collapse together, leaving the student unable to discriminate between categories. Therefore, the optimal choice requires a balance between them.

\subsubsection{The necessity of $\mathcal{L}_{dist}$}
\
The CDM transfers the core class features using \Canofeat{}s via $\mathcal{L}_{dist}$. Without this loss, the student can fail to learn the encoded features of the \Canoimg{}s, which can negatively affect the student's clean accuracy and adversarial robustness, as shown in Table~\ref{tab:supp-imagenet-quant-ldist}. 

\begin{table}[]
    \centering
    \caption{Quantitative comparisons between \textbf{\textit{\CanoDistill{}}} and baselines on ImageNet \cite{deng2009imagenet} with a ResNet50 \cite{resnet}. Vanilla: The original student network. $\text{Data}_{\text{DM}}$: the portion of the subset on which the diffusion model serves as the teacher. \DMFit{}: Feature distillation by $\mathcal{L}_{dist}$ on the whole dataset; \DMDistill{}: Using the framework of \textbf{\textit{\CanoDistill{}}}, but replace \Canoimg{}s by samples generated with CFG after \DDIMinv{}, and use their corresponding features in the CDM. Higher is better. \textcolor{venetianred}{Green} is lower than the vanilla model. Without $\mathcal{L}_{dist}$, the student cannot learn the teacher's encoding of \Cano{}s, limiting the student's adversarial robustness.}
    \label{tab:supp-imagenet-quant-ldist}
    \resizebox{\linewidth}{!}{
    \begin{tabular}{c|c|c|cccc}
       Model & $\text{Data}_{\text{DM}}$ & Clean & PGD \cite{pgd} & CW \cite{cw} & APGD-DLR \cite{autoattack} & APGD-CE \cite{autoattack} \\
       \midrule
        Vanilla & / & 75.9 & 15.6 & 13.7 & 17.2 & 16.7 \\
        \midrule
        \midrule
        DiffAug \cite{diffaug} & 100\% & 76.0 & 15.9 & \textcolor{venetianred}{13.1} & 17.2 & 17.0 \\
        \DMFit{}                & 100\% & \textcolor{venetianred}{75.7} & 15.7 & 14.1 & \textcolor{venetianred}{17.0} & 16.7 \\
        \midrule
        \midrule
        \DMDistill{} & 10\% & 75.7 & 20.8 & 20.3 & 20.8 & 21.4 \\
        \rowcolor{gray!20} \textit{\textbf{\CanoDistill{}}} & 10\% &  75.9 &  \textbf{21.9} & \textbf{21.7} & \textbf{22.5} & \textbf{22.3} \\
        \midrule
        No $\mathcal{L}_{dist}$ & 10\% & 75.6 & 20.3 & 19.3 & 20.5 & 21.9 \\
    \end{tabular}
    }
\end{table}

\subsubsection{The weights of losses, $\lambda_{cs},\lambda_{dist},\lambda_{cka}$}
\begin{figure}[htp]
  \centering
  \begin{subfigure}[b]{0.32\textwidth}
    \centering
    \includegraphics[width=\textwidth]{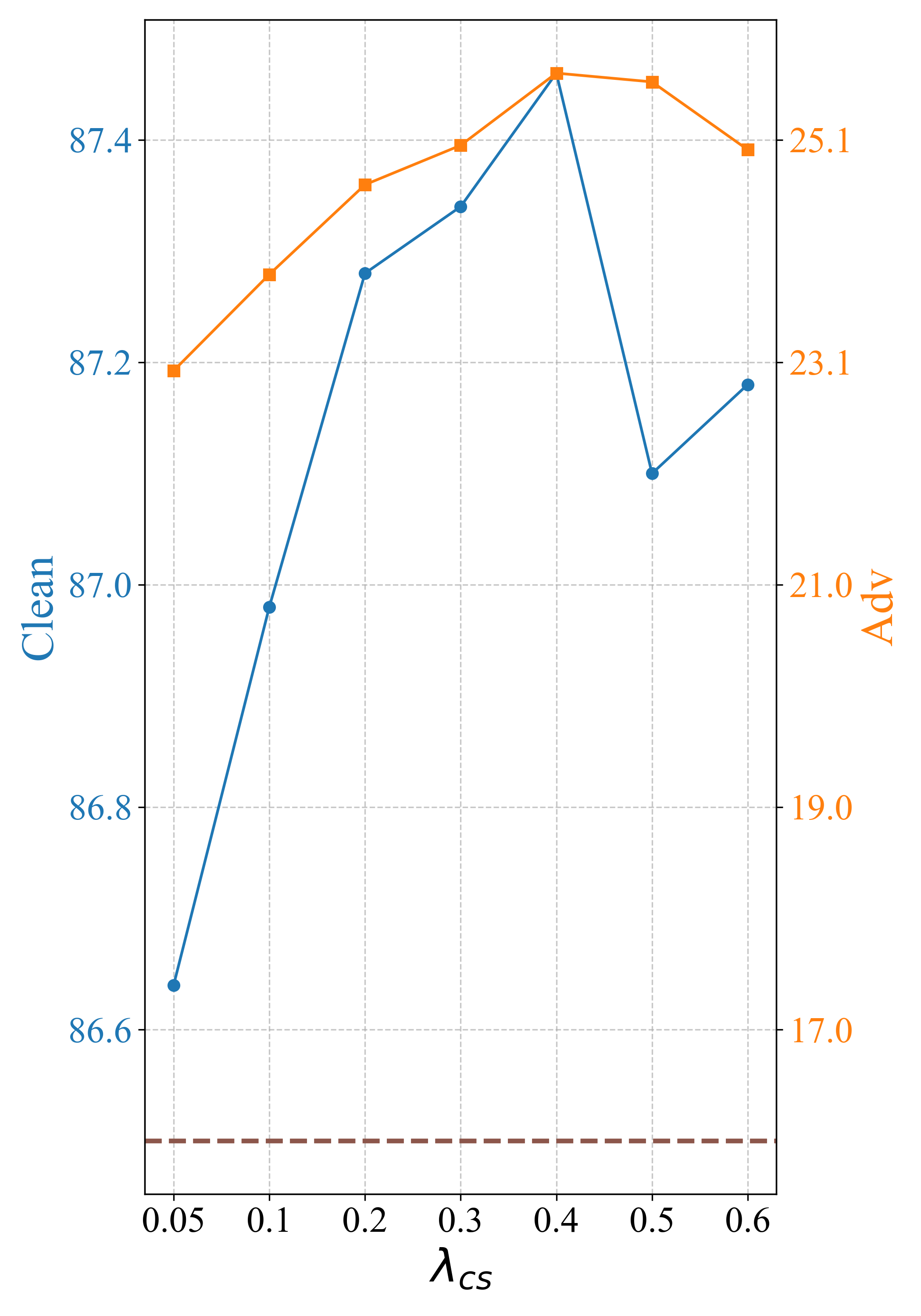}
    \caption{}%
    \label{fig:l-aug-abl}
  \end{subfigure}
  \begin{subfigure}[b]{0.32\textwidth}
    \centering
    \includegraphics[width=\textwidth]{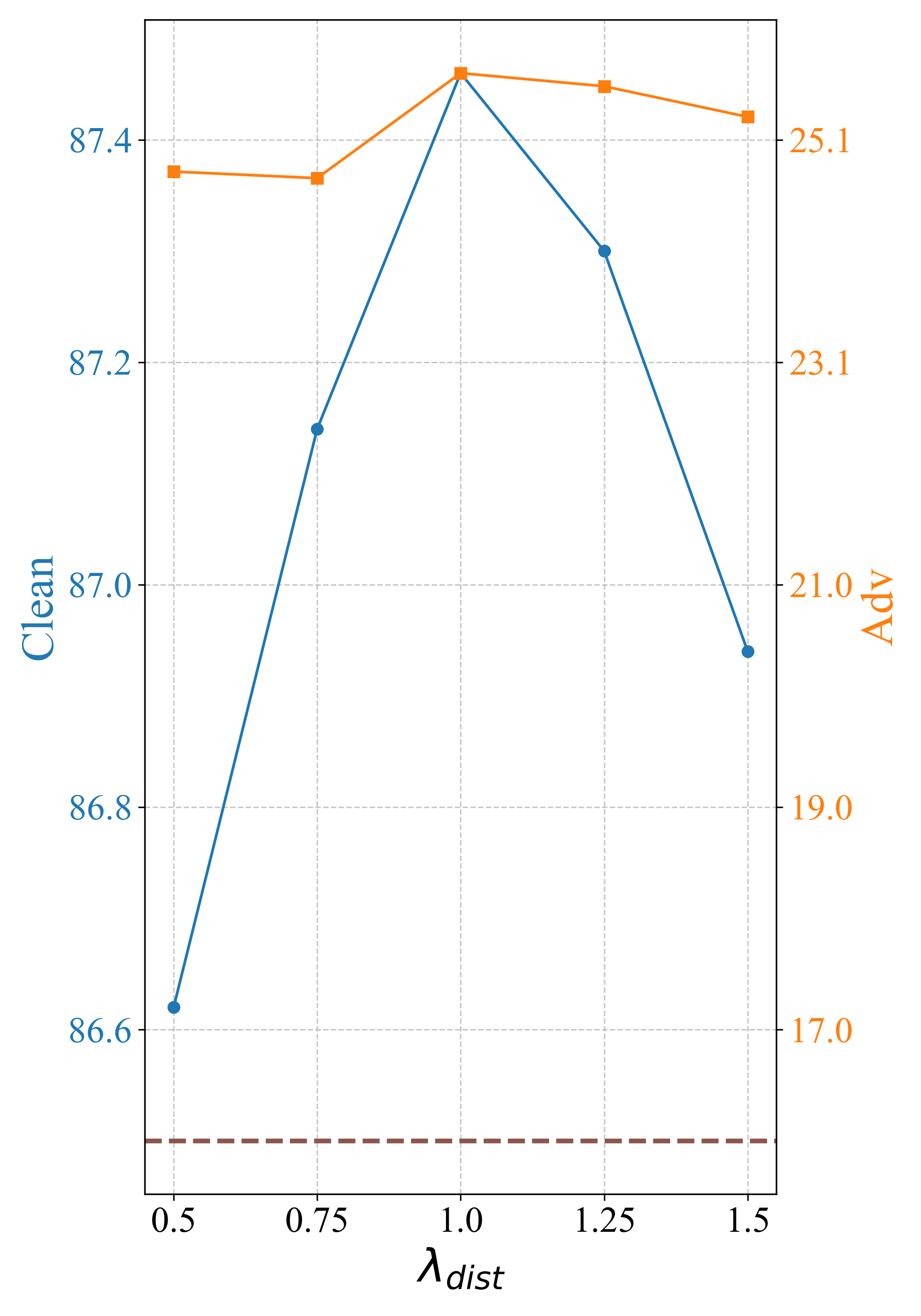}
    \caption{}%
    \label{fig:l-dist-abl}
  \end{subfigure}
  \begin{subfigure}[b]{0.32\textwidth}
    \centering
    \includegraphics[width=\textwidth]{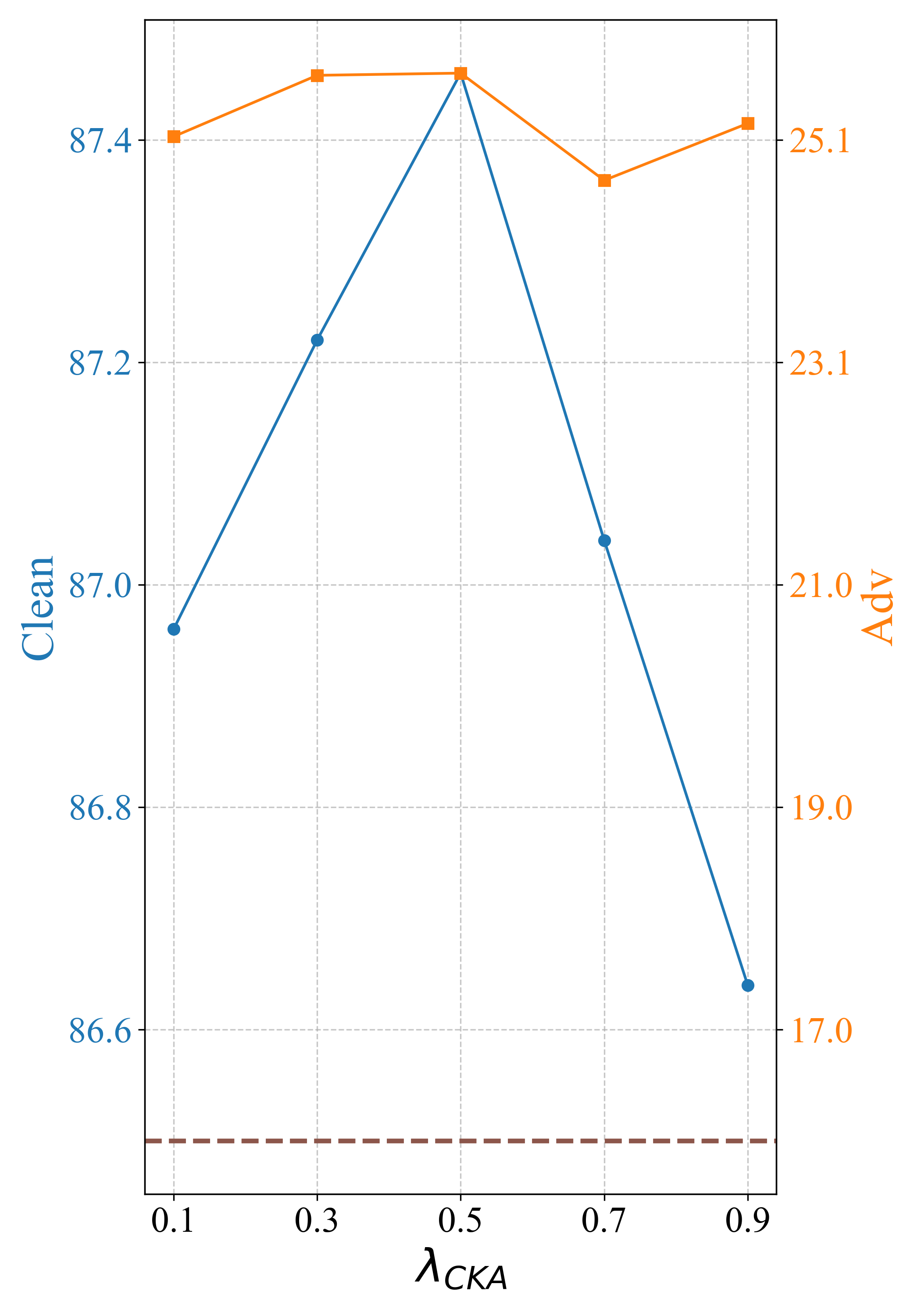}
    \caption{}%
    \label{fig:l-cka-abl}
  \end{subfigure}
  \caption{Ablation studies on $\lambda_{cs},\lambda_{dist},\lambda_{cka}$, introduced in Section~\ref{sec:canodistill-aug}. We select $\lambda_{cs}=0.4,\lambda_{dist}=1.0,\lambda_{cka}=0.5$ in all of our experiments on ImageNet. }
  \label{fig:supp-abl-loss}
\end{figure}

\textbf{\textit{\CanoDistill{}}} involves 3 losses, $\lambda_{cs},\lambda_{dist},\lambda_{cka}$, each having its own weights. Here, we perform thorough ablation studies on $\lambda_{cs},\lambda_{dist},\lambda_{cka}$ on ImageNet100. The results are given in Figure~\ref{fig:supp-abl-loss}. We perform the ablation study on one loss function by fixing the other weights to their own optimal values. We empirically conclude this setting: $\lambda_{cs}=0.4,\lambda_{dist}=1.0,\lambda_{cka}=0.5$, for all experiments on ImageNet. For CIFAR10, we perform grid search over several parameter combinations and fix $\lambda_{cs}=0.2,\lambda_{dist}=0.25,\lambda_{cka}=0.5$. 

\subsubsection{The design of $\mathcal{L}_{cano}$}
\label{sec:supp-lcano-ce-nopos-discussion}
We design $\mathcal{L}_{align}$ and $\mathcal{L}_{cano}$ to both have push-together and pull-away effects, inspired by \citet{supcon}. In Eq. \eqref{eq:cano-canodistill}, each \Canoimg{} treats other \Canoimg{}s of the same class (excluding itself) as positive examples. If a class happens to contribute only a single \Canoimg{}, no such positives exist. In that case, we optimize only the "pull-away" term—the denominator that separates the anchor from negatives in other classes, so the loss remains well-defined and informative. In this case, $\mathcal{L}_{cano}$ becomes:
\begin{equation}
    \mathcal{L}_{cano}=\frac{1}{b}
        \sum_{i=1}^b
        \log \sum_{k \ne i} \exp\!\bigl(\tilde{\bm{z}}_i \cdot \tilde{\bm{z}}_k  / \tau\bigr).
    \label{eq:cano-canodistill-supp-nopos}
\end{equation}
We perform a simple ablation study on using cross-entropy for discriminating between \Cano{}s from different classes, using ImageNet100. The results are given in Table~\ref{tab:l-cano-ce-loss}.
\begin{table}[]
    \centering
    \caption{The ablation study on using cross-entropy as $\mathcal{L}_{cano}$. Our design in Eq.~\ref{eq:cano-canodistill} achieves better results.}
    \label{tab:l-cano-ce-loss}
    \begin{tabular}{c|cc}
       $\mathcal{L}_{cano}$  & Clean & AutoAttack \cite{autoattack} \\
       \midrule
       Vanilla \cite{resnet}  & 86.5 & 15.9 \\
       \midrule
       Cross-entropy  & 86.8 & 25.3 \\
       \rowcolor{gray!20} Ours & 87.5 & 25.7
    \end{tabular}
\end{table}
Our design achieves better results in both clean accuracy and adversarial robustness, which is in line with the previous claim \cite{supcon}. In our case, \Cano{}s are far less than the original images and are easier to classify, which can cause overfitting issues when using cross-entropy and lead to suboptimal results. 

\subsubsection{The magnitude of classifier-free guidance}
\begin{figure}
    \centering
    \includegraphics[width=1.0\linewidth]{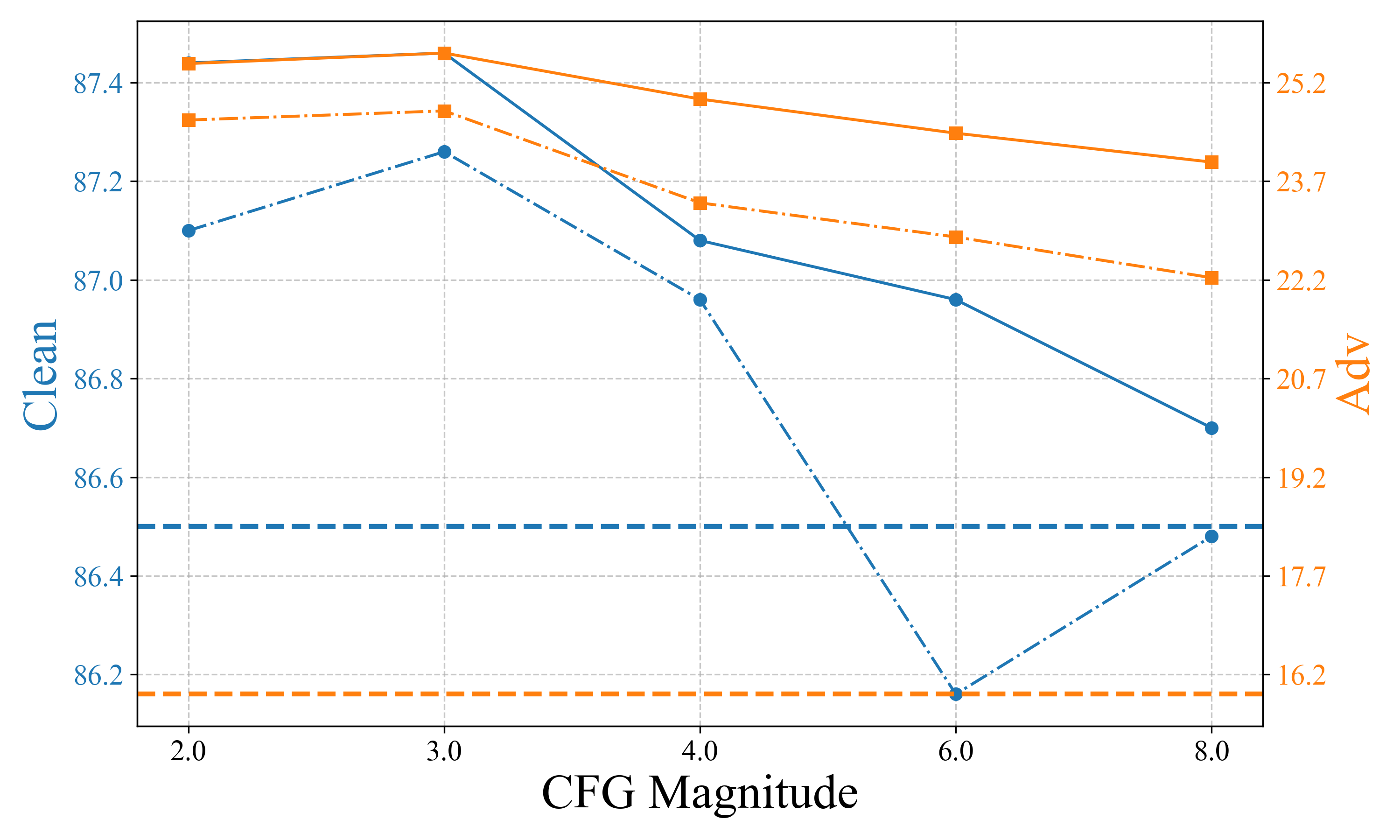}
    \caption{The ablation study on the magnitude of CFG used in our experiments on ImageNet. The student is trained with \textbf{\textit{\CanoDistill{}}} (solid lines) and \DMDistill{} (dash-dot lines). The dashed lines are the baselines. Larger CFG magnitudes do not necessarily contribute to a better performance, indicating that our design in \textbf{\textit{\CanoDistill{}}} is not simply a converging prior on the features (Section~\ref{sec:supp-limit-nmi-not-converging}). }
    \label{fig:cfg-abl}
\end{figure}
On ImageNet, we use CFG after projecting away the \ExtraDir{}s. We perform an ablation study on the CFG magnitude. The results are shown in Figure~\ref{fig:cfg-abl}. Notably, larger CFG magnitudes do not correspond to better performance. An overly large CFG scale can even worsen student performance. This is because our \textbf{\textit{\CanoDistill{}}} are not simply providing a converging prior over the student features, as discussed in Section~\ref{sec:supp-limit-nmi-not-converging}. We choose the magnitude to be 3 for both \textbf{\textit{\CanoDistill{}}} and \DMDistill{}. 

\subsection{Generalization of \textbf{\textit{\CanoDistill{}}} to different student architectures}
\label{sec:generalization-swin}
We demonstrate that \textbf{\textit{\CanoDistill{}}} is effective when the student is a transformer architecture. Specifically, we train a Swin-Tiny \cite{swin} model and a FAN-Tiny \cite{fan} model on ImageNet100. The result is given in Table~\ref{tab:swin-tiny}. We train the networks using the same setting as described in Section~\ref{sec:supp-training-detail}, except that we follow the timm data augmentation with Mixup \cite{mixup} and CutMix \cite{cutmix}, and we have a 5-epoch learning rate warm-up. 

\begin{table}[]
    \centering
    \caption{Comparison between a vanilla Swin-Tiny \cite{swin} model, a FAN-Small \cite{fan} model, and the ones trained with \textbf{\textit{\CanoDistill{}}}. Our method is effective with transformer students. It also proves that \textbf{\textit{\CanoDistill{}}} is effective with modern data augmentation techniques such as Mixup \cite{mixup} and CutMix \cite{cutmix}.}
    \label{tab:swin-tiny}
    \begin{tabular}{c|cc}
       Model  & Clean & AutoAttack \cite{autoattack} \\
       \midrule
       Swin-Tiny \cite{swin}  & 81.8 & 9.3 \\
       \rowcolor{gray!20} \textbf{\textit{\CanoDistill{}}} & \textbf{84.4} & \textbf{13.8} \\
       \midrule
       \midrule
       FAN-Tiny \cite{fan}  & 84.3 & 25.8 \\
       \rowcolor{gray!20} \textbf{\textit{\CanoDistill{}}} & \textbf{84.4} & \textbf{29.1}
    \end{tabular}
\end{table}

\section{Discussion on the quantitative results}
\label{sec:discussion-quant-result}
Our proposed method, \textbf{\textit{\CanoDistill{}}}, consistently improves the adversarial robustness and generalization ability of the student model. While the baseline methods can achieve better results on some benchmarks (\textit{e.g.} the IM-C \cite{imagenet-c} test on DiffAug \cite{diffaug}), they can worsen the student's performance on other benchmarks (\textcolor{venetianred}{Red} marks). This phenomenon reveals the established observation: the multifacetedness of robustness. Despite this, \textbf{\textit{\CanoDistill{}}} still consistently improves the student's performance. The baseline methods that do not use our proposed feature distillation pipeline all necessitate access to the full dataset, while our methods achieve competitive performance with access only to 10\% of the data. 

The performance trends differ in CIFAR10 and ImageNet (Table~\ref{tab:main-quant}). On CIFAR10, even if simply distilling the raw diffusion features to the student via \DMFit{} can contribute to the performance on all benchmarks, while on ImageNet, all the baseline methods can perform worse than the vanilla model in certain cases. We assume that two factors can lead to such a difference. The first is that CIFAR10 is a simpler dataset than ImageNet. Most images in CIFAR10 contain purely foreground objects, while the images in ImageNet are much noisier and harder to classify. The evidence is that CIFAR10 is a nearly solved dataset, as the classification accuracy approaches 100\% \cite{vit}, while the top models on ImageNet can achieve $\sim$90\% \cite{best-imagenet}. 

More importantly, we identify a critical difference in the CDM on CIFAR10 and on ImageNet. That is, ImageNet-trained CDMs are typically trained in a low-resolution latent space of a pre-trained variational autoencoder (VAE) \cite{vae,ldm}. This low-resolution space loses detailed information compared to the pixel space, reducing the discriminative power. To validate, we train a ResNet50 \cite{resnet} in this latent space for image classification, receiving the input as the VAE-encoded images. Notably, the clean accuracy drastically drops from 86.5 to 76.6. We assume that the missing discriminative information can negatively affect the performance of all feature distillation methods, because the diffusion features lie in the low-resolution latent space. Due to the limits on computational resources, we leave the investigation on pixel-space DM on ImageNet as a future direction.

\section{On more sophisticated feature distillation frameworks}
In Section~\ref{sec:supp-ldist-hint-rkd-cka}, we demonstrate that our feature distillation loss outperforms the ones used in existing diffusion-based feature distillation frameworks. In this experiment, we use a single-layer distillation framework. That is, the alignment between the student and the teacher only happens at one layer, respectively. We do not consider more sophisticated feature distillation frameworks such as multi-layer alignments \cite{dreamteacher,vit-vit-distill-bad} due to limited computational resources. We believe investigating the combinations between \textbf{\textit{\CanoDistill{}}} and different feature distillation frameworks is a promising future direction.


\section{More visual results}
\label{sec:supp-more-visual-results}
We provide more visualizations of \Cano{}s using \Canoimg{}s obtained from the DiT \cite{dit} used in our \textbf{\textit{\CanoDistill{}}} experiments on ImageNet \cite{deng2009imagenet}, in Figure~\ref{tab:supp-visual-1}, \ref{tab:supp-visual-2}, \ref{tab:supp-visual-3}, \ref{tab:supp-visual-4}, \ref{tab:supp-visual-5}. All the classes are from ImageNet. 

\begin{table}[]
    \centering
    \resizebox{\linewidth}{!}{
    \begin{tabular}{c|ccc|ccc|ccc}
    Class & Original & CFG & \MethodName{} & Original & CFG & \MethodName{} & Original & CFG & \MethodName{} \\
    \midrule
       \rotatebox{90}{\parbox{1.64cm}{\centering \textbf{\hspace{0pt} African crocodile}}} & \includegraphics[width=0.11\textwidth]{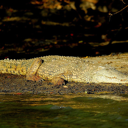}  & \includegraphics[width=0.11\textwidth]{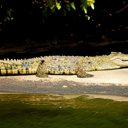}  & \includegraphics[width=0.11\textwidth]{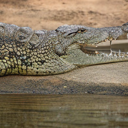}  & \includegraphics[width=0.11\textwidth]{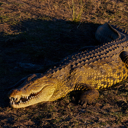}  & \includegraphics[width=0.11\textwidth]{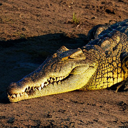}  & \includegraphics[width=0.11\textwidth]{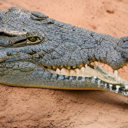}  & \includegraphics[width=0.11\textwidth]{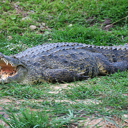}  & \includegraphics[width=0.11\textwidth]{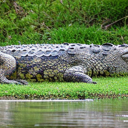}  & \includegraphics[width=0.11\textwidth]{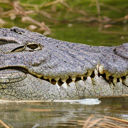} \\
\rotatebox{90}{\parbox{1.64cm}{\centering \textbf{\hspace{0pt} American chameleon}}} & \includegraphics[width=0.11\textwidth]{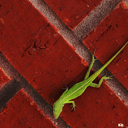}  & \includegraphics[width=0.11\textwidth]{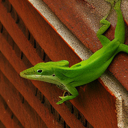}  & \includegraphics[width=0.11\textwidth]{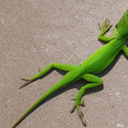}  & \includegraphics[width=0.11\textwidth]{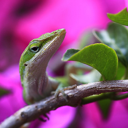}  & \includegraphics[width=0.11\textwidth]{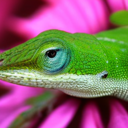}  & \includegraphics[width=0.11\textwidth]{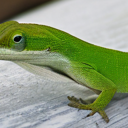}  & \includegraphics[width=0.11\textwidth]{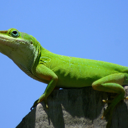}  & \includegraphics[width=0.11\textwidth]{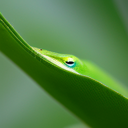}  & \includegraphics[width=0.11\textwidth]{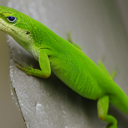} \\
\rotatebox{90}{\parbox{1.64cm}{\centering \textbf{\hspace{0pt} Indian cobra}}} & \includegraphics[width=0.11\textwidth]{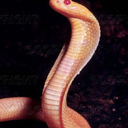}  & \includegraphics[width=0.11\textwidth]{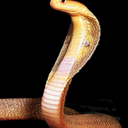}  & \includegraphics[width=0.11\textwidth]{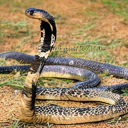}  & \includegraphics[width=0.11\textwidth]{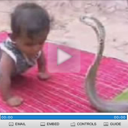}  & \includegraphics[width=0.11\textwidth]{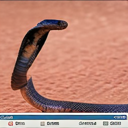}  & \includegraphics[width=0.11\textwidth]{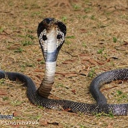}  & \includegraphics[width=0.11\textwidth]{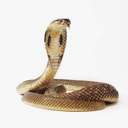}  & \includegraphics[width=0.11\textwidth]{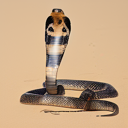}  & \includegraphics[width=0.11\textwidth]{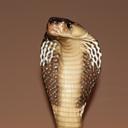} \\
\rotatebox{90}{\parbox{1.64cm}{\centering \textbf{\hspace{0pt} Acoustic guitar}}} & \includegraphics[width=0.11\textwidth]{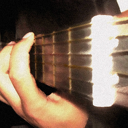}  & \includegraphics[width=0.11\textwidth]{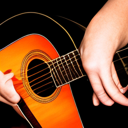}  & \includegraphics[width=0.11\textwidth]{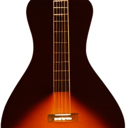}  & \includegraphics[width=0.11\textwidth]{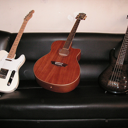}  & \includegraphics[width=0.11\textwidth]{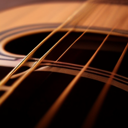}  & \includegraphics[width=0.11\textwidth]{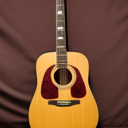}  & \includegraphics[width=0.11\textwidth]{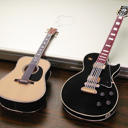}  & \includegraphics[width=0.11\textwidth]{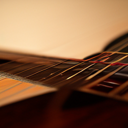}  & \includegraphics[width=0.11\textwidth]{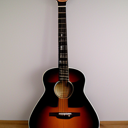} \\
\rotatebox{90}{\parbox{1.64cm}{\centering \textbf{\hspace{0pt} Airliner}}} & \includegraphics[width=0.11\textwidth]{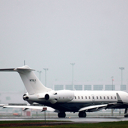}  & \includegraphics[width=0.11\textwidth]{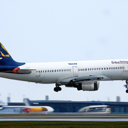}  & \includegraphics[width=0.11\textwidth]{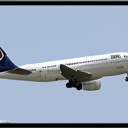}  & \includegraphics[width=0.11\textwidth]{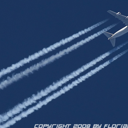}  & \includegraphics[width=0.11\textwidth]{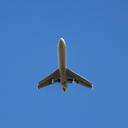}  & \includegraphics[width=0.11\textwidth]{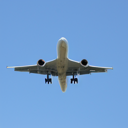}  & \includegraphics[width=0.11\textwidth]{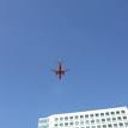}  & \includegraphics[width=0.11\textwidth]{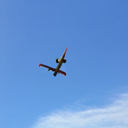}  & \includegraphics[width=0.11\textwidth]{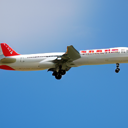} \\
\rotatebox{90}{\parbox{1.64cm}{\centering \textbf{\hspace{0pt} Airship}}} & \includegraphics[width=0.11\textwidth]{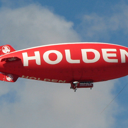}  & \includegraphics[width=0.11\textwidth]{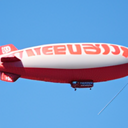}  & \includegraphics[width=0.11\textwidth]{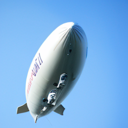}  & \includegraphics[width=0.11\textwidth]{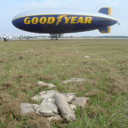}  & \includegraphics[width=0.11\textwidth]{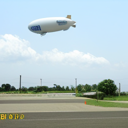}  & \includegraphics[width=0.11\textwidth]{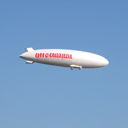}  & \includegraphics[width=0.11\textwidth]{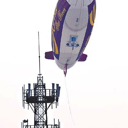}  & \includegraphics[width=0.11\textwidth]{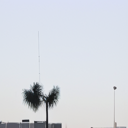}  & \includegraphics[width=0.11\textwidth]{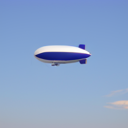} \\
\rotatebox{90}{\parbox{1.64cm}{\centering \textbf{\hspace{0pt} Amphibian}}} & \includegraphics[width=0.11\textwidth]{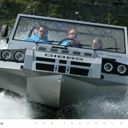}  & \includegraphics[width=0.11\textwidth]{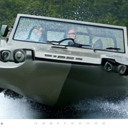}  & \includegraphics[width=0.11\textwidth]{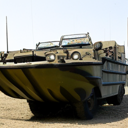}  & \includegraphics[width=0.11\textwidth]{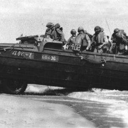}  & \includegraphics[width=0.11\textwidth]{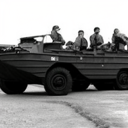}  & \includegraphics[width=0.11\textwidth]{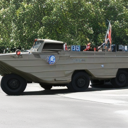}  & \includegraphics[width=0.11\textwidth]{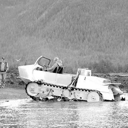}  & \includegraphics[width=0.11\textwidth]{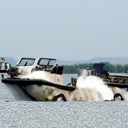}  & \includegraphics[width=0.11\textwidth]{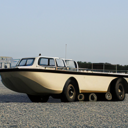} \\
\rotatebox{90}{\parbox{1.64cm}{\centering \textbf{\hspace{0pt} Analog clock}}} & \includegraphics[width=0.11\textwidth]{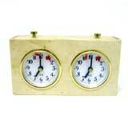}  & \includegraphics[width=0.11\textwidth]{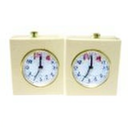}  & \includegraphics[width=0.11\textwidth]{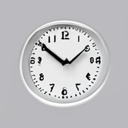}  & \includegraphics[width=0.11\textwidth]{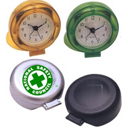}  & \includegraphics[width=0.11\textwidth]{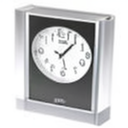}  & \includegraphics[width=0.11\textwidth]{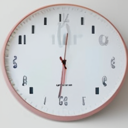}  & \includegraphics[width=0.11\textwidth]{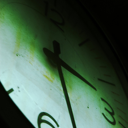}  & \includegraphics[width=0.11\textwidth]{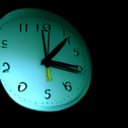}  & \includegraphics[width=0.11\textwidth]{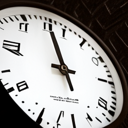} \\
\rotatebox{90}{\parbox{1.64cm}{\centering \textbf{\hspace{0pt} Assault rifle}}} & \includegraphics[width=0.11\textwidth]{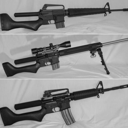}  & \includegraphics[width=0.11\textwidth]{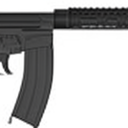}  & \includegraphics[width=0.11\textwidth]{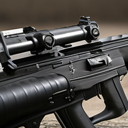}  & \includegraphics[width=0.11\textwidth]{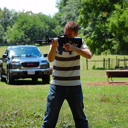}  & \includegraphics[width=0.11\textwidth]{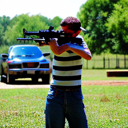}  & \includegraphics[width=0.11\textwidth]{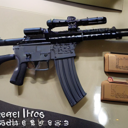}  & \includegraphics[width=0.11\textwidth]{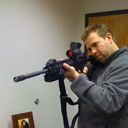}  & \includegraphics[width=0.11\textwidth]{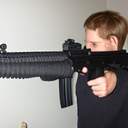}  & \includegraphics[width=0.11\textwidth]{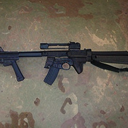} \\
\rotatebox{90}{\parbox{1.64cm}{\centering \textbf{\hspace{0pt} Bakery}}} & \includegraphics[width=0.11\textwidth]{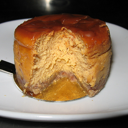}  & \includegraphics[width=0.11\textwidth]{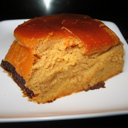}  & \includegraphics[width=0.11\textwidth]{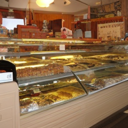}  & \includegraphics[width=0.11\textwidth]{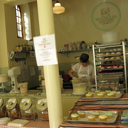}  & \includegraphics[width=0.11\textwidth]{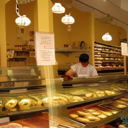}  & \includegraphics[width=0.11\textwidth]{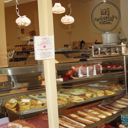}  & \includegraphics[width=0.11\textwidth]{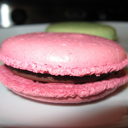}  & \includegraphics[width=0.11\textwidth]{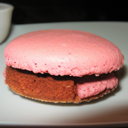}  & \includegraphics[width=0.11\textwidth]{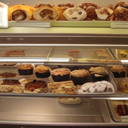} \\
\rotatebox{90}{\parbox{1.64cm}{\centering \textbf{\hspace{0pt} Balloon}}} & \includegraphics[width=0.11\textwidth]{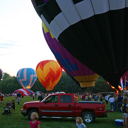}  & \includegraphics[width=0.11\textwidth]{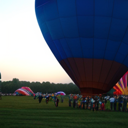}  & \includegraphics[width=0.11\textwidth]{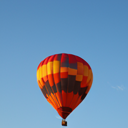}  & \includegraphics[width=0.11\textwidth]{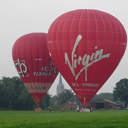}  & \includegraphics[width=0.11\textwidth]{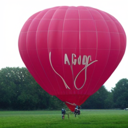}  & \includegraphics[width=0.11\textwidth]{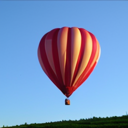}  & \includegraphics[width=0.11\textwidth]{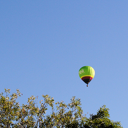}  & \includegraphics[width=0.11\textwidth]{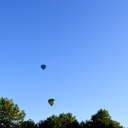}  & \includegraphics[width=0.11\textwidth]{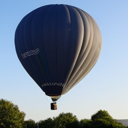} \\
\rotatebox{90}{\parbox{1.64cm}{\centering \textbf{\hspace{0pt} Ballpoint}}} & \includegraphics[width=0.11\textwidth]{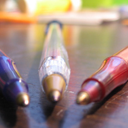}  & \includegraphics[width=0.11\textwidth]{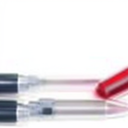}  & \includegraphics[width=0.11\textwidth]{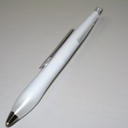}  & \includegraphics[width=0.11\textwidth]{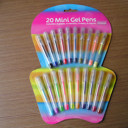}  & \includegraphics[width=0.11\textwidth]{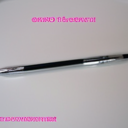}  & \includegraphics[width=0.11\textwidth]{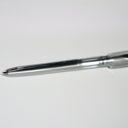}  & \includegraphics[width=0.11\textwidth]{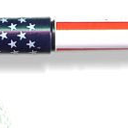}  & \includegraphics[width=0.11\textwidth]{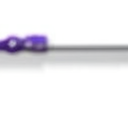}  & \includegraphics[width=0.11\textwidth]{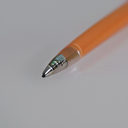} \\
\rotatebox{90}{\parbox{1.64cm}{\centering \textbf{\hspace{0pt} Banded gecko}}} & \includegraphics[width=0.11\textwidth]{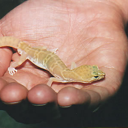}  & \includegraphics[width=0.11\textwidth]{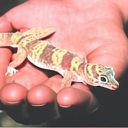}  & \includegraphics[width=0.11\textwidth]{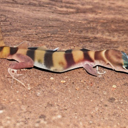}  & \includegraphics[width=0.11\textwidth]{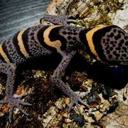}  & \includegraphics[width=0.11\textwidth]{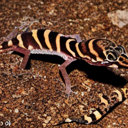}  & \includegraphics[width=0.11\textwidth]{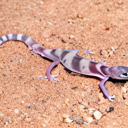}  & \includegraphics[width=0.11\textwidth]{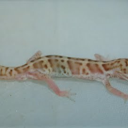}  & \includegraphics[width=0.11\textwidth]{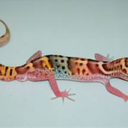}  & \includegraphics[width=0.11\textwidth]{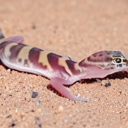} \\
\rotatebox{90}{\parbox{1.64cm}{\centering \textbf{\hspace{0pt} Bannister}}} & \includegraphics[width=0.11\textwidth]{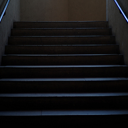}  & \includegraphics[width=0.11\textwidth]{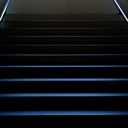}  & \includegraphics[width=0.11\textwidth]{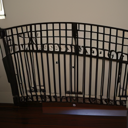}  & \includegraphics[width=0.11\textwidth]{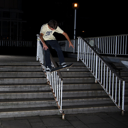}  & \includegraphics[width=0.11\textwidth]{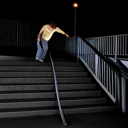}  & \includegraphics[width=0.11\textwidth]{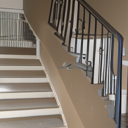}  & \includegraphics[width=0.11\textwidth]{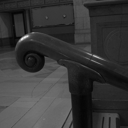}  & \includegraphics[width=0.11\textwidth]{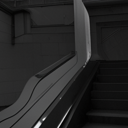}  & \includegraphics[width=0.11\textwidth]{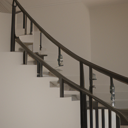} \\
\rotatebox{90}{\parbox{1.64cm}{\centering \textbf{\hspace{0pt} Barn spider}}} & \includegraphics[width=0.11\textwidth]{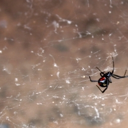}  & \includegraphics[width=0.11\textwidth]{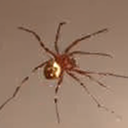}  & \includegraphics[width=0.11\textwidth]{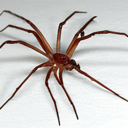}  & \includegraphics[width=0.11\textwidth]{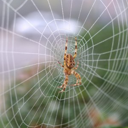}  & \includegraphics[width=0.11\textwidth]{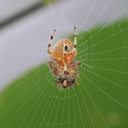}  & \includegraphics[width=0.11\textwidth]{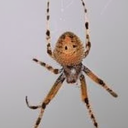}  & \includegraphics[width=0.11\textwidth]{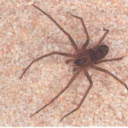}  & \includegraphics[width=0.11\textwidth]{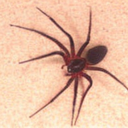}  & \includegraphics[width=0.11\textwidth]{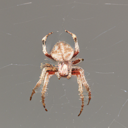} \\
\rotatebox{90}{\parbox{1.64cm}{\centering \textbf{\hspace{0pt} Bee eater}}} & \includegraphics[width=0.11\textwidth]{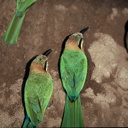}  & \includegraphics[width=0.11\textwidth]{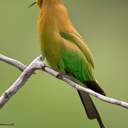}  & \includegraphics[width=0.11\textwidth]{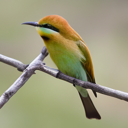}  & \includegraphics[width=0.11\textwidth]{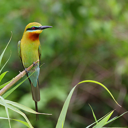}  & \includegraphics[width=0.11\textwidth]{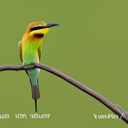}  & \includegraphics[width=0.11\textwidth]{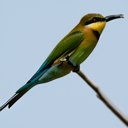}  & \includegraphics[width=0.11\textwidth]{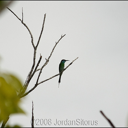}  & \includegraphics[width=0.11\textwidth]{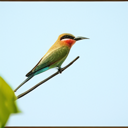}  & \includegraphics[width=0.11\textwidth]{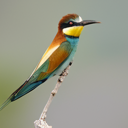} \\
    \end{tabular}
    }
    \captionof{figure}{Visualizations of \Cano{}s using \Canoimg{}s obtained from the DiT \cite{dit} used in our \textbf{\textit{\CanoDistill{}}} experiments on ImageNet.}
    \label{tab:supp-visual-1}
\end{table}

\begin{table}[]
    \centering
    \resizebox{\linewidth}{!}{
    \begin{tabular}{c|ccc|ccc|ccc}
    Class & Original & CFG & \MethodName{} & Original & CFG & \MethodName{} & Original & CFG & \MethodName{} \\
    \midrule
       \rotatebox{90}{\parbox{1.64cm}{\centering \textbf{\hspace{0pt} Black grouse}}} & \includegraphics[width=0.11\textwidth]{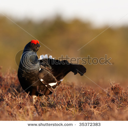}  & \includegraphics[width=0.11\textwidth]{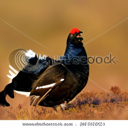}  & \includegraphics[width=0.11\textwidth]{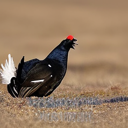}  & \includegraphics[width=0.11\textwidth]{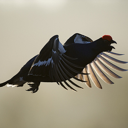}  & \includegraphics[width=0.11\textwidth]{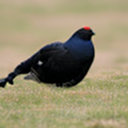}  & \includegraphics[width=0.11\textwidth]{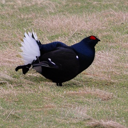}  & \includegraphics[width=0.11\textwidth]{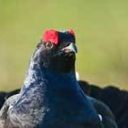}  & \includegraphics[width=0.11\textwidth]{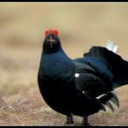}  & \includegraphics[width=0.11\textwidth]{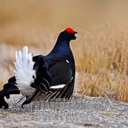} \\
\rotatebox{90}{\parbox{1.64cm}{\centering \textbf{\hspace{0pt} Bottlecap}}} & \includegraphics[width=0.11\textwidth]{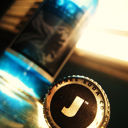}  & \includegraphics[width=0.11\textwidth]{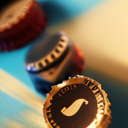}  & \includegraphics[width=0.11\textwidth]{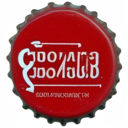}  & \includegraphics[width=0.11\textwidth]{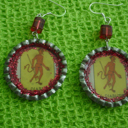}  & \includegraphics[width=0.11\textwidth]{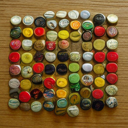}  & \includegraphics[width=0.11\textwidth]{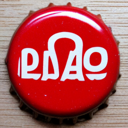}  & \includegraphics[width=0.11\textwidth]{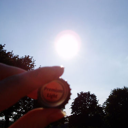}  & \includegraphics[width=0.11\textwidth]{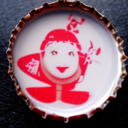}  & \includegraphics[width=0.11\textwidth]{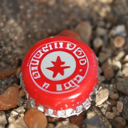} \\
\rotatebox{90}{\parbox{1.64cm}{\centering \textbf{\hspace{0pt} Drake}}} & \includegraphics[width=0.11\textwidth]{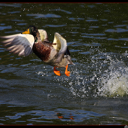}  & \includegraphics[width=0.11\textwidth]{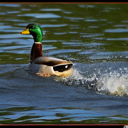}  & \includegraphics[width=0.11\textwidth]{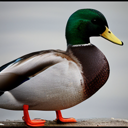}  & \includegraphics[width=0.11\textwidth]{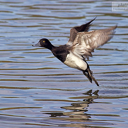}  & \includegraphics[width=0.11\textwidth]{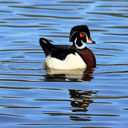}  & \includegraphics[width=0.11\textwidth]{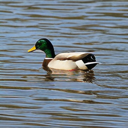}  & \includegraphics[width=0.11\textwidth]{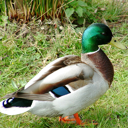}  & \includegraphics[width=0.11\textwidth]{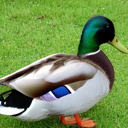}  & \includegraphics[width=0.11\textwidth]{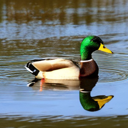} \\
\rotatebox{90}{\parbox{1.64cm}{\centering \textbf{\hspace{0pt} Frilled lizard}}} & \includegraphics[width=0.11\textwidth]{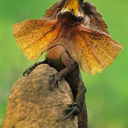}  & \includegraphics[width=0.11\textwidth]{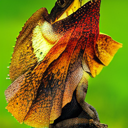}  & \includegraphics[width=0.11\textwidth]{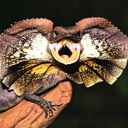}  & \includegraphics[width=0.11\textwidth]{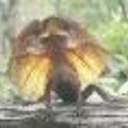}  & \includegraphics[width=0.11\textwidth]{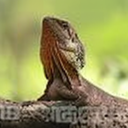}  & \includegraphics[width=0.11\textwidth]{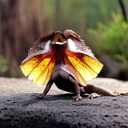}  & \includegraphics[width=0.11\textwidth]{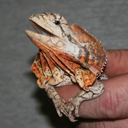}  & \includegraphics[width=0.11\textwidth]{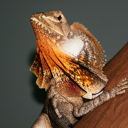}  & \includegraphics[width=0.11\textwidth]{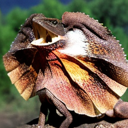} \\
\rotatebox{90}{\parbox{1.64cm}{\centering \textbf{\hspace{0pt} Hammerhead}}} & \includegraphics[width=0.11\textwidth]{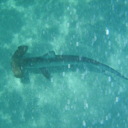}  & \includegraphics[width=0.11\textwidth]{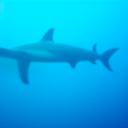}  & \includegraphics[width=0.11\textwidth]{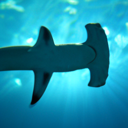}  & \includegraphics[width=0.11\textwidth]{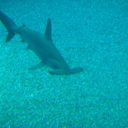}  & \includegraphics[width=0.11\textwidth]{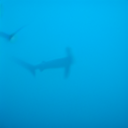}  & \includegraphics[width=0.11\textwidth]{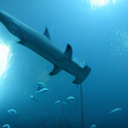}  & \includegraphics[width=0.11\textwidth]{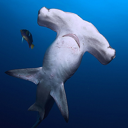}  & \includegraphics[width=0.11\textwidth]{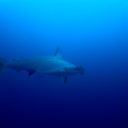}  & \includegraphics[width=0.11\textwidth]{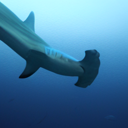} \\
\rotatebox{90}{\parbox{1.64cm}{\centering \textbf{\hspace{0pt} Hornbill}}} & \includegraphics[width=0.11\textwidth]{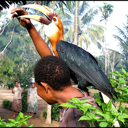}  & \includegraphics[width=0.11\textwidth]{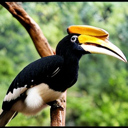}  & \includegraphics[width=0.11\textwidth]{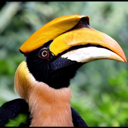}  & \includegraphics[width=0.11\textwidth]{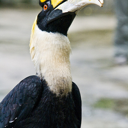}  & \includegraphics[width=0.11\textwidth]{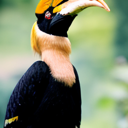}  & \includegraphics[width=0.11\textwidth]{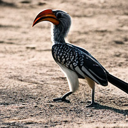}  & \includegraphics[width=0.11\textwidth]{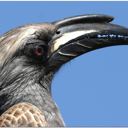}  & \includegraphics[width=0.11\textwidth]{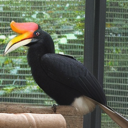}  & \includegraphics[width=0.11\textwidth]{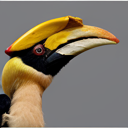} \\
\rotatebox{90}{\parbox{1.64cm}{\centering \textbf{\hspace{0pt} Kite}}} & \includegraphics[width=0.11\textwidth]{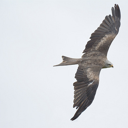}  & \includegraphics[width=0.11\textwidth]{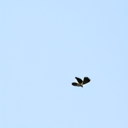}  & \includegraphics[width=0.11\textwidth]{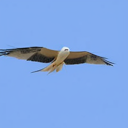}  & \includegraphics[width=0.11\textwidth]{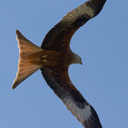}  & \includegraphics[width=0.11\textwidth]{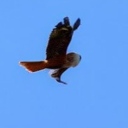}  & \includegraphics[width=0.11\textwidth]{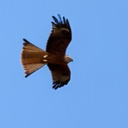}  & \includegraphics[width=0.11\textwidth]{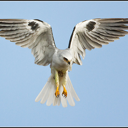}  & \includegraphics[width=0.11\textwidth]{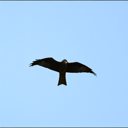}  & \includegraphics[width=0.11\textwidth]{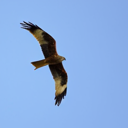} \\
\rotatebox{90}{\parbox{1.64cm}{\centering \textbf{\hspace{0pt} Mud turtle}}} & \includegraphics[width=0.11\textwidth]{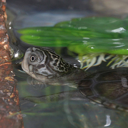}  & \includegraphics[width=0.11\textwidth]{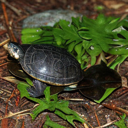}  & \includegraphics[width=0.11\textwidth]{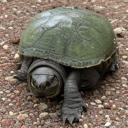}  & \includegraphics[width=0.11\textwidth]{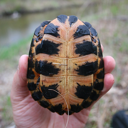}  & \includegraphics[width=0.11\textwidth]{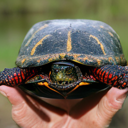}  & \includegraphics[width=0.11\textwidth]{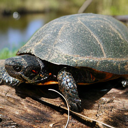}  & \includegraphics[width=0.11\textwidth]{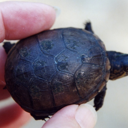}  & \includegraphics[width=0.11\textwidth]{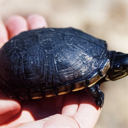}  & \includegraphics[width=0.11\textwidth]{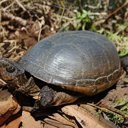} \\
\rotatebox{90}{\parbox{1.64cm}{\centering \textbf{\hspace{0pt} Peacock}}} & \includegraphics[width=0.11\textwidth]{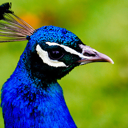}  & \includegraphics[width=0.11\textwidth]{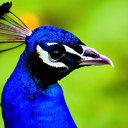}  & \includegraphics[width=0.11\textwidth]{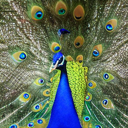}  & \includegraphics[width=0.11\textwidth]{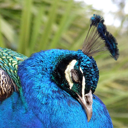}  & \includegraphics[width=0.11\textwidth]{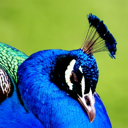}  & \includegraphics[width=0.11\textwidth]{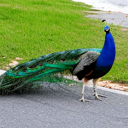}  & \includegraphics[width=0.11\textwidth]{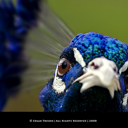}  & \includegraphics[width=0.11\textwidth]{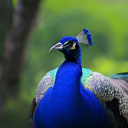}  & \includegraphics[width=0.11\textwidth]{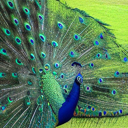} \\
\rotatebox{90}{\parbox{1.64cm}{\centering \textbf{\hspace{0pt} Ruffed grouse}}} & \includegraphics[width=0.11\textwidth]{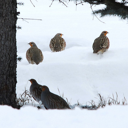}  & \includegraphics[width=0.11\textwidth]{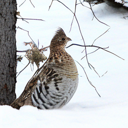}  & \includegraphics[width=0.11\textwidth]{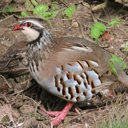}  & \includegraphics[width=0.11\textwidth]{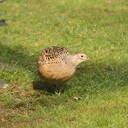}  & \includegraphics[width=0.11\textwidth]{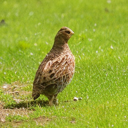}  & \includegraphics[width=0.11\textwidth]{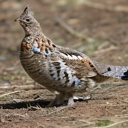}  & \includegraphics[width=0.11\textwidth]{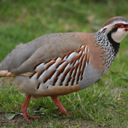}  & \includegraphics[width=0.11\textwidth]{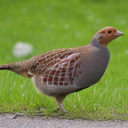}  & \includegraphics[width=0.11\textwidth]{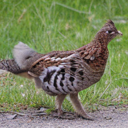} \\
\rotatebox{90}{\parbox{1.64cm}{\centering \textbf{\hspace{0pt} Scorpion}}} & \includegraphics[width=0.11\textwidth]{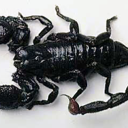}  & \includegraphics[width=0.11\textwidth]{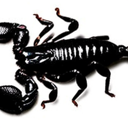}  & \includegraphics[width=0.11\textwidth]{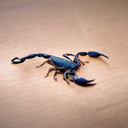}  & \includegraphics[width=0.11\textwidth]{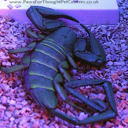}  & \includegraphics[width=0.11\textwidth]{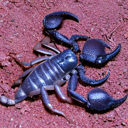}  & \includegraphics[width=0.11\textwidth]{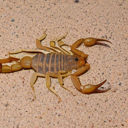}  & \includegraphics[width=0.11\textwidth]{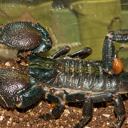}  & \includegraphics[width=0.11\textwidth]{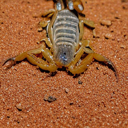}  & \includegraphics[width=0.11\textwidth]{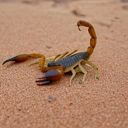} \\
\rotatebox{90}{\parbox{1.64cm}{\centering \textbf{\hspace{0pt} Tench}}} & \includegraphics[width=0.11\textwidth]{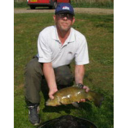}  & \includegraphics[width=0.11\textwidth]{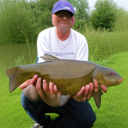}  & \includegraphics[width=0.11\textwidth]{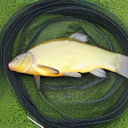}  & \includegraphics[width=0.11\textwidth]{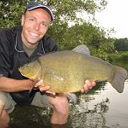}  & \includegraphics[width=0.11\textwidth]{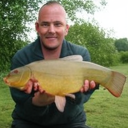}  & \includegraphics[width=0.11\textwidth]{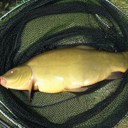}  & \includegraphics[width=0.11\textwidth]{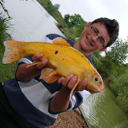}  & \includegraphics[width=0.11\textwidth]{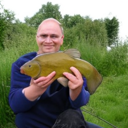}  & \includegraphics[width=0.11\textwidth]{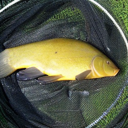} \\
\rotatebox{90}{\parbox{1.64cm}{\centering \textbf{\hspace{0pt} Thunder snake}}} & \includegraphics[width=0.11\textwidth]{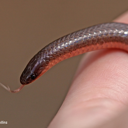}  & \includegraphics[width=0.11\textwidth]{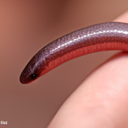}  & \includegraphics[width=0.11\textwidth]{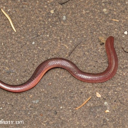}  & \includegraphics[width=0.11\textwidth]{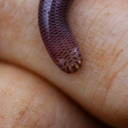}  & \includegraphics[width=0.11\textwidth]{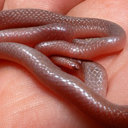}  & \includegraphics[width=0.11\textwidth]{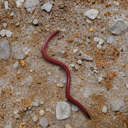}  & \includegraphics[width=0.11\textwidth]{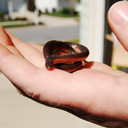}  & \includegraphics[width=0.11\textwidth]{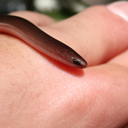}  & \includegraphics[width=0.11\textwidth]{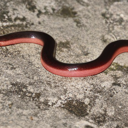} \\
\rotatebox{90}{\parbox{1.64cm}{\centering \textbf{\hspace{0pt} Tick}}} & \includegraphics[width=0.11\textwidth]{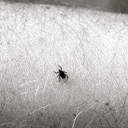}  & \includegraphics[width=0.11\textwidth]{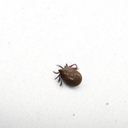}  & \includegraphics[width=0.11\textwidth]{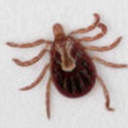}  & \includegraphics[width=0.11\textwidth]{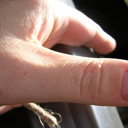}  & \includegraphics[width=0.11\textwidth]{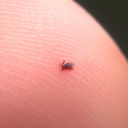}  & \includegraphics[width=0.11\textwidth]{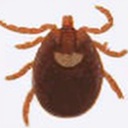}  & \includegraphics[width=0.11\textwidth]{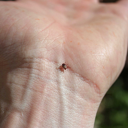}  & \includegraphics[width=0.11\textwidth]{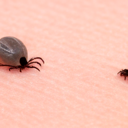}  & \includegraphics[width=0.11\textwidth]{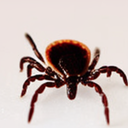} \\
\rotatebox{90}{\parbox{1.64cm}{\centering \textbf{\hspace{0pt} Tree frog}}} & \includegraphics[width=0.11\textwidth]{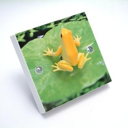}  & \includegraphics[width=0.11\textwidth]{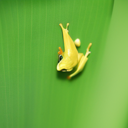}  & \includegraphics[width=0.11\textwidth]{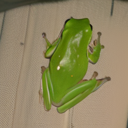}  & \includegraphics[width=0.11\textwidth]{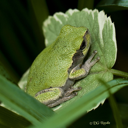}  & \includegraphics[width=0.11\textwidth]{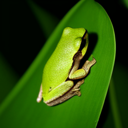}  & \includegraphics[width=0.11\textwidth]{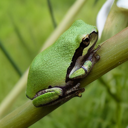}  & \includegraphics[width=0.11\textwidth]{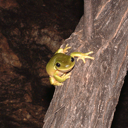}  & \includegraphics[width=0.11\textwidth]{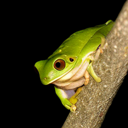}  & \includegraphics[width=0.11\textwidth]{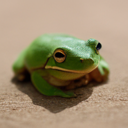} \\
\rotatebox{90}{\parbox{1.64cm}{\centering \textbf{\hspace{0pt} Tusker}}} & \includegraphics[width=0.11\textwidth]{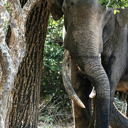}  & \includegraphics[width=0.11\textwidth]{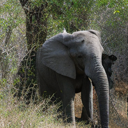}  & \includegraphics[width=0.11\textwidth]{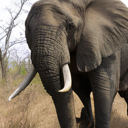}  & \includegraphics[width=0.11\textwidth]{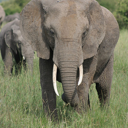}  & \includegraphics[width=0.11\textwidth]{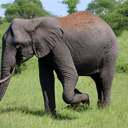}  & \includegraphics[width=0.11\textwidth]{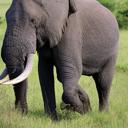}  & \includegraphics[width=0.11\textwidth]{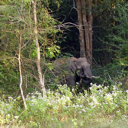}  & \includegraphics[width=0.11\textwidth]{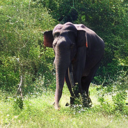}  & \includegraphics[width=0.11\textwidth]{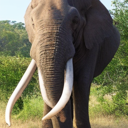} \\
    \end{tabular}
    }
    \captionof{figure}{Visualizations of \Cano{}s using \Canoimg{}s obtained from the DiT \cite{dit} used in our \textbf{\textit{\CanoDistill{}}} experiments on ImageNet.}
    \label{tab:supp-visual-2}
\end{table}

\begin{table}[]
    \centering
    \resizebox{\linewidth}{!}{
    \begin{tabular}{c|ccc|ccc|ccc}
    Class & Original & CFG & \MethodName{} & Original & CFG & \MethodName{} & Original & CFG & \MethodName{} \\
    \midrule
    \rotatebox{90}{\parbox{1.64cm}{\centering \textbf{\hspace{0pt} American lobster}}} & \includegraphics[width=0.11\textwidth]{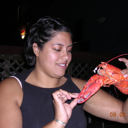}  & \includegraphics[width=0.11\textwidth]{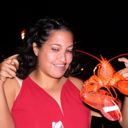}  & \includegraphics[width=0.11\textwidth]{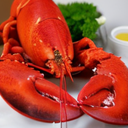}  & \includegraphics[width=0.11\textwidth]{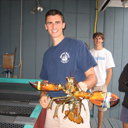}  & \includegraphics[width=0.11\textwidth]{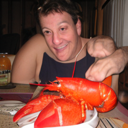}  & \includegraphics[width=0.11\textwidth]{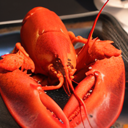}  & \includegraphics[width=0.11\textwidth]{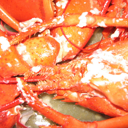}  & \includegraphics[width=0.11\textwidth]{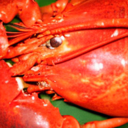}  & \includegraphics[width=0.11\textwidth]{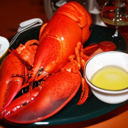} \\
\rotatebox{90}{\parbox{1.64cm}{\centering \textbf{\hspace{0pt} Chihuahua}}} & \includegraphics[width=0.11\textwidth]{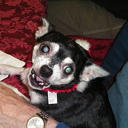}  & \includegraphics[width=0.11\textwidth]{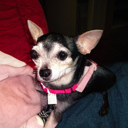}  & \includegraphics[width=0.11\textwidth]{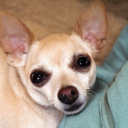}  & \includegraphics[width=0.11\textwidth]{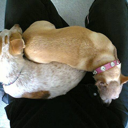}  & \includegraphics[width=0.11\textwidth]{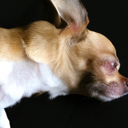}  & \includegraphics[width=0.11\textwidth]{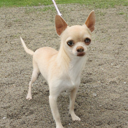}  & \includegraphics[width=0.11\textwidth]{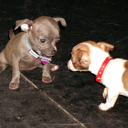}  & \includegraphics[width=0.11\textwidth]{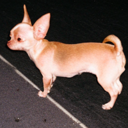}  & \includegraphics[width=0.11\textwidth]{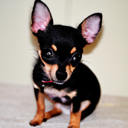} \\
\rotatebox{90}{\parbox{1.64cm}{\centering \textbf{\hspace{0pt} English foxhound}}} & \includegraphics[width=0.11\textwidth]{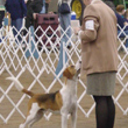}  & \includegraphics[width=0.11\textwidth]{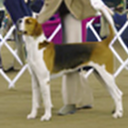}  & \includegraphics[width=0.11\textwidth]{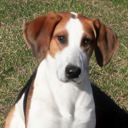}  & \includegraphics[width=0.11\textwidth]{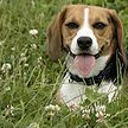}  & \includegraphics[width=0.11\textwidth]{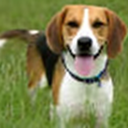}  & \includegraphics[width=0.11\textwidth]{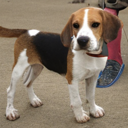}  & \includegraphics[width=0.11\textwidth]{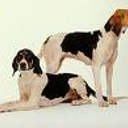}  & \includegraphics[width=0.11\textwidth]{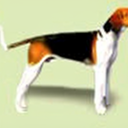}  & \includegraphics[width=0.11\textwidth]{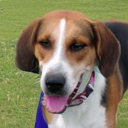} \\
\rotatebox{90}{\parbox{1.64cm}{\centering \textbf{\hspace{0pt} Maltese dog}}} & \includegraphics[width=0.11\textwidth]{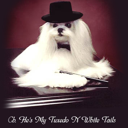}  & \includegraphics[width=0.11\textwidth]{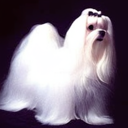}  & \includegraphics[width=0.11\textwidth]{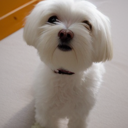}  & \includegraphics[width=0.11\textwidth]{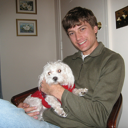}  & \includegraphics[width=0.11\textwidth]{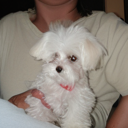}  & \includegraphics[width=0.11\textwidth]{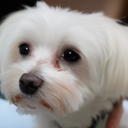}  & \includegraphics[width=0.11\textwidth]{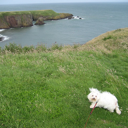}  & \includegraphics[width=0.11\textwidth]{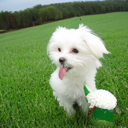}  & \includegraphics[width=0.11\textwidth]{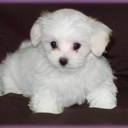} \\
\rotatebox{90}{\parbox{1.64cm}{\centering \textbf{\hspace{0pt} Cleaver}}} & \includegraphics[width=0.11\textwidth]{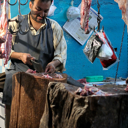}  & \includegraphics[width=0.11\textwidth]{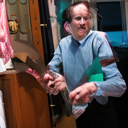}  & \includegraphics[width=0.11\textwidth]{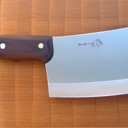}  & \includegraphics[width=0.11\textwidth]{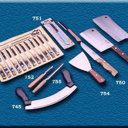}  & \includegraphics[width=0.11\textwidth]{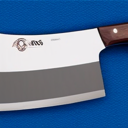}  & \includegraphics[width=0.11\textwidth]{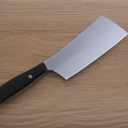}  & \includegraphics[width=0.11\textwidth]{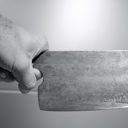}  & \includegraphics[width=0.11\textwidth]{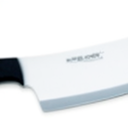}  & \includegraphics[width=0.11\textwidth]{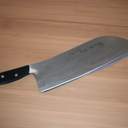} \\
\rotatebox{90}{\parbox{1.64cm}{\centering \textbf{\hspace{0pt} Conch}}} & \includegraphics[width=0.11\textwidth]{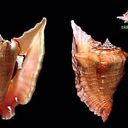}  & \includegraphics[width=0.11\textwidth]{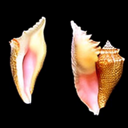}  & \includegraphics[width=0.11\textwidth]{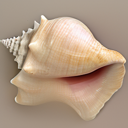}  & \includegraphics[width=0.11\textwidth]{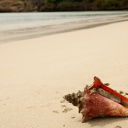}  & \includegraphics[width=0.11\textwidth]{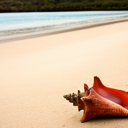}  & \includegraphics[width=0.11\textwidth]{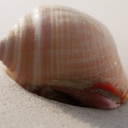}  & \includegraphics[width=0.11\textwidth]{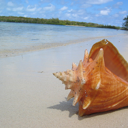}  & \includegraphics[width=0.11\textwidth]{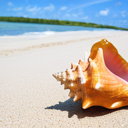}  & \includegraphics[width=0.11\textwidth]{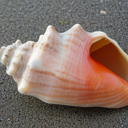} \\
\rotatebox{90}{\parbox{1.64cm}{\centering \textbf{\hspace{0pt} Dugong}}} & \includegraphics[width=0.11\textwidth]{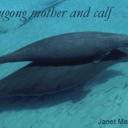}  & \includegraphics[width=0.11\textwidth]{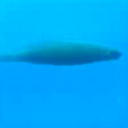}  & \includegraphics[width=0.11\textwidth]{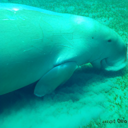}  & \includegraphics[width=0.11\textwidth]{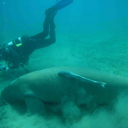}  & \includegraphics[width=0.11\textwidth]{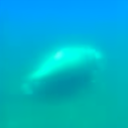}  & \includegraphics[width=0.11\textwidth]{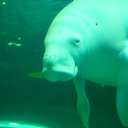}  & \includegraphics[width=0.11\textwidth]{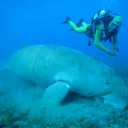}  & \includegraphics[width=0.11\textwidth]{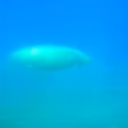}  & \includegraphics[width=0.11\textwidth]{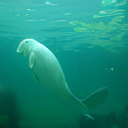} \\
\rotatebox{90}{\parbox{1.64cm}{\centering \textbf{\hspace{0pt} Isopod}}} & \includegraphics[width=0.11\textwidth]{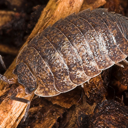}  & \includegraphics[width=0.11\textwidth]{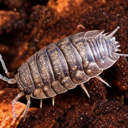}  & \includegraphics[width=0.11\textwidth]{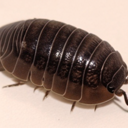}  & \includegraphics[width=0.11\textwidth]{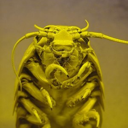}  & \includegraphics[width=0.11\textwidth]{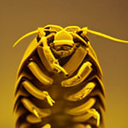}  & \includegraphics[width=0.11\textwidth]{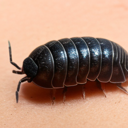}  & \includegraphics[width=0.11\textwidth]{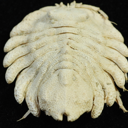}  & \includegraphics[width=0.11\textwidth]{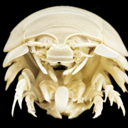}  & \includegraphics[width=0.11\textwidth]{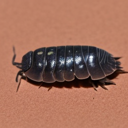} \\
\rotatebox{90}{\parbox{1.64cm}{\centering \textbf{\hspace{0pt} Jellyfish}}} & \includegraphics[width=0.11\textwidth]{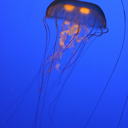}  & \includegraphics[width=0.11\textwidth]{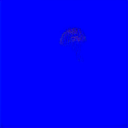}  & \includegraphics[width=0.11\textwidth]{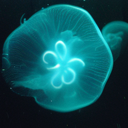}  & \includegraphics[width=0.11\textwidth]{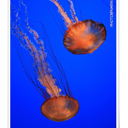}  & \includegraphics[width=0.11\textwidth]{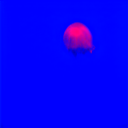}  & \includegraphics[width=0.11\textwidth]{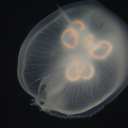}  & \includegraphics[width=0.11\textwidth]{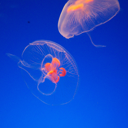}  & \includegraphics[width=0.11\textwidth]{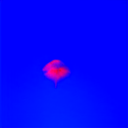}  & \includegraphics[width=0.11\textwidth]{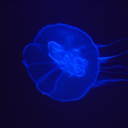} \\
\rotatebox{90}{\parbox{1.64cm}{\centering \textbf{\hspace{0pt} King crab}}} & \includegraphics[width=0.11\textwidth]{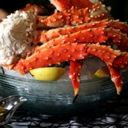}  & \includegraphics[width=0.11\textwidth]{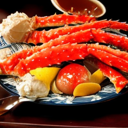}  & \includegraphics[width=0.11\textwidth]{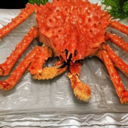}  & \includegraphics[width=0.11\textwidth]{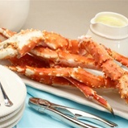}  & \includegraphics[width=0.11\textwidth]{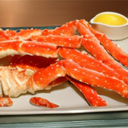}  & \includegraphics[width=0.11\textwidth]{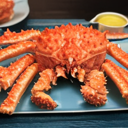}  & \includegraphics[width=0.11\textwidth]{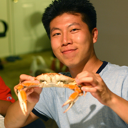}  & \includegraphics[width=0.11\textwidth]{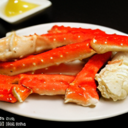}  & \includegraphics[width=0.11\textwidth]{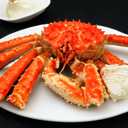} \\
\rotatebox{90}{\parbox{1.64cm}{\centering \textbf{\hspace{0pt} Limpkin}}} & \includegraphics[width=0.11\textwidth]{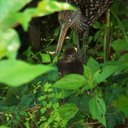}  & \includegraphics[width=0.11\textwidth]{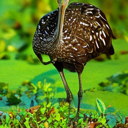}  & \includegraphics[width=0.11\textwidth]{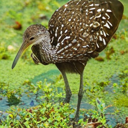}  & \includegraphics[width=0.11\textwidth]{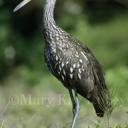}  & \includegraphics[width=0.11\textwidth]{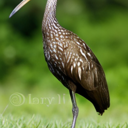}  & \includegraphics[width=0.11\textwidth]{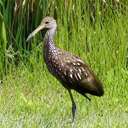}  & \includegraphics[width=0.11\textwidth]{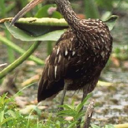}  & \includegraphics[width=0.11\textwidth]{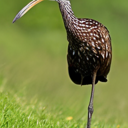}  & \includegraphics[width=0.11\textwidth]{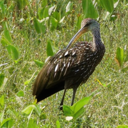} \\
\rotatebox{90}{\parbox{1.64cm}{\centering \textbf{\hspace{0pt} Otterhound}}} & \includegraphics[width=0.11\textwidth]{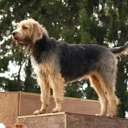}  & \includegraphics[width=0.11\textwidth]{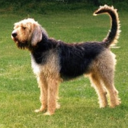}  & \includegraphics[width=0.11\textwidth]{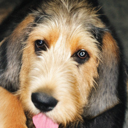}  & \includegraphics[width=0.11\textwidth]{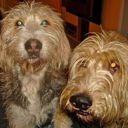}  & \includegraphics[width=0.11\textwidth]{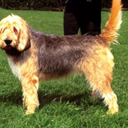}  & \includegraphics[width=0.11\textwidth]{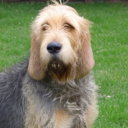}  & \includegraphics[width=0.11\textwidth]{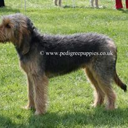}  & \includegraphics[width=0.11\textwidth]{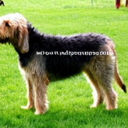}  & \includegraphics[width=0.11\textwidth]{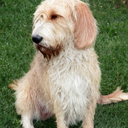} \\
\rotatebox{90}{\parbox{1.64cm}{\centering \textbf{\hspace{0pt} Oystercatcher}}} & \includegraphics[width=0.11\textwidth]{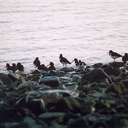}  & \includegraphics[width=0.11\textwidth]{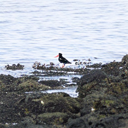}  & \includegraphics[width=0.11\textwidth]{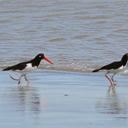}  & \includegraphics[width=0.11\textwidth]{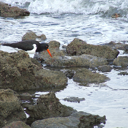}  & \includegraphics[width=0.11\textwidth]{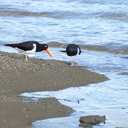}  & \includegraphics[width=0.11\textwidth]{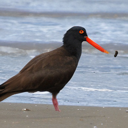}  & \includegraphics[width=0.11\textwidth]{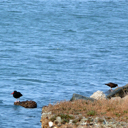}  & \includegraphics[width=0.11\textwidth]{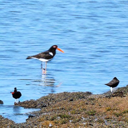}  & \includegraphics[width=0.11\textwidth]{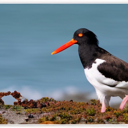} \\
\rotatebox{90}{\parbox{1.64cm}{\centering \textbf{\hspace{0pt} Red backed sandpiper}}} & \includegraphics[width=0.11\textwidth]{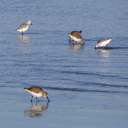}  & \includegraphics[width=0.11\textwidth]{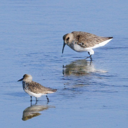}  & \includegraphics[width=0.11\textwidth]{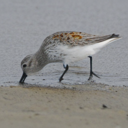}  & \includegraphics[width=0.11\textwidth]{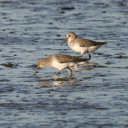}  & \includegraphics[width=0.11\textwidth]{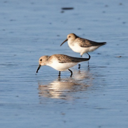}  & \includegraphics[width=0.11\textwidth]{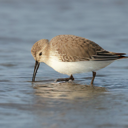}  & \includegraphics[width=0.11\textwidth]{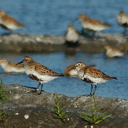}  & \includegraphics[width=0.11\textwidth]{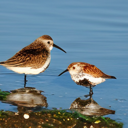}  & \includegraphics[width=0.11\textwidth]{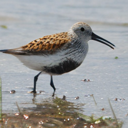} \\
\rotatebox{90}{\parbox{1.64cm}{\centering \textbf{\hspace{0pt} Sea anemone}}} & \includegraphics[width=0.11\textwidth]{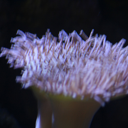}  & \includegraphics[width=0.11\textwidth]{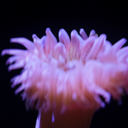}  & \includegraphics[width=0.11\textwidth]{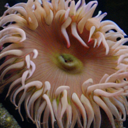}  & \includegraphics[width=0.11\textwidth]{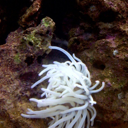}  & \includegraphics[width=0.11\textwidth]{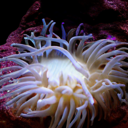}  & \includegraphics[width=0.11\textwidth]{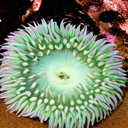}  & \includegraphics[width=0.11\textwidth]{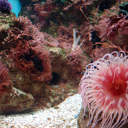}  & \includegraphics[width=0.11\textwidth]{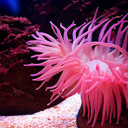}  & \includegraphics[width=0.11\textwidth]{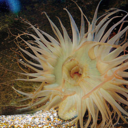} \\
\rotatebox{90}{\parbox{1.64cm}{\centering \textbf{\hspace{0pt} Snail}}} & \includegraphics[width=0.11\textwidth]{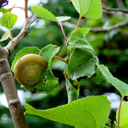}  & \includegraphics[width=0.11\textwidth]{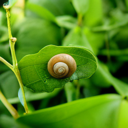}  & \includegraphics[width=0.11\textwidth]{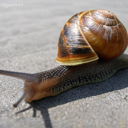}  & \includegraphics[width=0.11\textwidth]{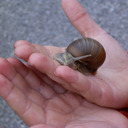}  & \includegraphics[width=0.11\textwidth]{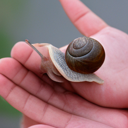}  & \includegraphics[width=0.11\textwidth]{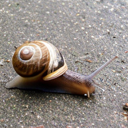}  & \includegraphics[width=0.11\textwidth]{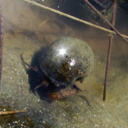}  & \includegraphics[width=0.11\textwidth]{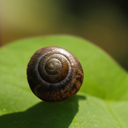}  & \includegraphics[width=0.11\textwidth]{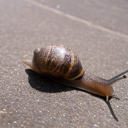} \\
    \end{tabular}
    }
    \captionof{figure}{Visualizations of \Cano{}s using \Canoimg{}s obtained from the DiT \cite{dit} used in our \textbf{\textit{\CanoDistill{}}} experiments on ImageNet.}
    \label{tab:supp-visual-3}
\end{table}

\begin{table}[]
    \centering
    \resizebox{\linewidth}{!}{
    \begin{tabular}{c|ccc|ccc|ccc}
    Class & Original & CFG & \MethodName{} & Original & CFG & \MethodName{} & Original & CFG & \MethodName{} \\
    \midrule
    \rotatebox{90}{\parbox{1.64cm}{\centering \textbf{\hspace{0pt} Beer bottle}}} & \includegraphics[width=0.11\textwidth]{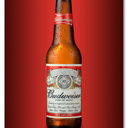}  & \includegraphics[width=0.11\textwidth]{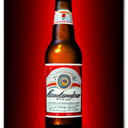}  & \includegraphics[width=0.11\textwidth]{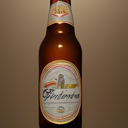}  & \includegraphics[width=0.11\textwidth]{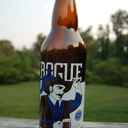}  & \includegraphics[width=0.11\textwidth]{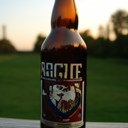}  & \includegraphics[width=0.11\textwidth]{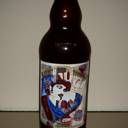}  & \includegraphics[width=0.11\textwidth]{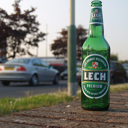}  & \includegraphics[width=0.11\textwidth]{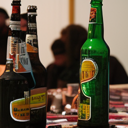}  & \includegraphics[width=0.11\textwidth]{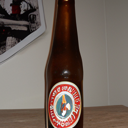} \\
\rotatebox{90}{\parbox{1.64cm}{\centering \textbf{\hspace{0pt} Cab}}} & \includegraphics[width=0.11\textwidth]{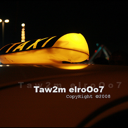}  & \includegraphics[width=0.11\textwidth]{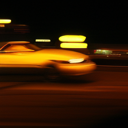}  & \includegraphics[width=0.11\textwidth]{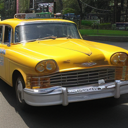}  & \includegraphics[width=0.11\textwidth]{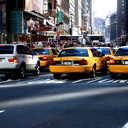}  & \includegraphics[width=0.11\textwidth]{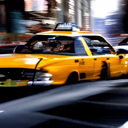}  & \includegraphics[width=0.11\textwidth]{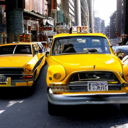}  & \includegraphics[width=0.11\textwidth]{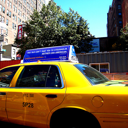}  & \includegraphics[width=0.11\textwidth]{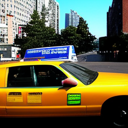}  & \includegraphics[width=0.11\textwidth]{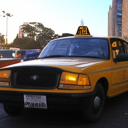} \\
\rotatebox{90}{\parbox{1.64cm}{\centering \textbf{\hspace{0pt} Crossword puzzle}}} & \includegraphics[width=0.11\textwidth]{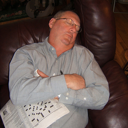}  & \includegraphics[width=0.11\textwidth]{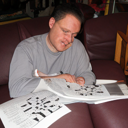}  & \includegraphics[width=0.11\textwidth]{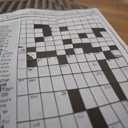}  & \includegraphics[width=0.11\textwidth]{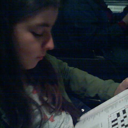}  & \includegraphics[width=0.11\textwidth]{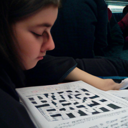}  & \includegraphics[width=0.11\textwidth]{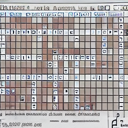}  & \includegraphics[width=0.11\textwidth]{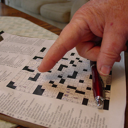}  & \includegraphics[width=0.11\textwidth]{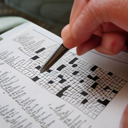}  & \includegraphics[width=0.11\textwidth]{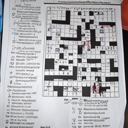} \\
\rotatebox{90}{\parbox{1.64cm}{\centering \textbf{\hspace{0pt} Folding chair}}} & \includegraphics[width=0.11\textwidth]{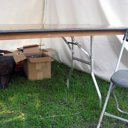}  & \includegraphics[width=0.11\textwidth]{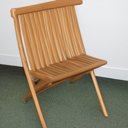}  & \includegraphics[width=0.11\textwidth]{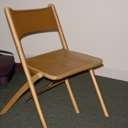}  & \includegraphics[width=0.11\textwidth]{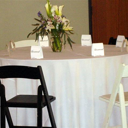}  & \includegraphics[width=0.11\textwidth]{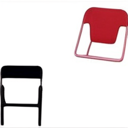}  & \includegraphics[width=0.11\textwidth]{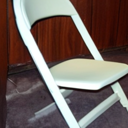}  & \includegraphics[width=0.11\textwidth]{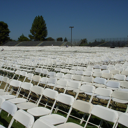}  & \includegraphics[width=0.11\textwidth]{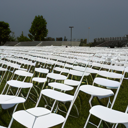}  & \includegraphics[width=0.11\textwidth]{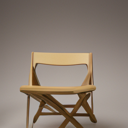} \\
\rotatebox{90}{\parbox{1.64cm}{\centering \textbf{\hspace{0pt} Garbage truck}}} & \includegraphics[width=0.11\textwidth]{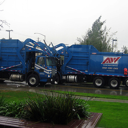}  & \includegraphics[width=0.11\textwidth]{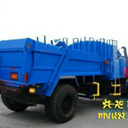}  & \includegraphics[width=0.11\textwidth]{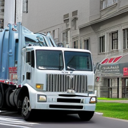}  & \includegraphics[width=0.11\textwidth]{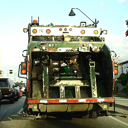}  & \includegraphics[width=0.11\textwidth]{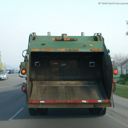}  & \includegraphics[width=0.11\textwidth]{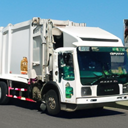}  & \includegraphics[width=0.11\textwidth]{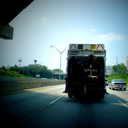}  & \includegraphics[width=0.11\textwidth]{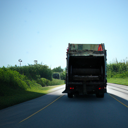}  & \includegraphics[width=0.11\textwidth]{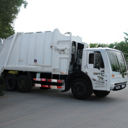} \\
\rotatebox{90}{\parbox{1.64cm}{\centering \textbf{\hspace{0pt} Ibex}}} & \includegraphics[width=0.11\textwidth]{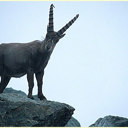}  & \includegraphics[width=0.11\textwidth]{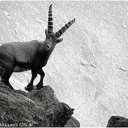}  & \includegraphics[width=0.11\textwidth]{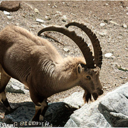}  & \includegraphics[width=0.11\textwidth]{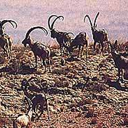}  & \includegraphics[width=0.11\textwidth]{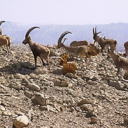}  & \includegraphics[width=0.11\textwidth]{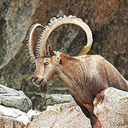}  & \includegraphics[width=0.11\textwidth]{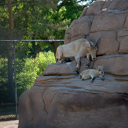}  & \includegraphics[width=0.11\textwidth]{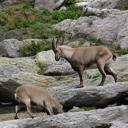}  & \includegraphics[width=0.11\textwidth]{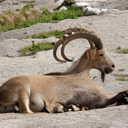} \\
\rotatebox{90}{\parbox{1.64cm}{\centering \textbf{\hspace{0pt} Lemon}}} & \includegraphics[width=0.11\textwidth]{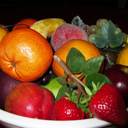}  & \includegraphics[width=0.11\textwidth]{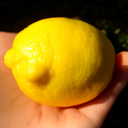}  & \includegraphics[width=0.11\textwidth]{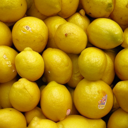}  & \includegraphics[width=0.11\textwidth]{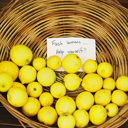}  & \includegraphics[width=0.11\textwidth]{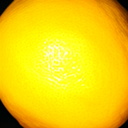}  & \includegraphics[width=0.11\textwidth]{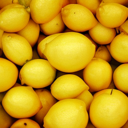}  & \includegraphics[width=0.11\textwidth]{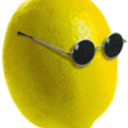}  & \includegraphics[width=0.11\textwidth]{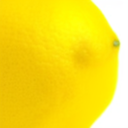}  & \includegraphics[width=0.11\textwidth]{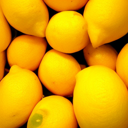} \\
\rotatebox{90}{\parbox{1.64cm}{\centering \textbf{\hspace{0pt} Orange}}} & \includegraphics[width=0.11\textwidth]{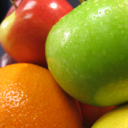}  & \includegraphics[width=0.11\textwidth]{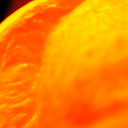}  & \includegraphics[width=0.11\textwidth]{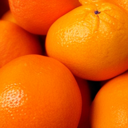}  & \includegraphics[width=0.11\textwidth]{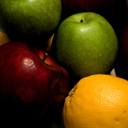}  & \includegraphics[width=0.11\textwidth]{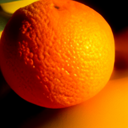}  & \includegraphics[width=0.11\textwidth]{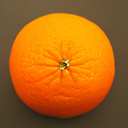}  & \includegraphics[width=0.11\textwidth]{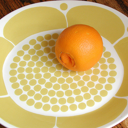}  & \includegraphics[width=0.11\textwidth]{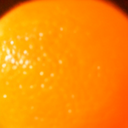}  & \includegraphics[width=0.11\textwidth]{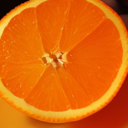} \\
\rotatebox{90}{\parbox{1.64cm}{\centering \textbf{\hspace{0pt} Television}}} & \includegraphics[width=0.11\textwidth]{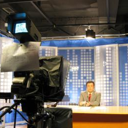}  & \includegraphics[width=0.11\textwidth]{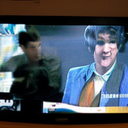}  & \includegraphics[width=0.11\textwidth]{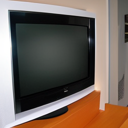}  & \includegraphics[width=0.11\textwidth]{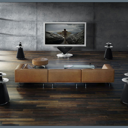}  & \includegraphics[width=0.11\textwidth]{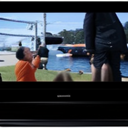}  & \includegraphics[width=0.11\textwidth]{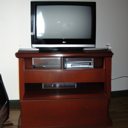}  & \includegraphics[width=0.11\textwidth]{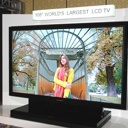}  & \includegraphics[width=0.11\textwidth]{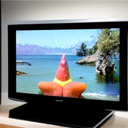}  & \includegraphics[width=0.11\textwidth]{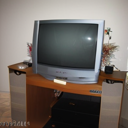} \\
\rotatebox{90}{\parbox{1.64cm}{\centering \textbf{\hspace{0pt} Toaster}}} & \includegraphics[width=0.11\textwidth]{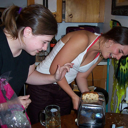}  & \includegraphics[width=0.11\textwidth]{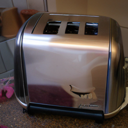}  & \includegraphics[width=0.11\textwidth]{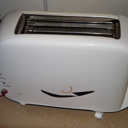}  & \includegraphics[width=0.11\textwidth]{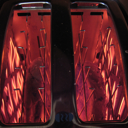}  & \includegraphics[width=0.11\textwidth]{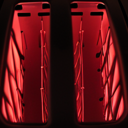}  & \includegraphics[width=0.11\textwidth]{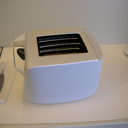}  & \includegraphics[width=0.11\textwidth]{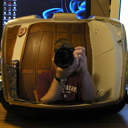}  & \includegraphics[width=0.11\textwidth]{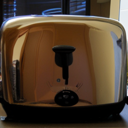}  & \includegraphics[width=0.11\textwidth]{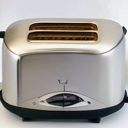} \\
\rotatebox{90}{\parbox{1.64cm}{\centering \textbf{\hspace{0pt} Traffic light}}} & \includegraphics[width=0.11\textwidth]{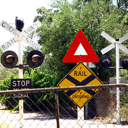}  & \includegraphics[width=0.11\textwidth]{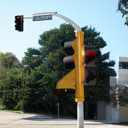}  & \includegraphics[width=0.11\textwidth]{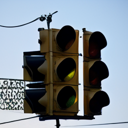}  & \includegraphics[width=0.11\textwidth]{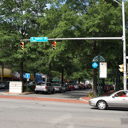}  & \includegraphics[width=0.11\textwidth]{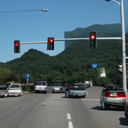}  & \includegraphics[width=0.11\textwidth]{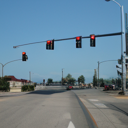}  & \includegraphics[width=0.11\textwidth]{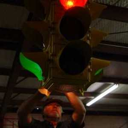}  & \includegraphics[width=0.11\textwidth]{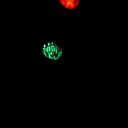}  & \includegraphics[width=0.11\textwidth]{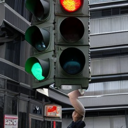} \\
\rotatebox{90}{\parbox{1.64cm}{\centering \textbf{\hspace{0pt} Upright}}} & \includegraphics[width=0.11\textwidth]{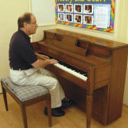}  & \includegraphics[width=0.11\textwidth]{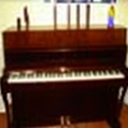}  & \includegraphics[width=0.11\textwidth]{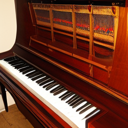}  & \includegraphics[width=0.11\textwidth]{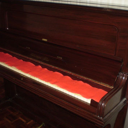}  & \includegraphics[width=0.11\textwidth]{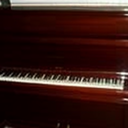}  & \includegraphics[width=0.11\textwidth]{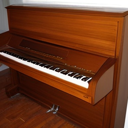}  & \includegraphics[width=0.11\textwidth]{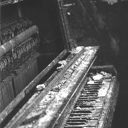}  & \includegraphics[width=0.11\textwidth]{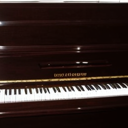}  & \includegraphics[width=0.11\textwidth]{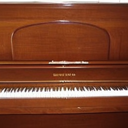} \\
\rotatebox{90}{\parbox{1.64cm}{\centering \textbf{\hspace{0pt} Violin}}} & \includegraphics[width=0.11\textwidth]{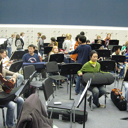}  & \includegraphics[width=0.11\textwidth]{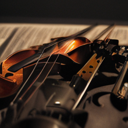}  & \includegraphics[width=0.11\textwidth]{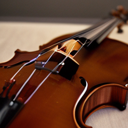}  & \includegraphics[width=0.11\textwidth]{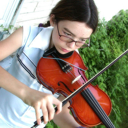}  & \includegraphics[width=0.11\textwidth]{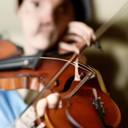}  & \includegraphics[width=0.11\textwidth]{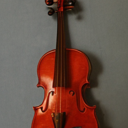}  & \includegraphics[width=0.11\textwidth]{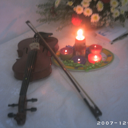}  & \includegraphics[width=0.11\textwidth]{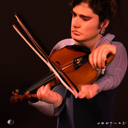}  & \includegraphics[width=0.11\textwidth]{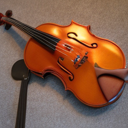} \\
\rotatebox{90}{\parbox{1.64cm}{\centering \textbf{\hspace{0pt} Water bottle}}} & \includegraphics[width=0.11\textwidth]{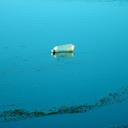}  & \includegraphics[width=0.11\textwidth]{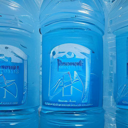}  & \includegraphics[width=0.11\textwidth]{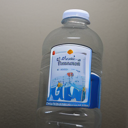}  & \includegraphics[width=0.11\textwidth]{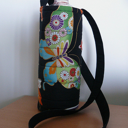}  & \includegraphics[width=0.11\textwidth]{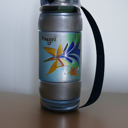}  & \includegraphics[width=0.11\textwidth]{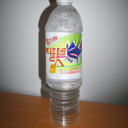}  & \includegraphics[width=0.11\textwidth]{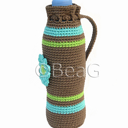}  & \includegraphics[width=0.11\textwidth]{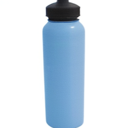}  & \includegraphics[width=0.11\textwidth]{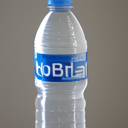} \\
\rotatebox{90}{\parbox{1.64cm}{\centering \textbf{\hspace{0pt} Wine bottle}}} & \includegraphics[width=0.11\textwidth]{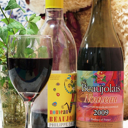}  & \includegraphics[width=0.11\textwidth]{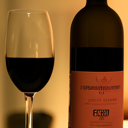}  & \includegraphics[width=0.11\textwidth]{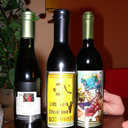}  & \includegraphics[width=0.11\textwidth]{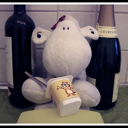}  & \includegraphics[width=0.11\textwidth]{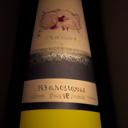}  & \includegraphics[width=0.11\textwidth]{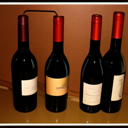}  & \includegraphics[width=0.11\textwidth]{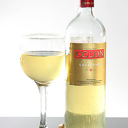}  & \includegraphics[width=0.11\textwidth]{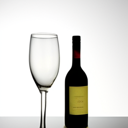}  & \includegraphics[width=0.11\textwidth]{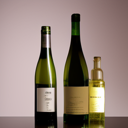} \\
\rotatebox{90}{\parbox{1.64cm}{\centering \textbf{\hspace{0pt} Wreck}}} & \includegraphics[width=0.11\textwidth]{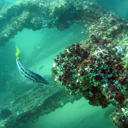}  & \includegraphics[width=0.11\textwidth]{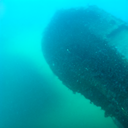}  & \includegraphics[width=0.11\textwidth]{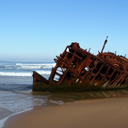}  & \includegraphics[width=0.11\textwidth]{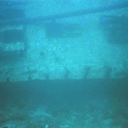}  & \includegraphics[width=0.11\textwidth]{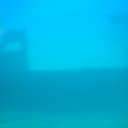}  & \includegraphics[width=0.11\textwidth]{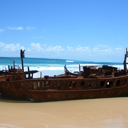}  & \includegraphics[width=0.11\textwidth]{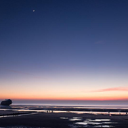}  & \includegraphics[width=0.11\textwidth]{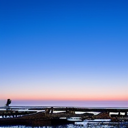}  & \includegraphics[width=0.11\textwidth]{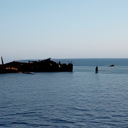} \\
    \end{tabular}
    }
    \captionof{figure}{Visualizations of \Cano{}s using \Canoimg{}s obtained from the DiT \cite{dit} used in our \textbf{\textit{\CanoDistill{}}} experiments on ImageNet.}
    \label{tab:supp-visual-4}
\end{table}

\begin{table}[]
    \centering
    \resizebox{\linewidth}{!}{
    \begin{tabular}{c|ccc|ccc|ccc}
    Class & Original & CFG & \MethodName{} & Original & CFG & \MethodName{} & Original & CFG & \MethodName{} \\
    \midrule
    \rotatebox{90}{\parbox{1.64cm}{\centering \textbf{\hspace{0pt} Crock pot}}} & \includegraphics[width=0.11\textwidth]{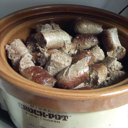}  & \includegraphics[width=0.11\textwidth]{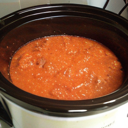}  & \includegraphics[width=0.11\textwidth]{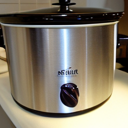}  & \includegraphics[width=0.11\textwidth]{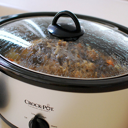}  & \includegraphics[width=0.11\textwidth]{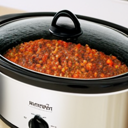}  & \includegraphics[width=0.11\textwidth]{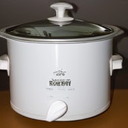}  & \includegraphics[width=0.11\textwidth]{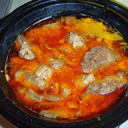}  & \includegraphics[width=0.11\textwidth]{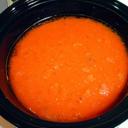}  & \includegraphics[width=0.11\textwidth]{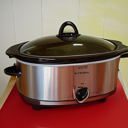} \\
\rotatebox{90}{\parbox{1.64cm}{\centering \textbf{\hspace{0pt} Petri dish}}} & \includegraphics[width=0.11\textwidth]{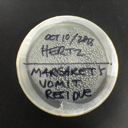}  & \includegraphics[width=0.11\textwidth]{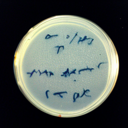}  & \includegraphics[width=0.11\textwidth]{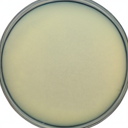}  & \includegraphics[width=0.11\textwidth]{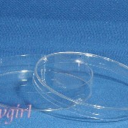}  & \includegraphics[width=0.11\textwidth]{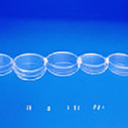}  & \includegraphics[width=0.11\textwidth]{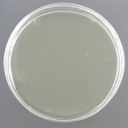}  & \includegraphics[width=0.11\textwidth]{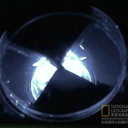}  & \includegraphics[width=0.11\textwidth]{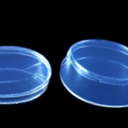}  & \includegraphics[width=0.11\textwidth]{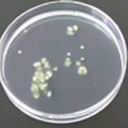} \\
\rotatebox{90}{\parbox{1.64cm}{\centering \textbf{\hspace{0pt} Beaker}}} & \includegraphics[width=0.11\textwidth]{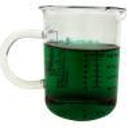}  & \includegraphics[width=0.11\textwidth]{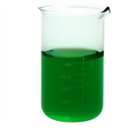}  & \includegraphics[width=0.11\textwidth]{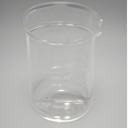}  & \includegraphics[width=0.11\textwidth]{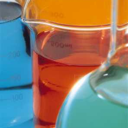}  & \includegraphics[width=0.11\textwidth]{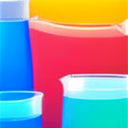}  & \includegraphics[width=0.11\textwidth]{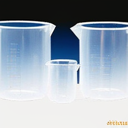}  & \includegraphics[width=0.11\textwidth]{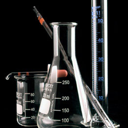}  & \includegraphics[width=0.11\textwidth]{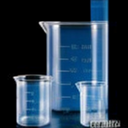}  & \includegraphics[width=0.11\textwidth]{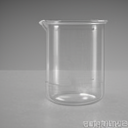} \\
\rotatebox{90}{\parbox{1.64cm}{\centering \textbf{\hspace{0pt} Broom}}} & \includegraphics[width=0.11\textwidth]{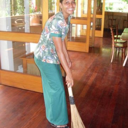}  & \includegraphics[width=0.11\textwidth]{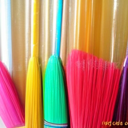}  & \includegraphics[width=0.11\textwidth]{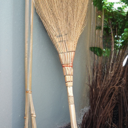}  & \includegraphics[width=0.11\textwidth]{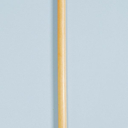}  & \includegraphics[width=0.11\textwidth]{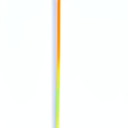}  & \includegraphics[width=0.11\textwidth]{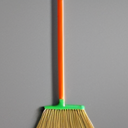}  & \includegraphics[width=0.11\textwidth]{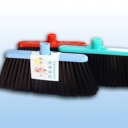}  & \includegraphics[width=0.11\textwidth]{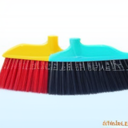}  & \includegraphics[width=0.11\textwidth]{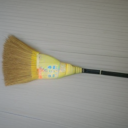} \\
\rotatebox{90}{\parbox{1.64cm}{\centering \textbf{\hspace{0pt} Cocktail shaker}}} & \includegraphics[width=0.11\textwidth]{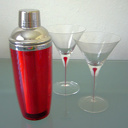}  & \includegraphics[width=0.11\textwidth]{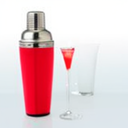}  & \includegraphics[width=0.11\textwidth]{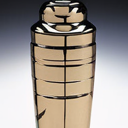}  & \includegraphics[width=0.11\textwidth]{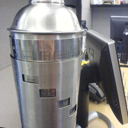}  & \includegraphics[width=0.11\textwidth]{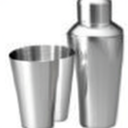}  & \includegraphics[width=0.11\textwidth]{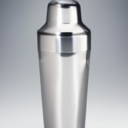}  & \includegraphics[width=0.11\textwidth]{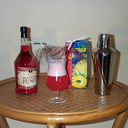}  & \includegraphics[width=0.11\textwidth]{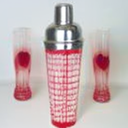}  & \includegraphics[width=0.11\textwidth]{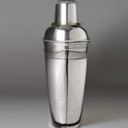} \\
\rotatebox{90}{\parbox{1.64cm}{\centering \textbf{\hspace{0pt} Coffee mug}}} & \includegraphics[width=0.11\textwidth]{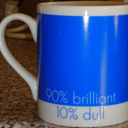}  & \includegraphics[width=0.11\textwidth]{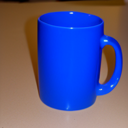}  & \includegraphics[width=0.11\textwidth]{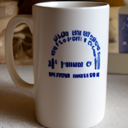}  & \includegraphics[width=0.11\textwidth]{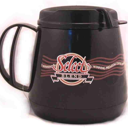}  & \includegraphics[width=0.11\textwidth]{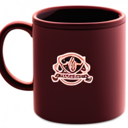}  & \includegraphics[width=0.11\textwidth]{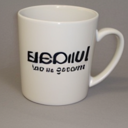}  & \includegraphics[width=0.11\textwidth]{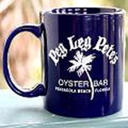}  & \includegraphics[width=0.11\textwidth]{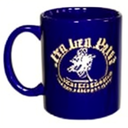}  & \includegraphics[width=0.11\textwidth]{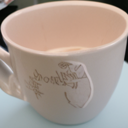} \\
\rotatebox{90}{\parbox{1.64cm}{\centering \textbf{\hspace{0pt} Cornet}}} & \includegraphics[width=0.11\textwidth]{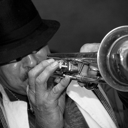}  & \includegraphics[width=0.11\textwidth]{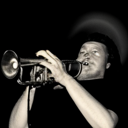}  & \includegraphics[width=0.11\textwidth]{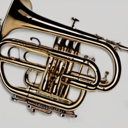}  & \includegraphics[width=0.11\textwidth]{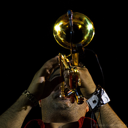}  & \includegraphics[width=0.11\textwidth]{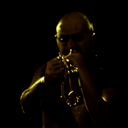}  & \includegraphics[width=0.11\textwidth]{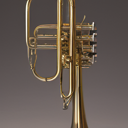}  & \includegraphics[width=0.11\textwidth]{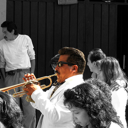}  & \includegraphics[width=0.11\textwidth]{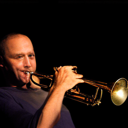}  & \includegraphics[width=0.11\textwidth]{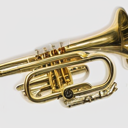} \\
\rotatebox{90}{\parbox{1.64cm}{\centering \textbf{\hspace{0pt} Crane}}} & \includegraphics[width=0.11\textwidth]{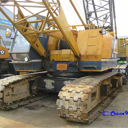}  & \includegraphics[width=0.11\textwidth]{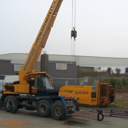}  & \includegraphics[width=0.11\textwidth]{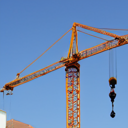}  & \includegraphics[width=0.11\textwidth]{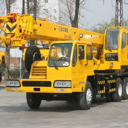}  & \includegraphics[width=0.11\textwidth]{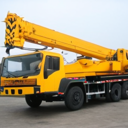}  & \includegraphics[width=0.11\textwidth]{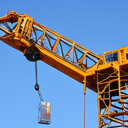}  & \includegraphics[width=0.11\textwidth]{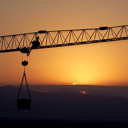}  & \includegraphics[width=0.11\textwidth]{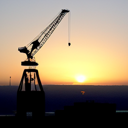}  & \includegraphics[width=0.11\textwidth]{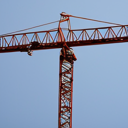} \\
\rotatebox{90}{\parbox{1.64cm}{\centering \textbf{\hspace{0pt} Honeycomb}}} & \includegraphics[width=0.11\textwidth]{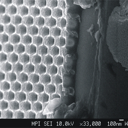}  & \includegraphics[width=0.11\textwidth]{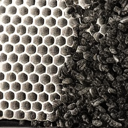}  & \includegraphics[width=0.11\textwidth]{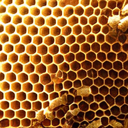}  & \includegraphics[width=0.11\textwidth]{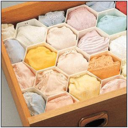}  & \includegraphics[width=0.11\textwidth]{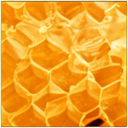}  & \includegraphics[width=0.11\textwidth]{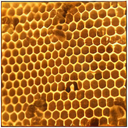}  & \includegraphics[width=0.11\textwidth]{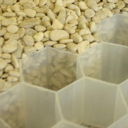}  & \includegraphics[width=0.11\textwidth]{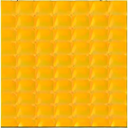}  & \includegraphics[width=0.11\textwidth]{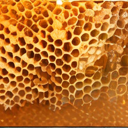} \\
\rotatebox{90}{\parbox{1.64cm}{\centering \textbf{\hspace{0pt} Jaguar}}} & \includegraphics[width=0.11\textwidth]{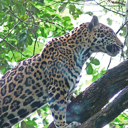}  & \includegraphics[width=0.11\textwidth]{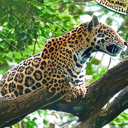}  & \includegraphics[width=0.11\textwidth]{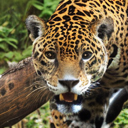}  & \includegraphics[width=0.11\textwidth]{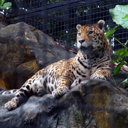}  & \includegraphics[width=0.11\textwidth]{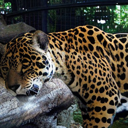}  & \includegraphics[width=0.11\textwidth]{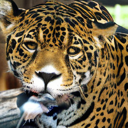}  & \includegraphics[width=0.11\textwidth]{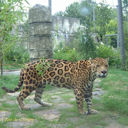}  & \includegraphics[width=0.11\textwidth]{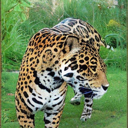}  & \includegraphics[width=0.11\textwidth]{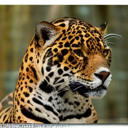} \\
\rotatebox{90}{\parbox{1.64cm}{\centering \textbf{\hspace{0pt} Mosquito net}}} & \includegraphics[width=0.11\textwidth]{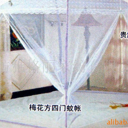}  & \includegraphics[width=0.11\textwidth]{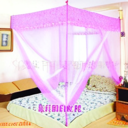}  & \includegraphics[width=0.11\textwidth]{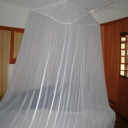}  & \includegraphics[width=0.11\textwidth]{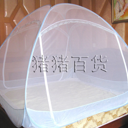}  & \includegraphics[width=0.11\textwidth]{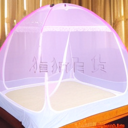}  & \includegraphics[width=0.11\textwidth]{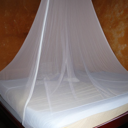}  & \includegraphics[width=0.11\textwidth]{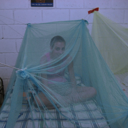}  & \includegraphics[width=0.11\textwidth]{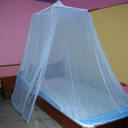}  & \includegraphics[width=0.11\textwidth]{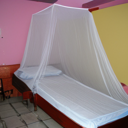} \\
\rotatebox{90}{\parbox{1.64cm}{\centering \textbf{\hspace{0pt} Mountain bike}}} & \includegraphics[width=0.11\textwidth]{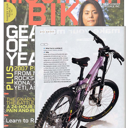}  & \includegraphics[width=0.11\textwidth]{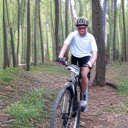}  & \includegraphics[width=0.11\textwidth]{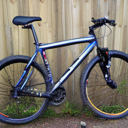}  & \includegraphics[width=0.11\textwidth]{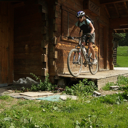}  & \includegraphics[width=0.11\textwidth]{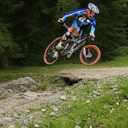}  & \includegraphics[width=0.11\textwidth]{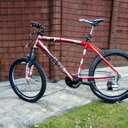}  & \includegraphics[width=0.11\textwidth]{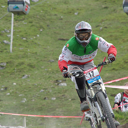}  & \includegraphics[width=0.11\textwidth]{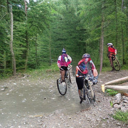}  & \includegraphics[width=0.11\textwidth]{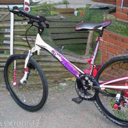} \\
\rotatebox{90}{\parbox{1.64cm}{\centering \textbf{\hspace{0pt} Nail}}} & \includegraphics[width=0.11\textwidth]{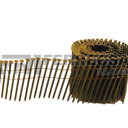}  & \includegraphics[width=0.11\textwidth]{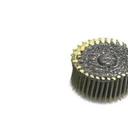}  & \includegraphics[width=0.11\textwidth]{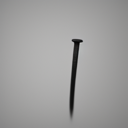}  & \includegraphics[width=0.11\textwidth]{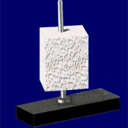}  & \includegraphics[width=0.11\textwidth]{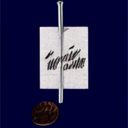}  & \includegraphics[width=0.11\textwidth]{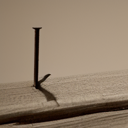}  & \includegraphics[width=0.11\textwidth]{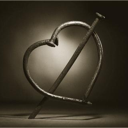}  & \includegraphics[width=0.11\textwidth]{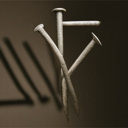}  & \includegraphics[width=0.11\textwidth]{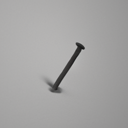} \\
\rotatebox{90}{\parbox{1.64cm}{\centering \textbf{\hspace{0pt} Park bench}}} & \includegraphics[width=0.11\textwidth]{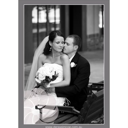}  & \includegraphics[width=0.11\textwidth]{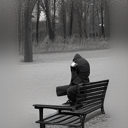}  & \includegraphics[width=0.11\textwidth]{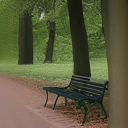}  & \includegraphics[width=0.11\textwidth]{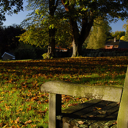}  & \includegraphics[width=0.11\textwidth]{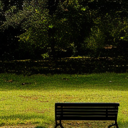}  & \includegraphics[width=0.11\textwidth]{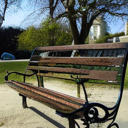}  & \includegraphics[width=0.11\textwidth]{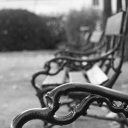}  & \includegraphics[width=0.11\textwidth]{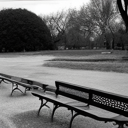}  & \includegraphics[width=0.11\textwidth]{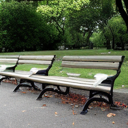} \\
\rotatebox{90}{\parbox{1.64cm}{\centering \textbf{\hspace{0pt} Sturgeon}}} & \includegraphics[width=0.11\textwidth]{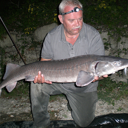}  & \includegraphics[width=0.11\textwidth]{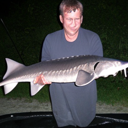}  & \includegraphics[width=0.11\textwidth]{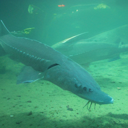}  & \includegraphics[width=0.11\textwidth]{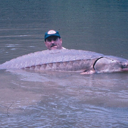}  & \includegraphics[width=0.11\textwidth]{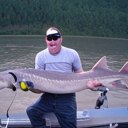}  & \includegraphics[width=0.11\textwidth]{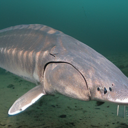}  & \includegraphics[width=0.11\textwidth]{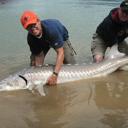}  & \includegraphics[width=0.11\textwidth]{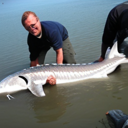}  & \includegraphics[width=0.11\textwidth]{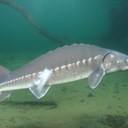} \\
\rotatebox{90}{\parbox{1.64cm}{\centering \textbf{\hspace{0pt} Tiger}}} & \includegraphics[width=0.11\textwidth]{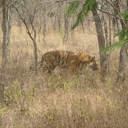}  & \includegraphics[width=0.11\textwidth]{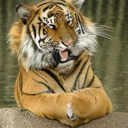}  & \includegraphics[width=0.11\textwidth]{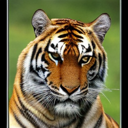}  & \includegraphics[width=0.11\textwidth]{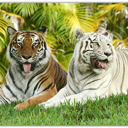}  & \includegraphics[width=0.11\textwidth]{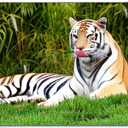}  & \includegraphics[width=0.11\textwidth]{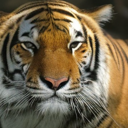}  & \includegraphics[width=0.11\textwidth]{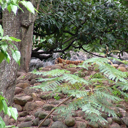}  & \includegraphics[width=0.11\textwidth]{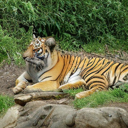}  & \includegraphics[width=0.11\textwidth]{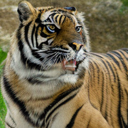} \\
    \end{tabular}
    }
    \captionof{figure}{Visualizations of \Cano{}s using \Canoimg{}s obtained from the DiT \cite{dit} used in our \textbf{\textit{\CanoDistill{}}} experiments on ImageNet.}
    \label{tab:supp-visual-5}
\end{table}

\section{Broader Impact}
\label{sec:broader-impact}
\MethodName{} introduces a new avenue into the field of DM research, focusing on interpreting the discriminative signals inside CDMs rather than directly probing the raw feature space. Our findings advance the interpretability of CDMs, contributing to safer usage of them. The application of \Cano{}s challenges the common assumption that the usage of DMs for discriminative tasks necessitates a large amount of data, providing new directions in diffusion-based feature distillation. The image generators used in our experiments may sometimes contain wrong information. That said, we are not focusing on generating high-fidelity images or improving the synthesizing quality.

\section{License information}
\subsection{Datasets information and license}
\label{sec:dataset-info}
\begin{itemize}[leftmargin=*]
    \item ImageNet1K \cite{deng2009imagenet}. This dataset contains 1.28M training images and 50000 images for validation. We report the top1 accuracy on the 50000 validation images. License: \href{https://image-net.org/}{Custom (research, non-commercial)}.
    \item ImageNet-C \cite{imagenet-c}. This dataset contains 15 types of 2D image corruption types that are generated by different algorithms. Higher accuracy on this dataset indicates a more robust model against corrupted images. License: CC BY 4.0. 
    \item ImageNet-A \cite{imagenet-a}. This dataset contains naturally existing adversarial examples that can drastically decrease the accuracy of ImageNet1K-trained CNNs. It is a 200-class subset of the ImageNet1K dataset. License: MIT license.
    \item ImageNet Reassessed Labels (ImageNet-ReaL) \cite{im-real}: This is a dataset with 50000 reassessed labels of the ImageNet validation set, aiming at testing the in-distribution generalization ability of a classifier. License: Apache 2.0 License.
    \item CIFAR10-C \cite{imagenet-c}. This dataset contains 15 types of 2D image corruption types that are generated by different algorithms. Higher accuracy on this dataset indicates a more robust model against corrupted images. License: CC BY 4.0. 
\end{itemize}

\subsection{Model and code license}
\label{sec:model-license}
\begin{itemize}[leftmargin=*]
\item Code for adversarial attacks \cite{attack-github}: MIT License.
\item PyTorch Image Model \cite{timm}: Apache 2.0 License.
\item Diffusion Transformer (DiT) \cite{dit}: Attribution-NonCommercial 4.0 International.
\item Stable Diffusion 2.1 \cite{ldm}: CreativeML Open RAIL++-M License.
\item EDM2 \cite{edm2}: Creative Commons BY-NC-SA 4.0 license. 
\item Supervised Contrastive Learning \cite{supcon}: BSD 2-Clause License.
\item Swin \cite{swin}: MIT License.
\item FAN \cite{fan}: Nvidia Source Code License-NC.
\end{itemize}




\end{document}